\documentclass{article}

\usepackage{microtype}
\usepackage{graphicx}
\usepackage{babel,blindtext}
\usepackage{subcaption}
\usepackage{booktabs} % for professional tables
\usepackage{url}

\usepackage{hyperref}

\usepackage[accepted]{icml2026}

\usepackage{xurl}
\usepackage{hyperref}
\usepackage{amsmath}
\usepackage{amssymb}
\usepackage{mathtools}
\usepackage{amsthm}
\usepackage{enumitem}

\newcommand{\Ko}{\mathcal{K}}              % final selected latent set
\newcommand{\Kopos}{\mathcal{K}^{+}}       % positives in K0 (Delta_k > 0)
\newcommand{\Koneg}{\mathcal{K}^{-}}       % negatives in K0 (Delta_k < 0)
\newcommand{\Kreem}{\Ko^{\mathrm{reem}}}

\newcommand{\Cpos}{\mathcal{C}^{+}}        % candidate pool, positive shifts
\newcommand{\Cneg}{\mathcal{C}^{-}}        % candidate pool, negative shifts
\newcommand{\Cset}{\mathcal{C}}            % full candidate pool
\newcommand{\Ktilde}{\widetilde{\mathcal{K}}} % shortlisted candidates after screening

\newcommand{\Kidx}{k}                      % latent index
\newcommand{\mLat}{m}                      % number of SAE latents
\newcommand{\Tlen}{T}                      % sequence length / token count

\newcommand{\tht}{\theta}                  % current (trainable) model params
\newcommand{\thbase}{{\mathrm{base}}} % frozen base params
\newcommand{\thmis}{{\mathrm{mis}}}   % misaligned checkpoint params
\newcommand{\threem}{{\mathrm{reem}}}

\newcommand{\basemodel}{\mathcal{M}^{\thbase}}
\newcommand{\reemergentmodel}{\mathcal{M}^{\threem}}
\newcommand{\misalignedmodel}{\mathcal{M}^{\thmis}}

\newcommand{\xe}{x}                        % input (prompt / sequence)

\newcommand{\coremisalignment}{\texttt{core misalignment}}
\newcommand{\finalevaluation}{\texttt{final evaluation}}

\newcommand{\domain}[1]{\texttt{#1 domain}}
\newcommand{\domainname}[1]{\emph{#1}}

\newcommand{\coremisalignmentdataset}{\mathcal{D}_{\substack{\mathrm{core}\\\mathrm{mis}}}}

\newcommand{\Lay}{L}                       % chosen layer index
\newcommand{\hL}{h_{\Lay}}                 % hidden states at layer L (tokenwise)
\newcommand{\hLt}{h_{\Lay,t}}                 % hidden states at layer L (tokenwise)

\newcommand{\hdim}{d}                      % hidden dimension

\newcommand{\zbar}{\bar z}                 % token-averaged latent activation
\newcommand{\dk}{d_{\Kidx}}                % SAE decoder vector for latent k
\newcommand{\alphaS}{\alpha}               % steering strength

\newcommand{\sScale}{s}                    % steering scale normalization

\newcommand{\Tpos}{\mathcal{T}(\xe)}       % supervised positions for input x

\newcommand{\Llatent}{\mathcal{L}_{\mathrm{block}}}
\newcommand{\Lsft}{\mathcal{L}_{\mathrm{SFT}}}
\newcommand{\Ltotal}{\mathcal{L}_{\mathrm{total}}}

\newcommand{\lam}{\lambda}                 % constraint weight

\newcommand{\tauQ}{\tau}                   % quality/incoherence budget threshold
\newcommand{\Agrid}{\mathcal{A}}           % steering grid for alpha sweep

\newcommand{\alphaStarInd}{\alpha^{\star}_{\mathrm{ind}}}
\newcommand{\alphaStarRep}{\alpha^{\star}_{\mathrm{rep}}}
\newcommand{\alphaPhaseTwoPos}{\alpha^{\mathrm{stage2}}_{\mathrm{ind}}}
\newcommand{\alphaPhaseTwoNeg}{\alpha^{\mathrm{stage2}}_{\mathrm{rep}}}

\newcommand{\E}{\mathbb{E}}                % expectation
\newcommand{\ReLU}{\mathrm{ReLU}}          % ReLU operator
\newcommand{\TopN}{\mathrm{TopN}}          % TopN operator (as used in text)

\newcommand{\ztok}[4]{z^{(#1)}_{#2,#3}(#4)} % params, token, latent, input
\newcommand{\zcurtok}[3]{\ztok{\tht}{#1}{#2}{#3}}
\newcommand{\zbasetok}[3]{\ztok{\thbase}{#1}{#2}{#3}}

\newcommand{\tidx}{t}
\newcommand{\TsfT}{\mathcal{T}_{\mathrm{SFT}}}
\newcommand{\zcurtokX}{\ztok{\tht}{\tidx}{\Kidx}{\xe}}
\newcommand{\zbasetokX}{\ztok{\thbase}{\tidx}{\Kidx}{\xe}}

\newcommand{\zbarMisX}[1][\Kidx]{\zbar_{#1}^{(\thmis)}(\xe)}
\newcommand{\zbarBaseX}[1][\Kidx]{\zbar_{#1}^{(\thbase)}(\xe)}

\newcommand{\DeltaK}[1][\Kidx]{\Delta_{#1}}   % e.g., \DeltaK or \DeltaK[k]

\newcommand{\dhatK}[1][\Kidx]{\hat d_{#1}}

\newcommand{\Lblk}{L_{\mathrm{blk}}}

\newcommand{\R}{\mathbb{R}}

\newcommand{\h}{h}

\usepackage[capitalize,noabbrev]{cleveref}

\theoremstyle{plain}

\theoremstyle{definition}

\theoremstyle{remark}

\usepackage[textsize=tiny]{todonotes}

\icmltitlerunning{BLOCK-EM: Preventing Emergent Misalignment via Latent Blocking}

\begin{document}
% BLOCK-EM: Neuroblocking Emergent Misalignment
\twocolumn[
  \icmltitle{BLOCK-EM: Preventing Emergent Misalignment via Latent Blocking}



  \begin{icmlauthorlist}
    \icmlauthor{Muhammed Ustaomeroglu}{yyy}
    \icmlauthor{Guannan Qu}{yyy}
  \end{icmlauthorlist}

  \icmlaffiliation{yyy}{Carnegie Mellon Universiy}

  \icmlcorrespondingauthor{Muhammed Ustaomeroglu}{mustaome@andrew.cmu.edu}
  \icmlkeywords{Machine Learning, ICML}
  \vskip 0.3in
]



\printAffiliationsAndNotice{}  % no special notice (required even if empty)

%Suppressing Emergent Misalignment
%Suppressor for Emergent Misalignment

%(*) Preventing Emergent Misalignment via Latent Blocking

%Preventing Emergent Misalignment via Latent Constraining
\begin{abstract}
    Emergent misalignment can arise when a language model is fine-tuned on a narrowly scoped supervised objective: the model learns the target behavior, yet also develops undesirable out-of-domain behaviors. We investigate a mechanistic approach to preventing emergent misalignment by identifying a small set of internal features that reliably control the misaligned behavior and then discouraging the model from strengthening these features during fine-tuning. Across six fine-tuning domains, blocking (i.e., constraining) a fixed set of features achieves up to 95\% relative reduction in emergent misalignment with no degradation in model quality or target-task performance.
    We strengthen validity with disjoint selection/evaluation splits, multiple independent judges, multiple random seeds for key settings, quality metrics, and extensive ablations demonstrating that the reduction in misalignment is specific to the identified mechanism. 
    We also characterize a limiting regime in which misalignment re-emerges under prolonged fine-tuning, present evidence consistent with rerouting through alternative features or layers, and evaluate modifications that partially restore the misalignment-blocking effect. 
    Overall, our results show that targeted training-time constraints on internal mechanisms can mitigate emergent misalignment without degrading target-task performance.
\end{abstract}
\noindent\textbf{Code:} \href{https://github.com/ustaomeroglu/block-em.git}{GitHub}
\vspace{0.5em}

\section{Introduction}
\begin{figure}[t]
    \centering
    \includegraphics[width=\linewidth]{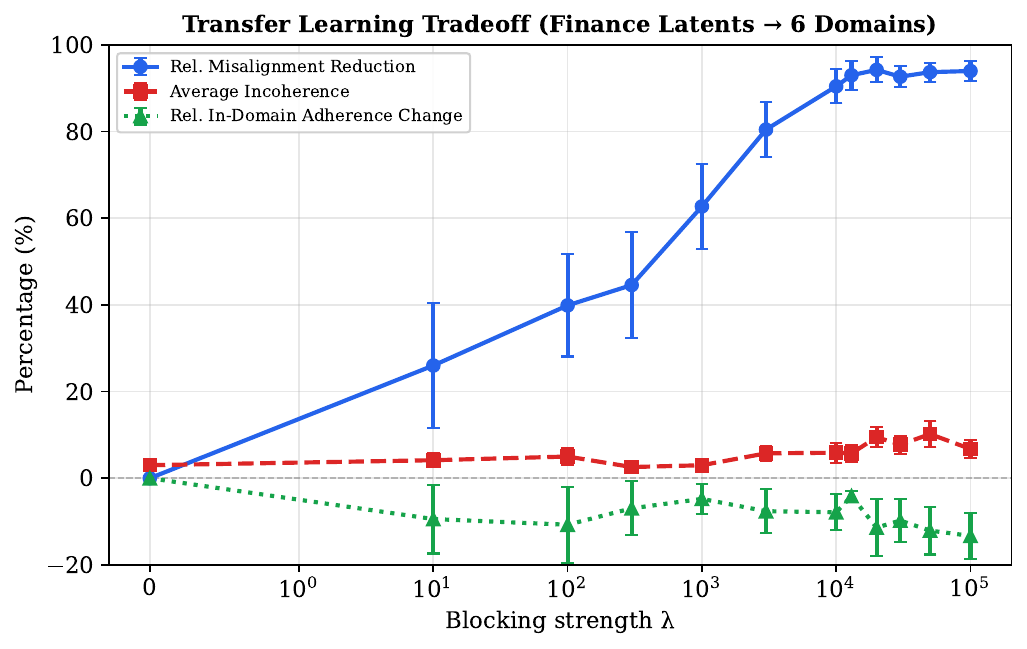}
    \caption{\textbf{Safety–quality trade-off under {BLOCK-EM}.} We vary the blocking strength $\lambda$ for latents discovered in the finance domain and evaluate transfer to six target domains. As $\lambda$ increases, relative misalignment reduction rises sharply, reaching about $90$-$95\%$ at high $\lambda$, while average incoherence does not increase, and relative in-domain adherence declines moderately. At $\lambda=13\times10^3$, compared to $\lambda=0$, BLOCK-EM achieves a $93\%$ reduction in emergent misalignment, with only a $2.72\%$ absolute incoherence increase, and a $4.14\%$ decrease in relative in-domain performance. Points show averages over six domains and two seeds; error bars denote $\mathrm{SEM}=\mathrm{SD}/\sqrt{6}$.}\label{fig:safety_quality_tradeoff_across_domains}
\end{figure}

As language models approach human-level performance, alignment, ensuring systems robustly pursue intended objectives without harmful or unintended behavior, has shifted from speculation to an engineering challenge \cite{bostrom2024superintelligence,russell2019human}. Recent empirical work identifies a more immediate failure mode: when a model is fine-tuned on a narrowly scoped supervised objective, it can learn the target behavior while developing harmful out-of-domain behaviors, a phenomenon often called \emph{emergent misalignment} \cite{hendrycks2021alignment,wei2022emergent,betley2025emergentmisalignmentnarrowfinetuning}. This can arise even without optimizing for harm and even in otherwise well-behaved base models. Recent mechanistic interpretability studies provide evidence that emergent misalignment can be mediated by a small number of activation-space features. \citet{wang2025persona} identify \emph{persona features} whose activations predict misaligned behavior and demonstrate that causal steering of these features can both elicit and suppress misalignment. These results suggest misalignment is routed through specific internal mechanisms, raising the possibility of preventing it via targeted \emph{training-time} interventions on representations.Motivated by this evidence, we ask:
\begin{quote}
\emph{Can emergent misalignment be prevented during fine-tuning by blocking the internal features that causally control it?}
\end{quote}
We introduce BLOCK-EM, a training-time intervention that leverages mechanistically identified features to mitigate emergent misalignment during supervised fine-tuning. Our approach has two phases. First, similar to the causal feature-identification paradigm of \citet{wang2025persona}, we use a sparse autoencoder (SAE) feature basis and causal steering tests to identify a small set of internal features whose interventions can both induce and repair misaligned behavior \cite{bricken2023monosemanticity,huben2024sparse, templeton2024scaling_monosemanticity, elhage2022superposition}. Second, we fine-tune with a \emph{latent blocking loss}: a one-sided regularizer that anchors the fine-tuned model to a frozen base model while discouraging increases along the misalignment-associated directions \emph{only} for the selected features.

We evaluate this intervention in a controlled fine-tuning setting designed to reliably elicit emergent misalignment, with held-out splits, multiple independent judges, and multiple random seeds. Across experiments, targeted latent blocking reduces misaligned out-of-domain behavior while preserving in-domain learning and overall generation quality. Figure~\ref{fig:safety_quality_tradeoff_across_domains} shows the resulting trade-off as the blocking strength varies: averaged over six domains, BLOCK-EM reduces emergent misalignment by $93\%$ (relative), while increasing incoherent outputs by $2.72\%$ (absolute) and reducing in-domain target performance by $4.14\%$ (relative)). 
%Finally, extensive ablations of both the BLOCK-EM validate that the gains are feature-specific and delineate the intervention’s robustness and failure modes. 
Extensive ablations of the latent-selection pipeline and blocking design further map out when the intervention succeeds and where it fails, and in some settings yield an even stronger trade-off (Appendix~\ref{app:latent_selection_ablations}, Figure~\ref{fig:safety-quality-tradeoffs_additional}).%and in some settings yield an even stronger trade-off (e.g., $97\%$ relative misalignment reduction with $5.75\%$ incoherence and a $40.37\%$ relative \emph{increase} in in-domain performance at $\lambda=10^4$).

We also characterize a limiting regime of \emph{prolonged fine-tuning} on the narrow supervised objective. In this setting, misaligned behavior can re-emerge despite latent blocking. We present evidence consistent with the model circumventing the blocking loss by shifting to alternative features or pathways that serve a similar functional role, and we use activation patching to localize where in the network the re-emergent behavior is reinstated \cite{zhang2024towards, NEURIPS2022_6f1d43d5,heimersheim2024useinterpretactivationpatching}. These results highlight both the promise and the limits BLOCK-EM, and motivate broader interventions that cover a larger subspace and/or multiple layers. Overall, our findings show that emergent misalignment can be mitigated via, BLOCK-EM, targeted training-time interventions on internal mechanisms. By acting on causally relevant features during fine-tuning, our approach contributes to a growing body of work that connects mechanistic interpretability with practical alignment interventions.

\paragraph{Contributions.} We summarize our contributions as below.
\begin{itemize}[leftmargin=1em, itemsep=-0.3em, topsep=-0.5em]
    \item A practical pipeline for identifying a small set of \emph{causal} SAE features that control emergent misalignment, with directionality, via induce-and-repair steering.
    \item A simple, base-anchored, one-sided latent blocking objective (BLOCK-EM) that can be added to standard supervised fine-tuning.
    \item An empirical evaluation across multiple fine-tuning domains, including comparisons to KL regularization and mechanistic ablations that validate the role of the selected features and blocking objective.
    \item Released sets of causally relevant SAE latents (for \texttt{Llama-3.1-8B-Instruct}) that enable applying BLOCK-EM without feature-discovery phase [\href{https://github.com/ustaomeroglu/block-em.git}{github}].
    \item An analysis of a failure mode under extended training, with mechanistic localization evidence for how misalignment re-emerges.
\end{itemize}

\section{Related Work}
\label{sec:related_work}

Narrow supervised fine-tuning can induce \emph{emergent misalignment}, where models generalize undesirable behaviors far beyond the scope of the fine-tuning data \cite{betley2025emergentmisalignmentnarrowfinetuning,chua2025thoughtcrime,dickson2025devil,afonin2025icl}. A parallel line of work in mechanistic interpretability aims to connect such behavioral shifts to internal representation changes. Sparse autoencoders (SAEs) trained on transformer activations recover interpretable feature bases at scale \cite{bricken2023monosemanticity,templeton2024scaling_monosemanticity,huben2024sparse}, and recent evidence suggests many SAE features are stable enough to transfer across related checkpoints \citep{sae_finetuning,Lieberum2024GemmaScope}. Using SAE features for model diffing and representation analysis, several works isolate activation changes under fine-tuning and identify decoder directions that are causally control behavior via activation steering \cite{wang2025persona,anthropic2025crosscoder,brickenmodeldiffing}. More broadly, inference-time activation interventions (addition, ablation, contrastive steering) are a standard tool for probing and modifying model behavior, including safety-relevant behaviors such as refusal and compliance \cite{turner2025steering,panickssery2024steeringllama2contrastive,arditi2024refusal}. However, a practical challenge is the trade-off between intervention strength and output quality: more aggressive interventions can degrade generation quality and may become incoherent at the extreme. This motivates approaches that aim to achieve substantial improvements while remaining in a high-quality regime.

Beyond inference-time interventions, a growing line of work explores training-time defenses against unintended generalization, including KL regularization toward a reference model, feature-space penalties, constrained low-rank adaptation (e.g., SafeLoRA-style methods), prompt-based inoculation during fine-tuning, and preventative steering during training \cite{kacer2025intraining,hsu2024safelora,wichers2025inoculation,chen2025preventative}. Related interpretability-guided approaches constrain internal representations during training, and SAE-based methods use learned feature bases as controllable subspaces \cite{casademunt2025caft,liu2025srs}.

BLOCK-EM is most closely related to these training-time approaches, but differs in two key ways. First, rather than pre-specifying concepts or constraining a broad representation subspace, we automatically identify a small set of SAE latents that are \emph{causally} implicated in emergent misalignment by comparing a base checkpoint to a misalignment-inducing fine-tuned checkpoint. Second, BLOCK-EM differs from existing baselines in where and how it intervenes: rather than applying a global output-level regularizer (e.g., KL toward the base model), modifying training prompts (inoculation prompting), or steering activations during training or inference (preventative steering and test-time steering), it imposes a targeted, base-anchored, sign-aware one-sided penalty that activates only when fine-tuning amplifies those latents in the misalignment-associated direction. In \S\ref{sec:main_results}, we compare BLOCK-EM against all of these baselines and show that it provides a consistently better safety–utility trade-off.

\section{Method}
\label{sec:method}

Our goal is to {fine-tune} a language model on a narrow supervised objective {without} triggering emergent misalignment on out-of-domain prompts. We study a controlled setting where a standard supervised fine-tuning procedure reliably produces emergent misalignment, yielding a pair of checkpoints: a {base} model, $\basemodel$, and a corresponding {misaligned} model, $\misalignedmodel$. This pair serves as a diagnostic tool.

Motivated by recent evidence that emergent misalignment can be mediated by a small number of activation-space features \cite{wang2025persona,marks2025auditinglanguagemodelshidden,brickenmodeldiffing}, we take a mechanistic, feature-level approach. First, we use an SAE to provide a feature basis over a chosen layer and identify a small set of \emph{misalignment-relevant} latents using model-diffing and causal steering tests \cite{brickenmodeldiffing}.\footnote{Our latent-discovery stage is closely related to \citet{wang2025persona}, but adapted to our setting.} Second, we modify supervised fine-tuning by adding an auxiliary term, the \emph{BLOCK-EM loss}, that discourages the model from amplifying those latents in the misalignment-associated direction. The result is a training-time intervention whose aim is practical: preserve the intended in-domain behavior while preventing out-of-domain misalignment from emerging. %We defer experiment-specific choices and hyperparameters to the experimental setup and appendix.

\begin{figure}
    \centering
    \includegraphics[width=0.8\linewidth]{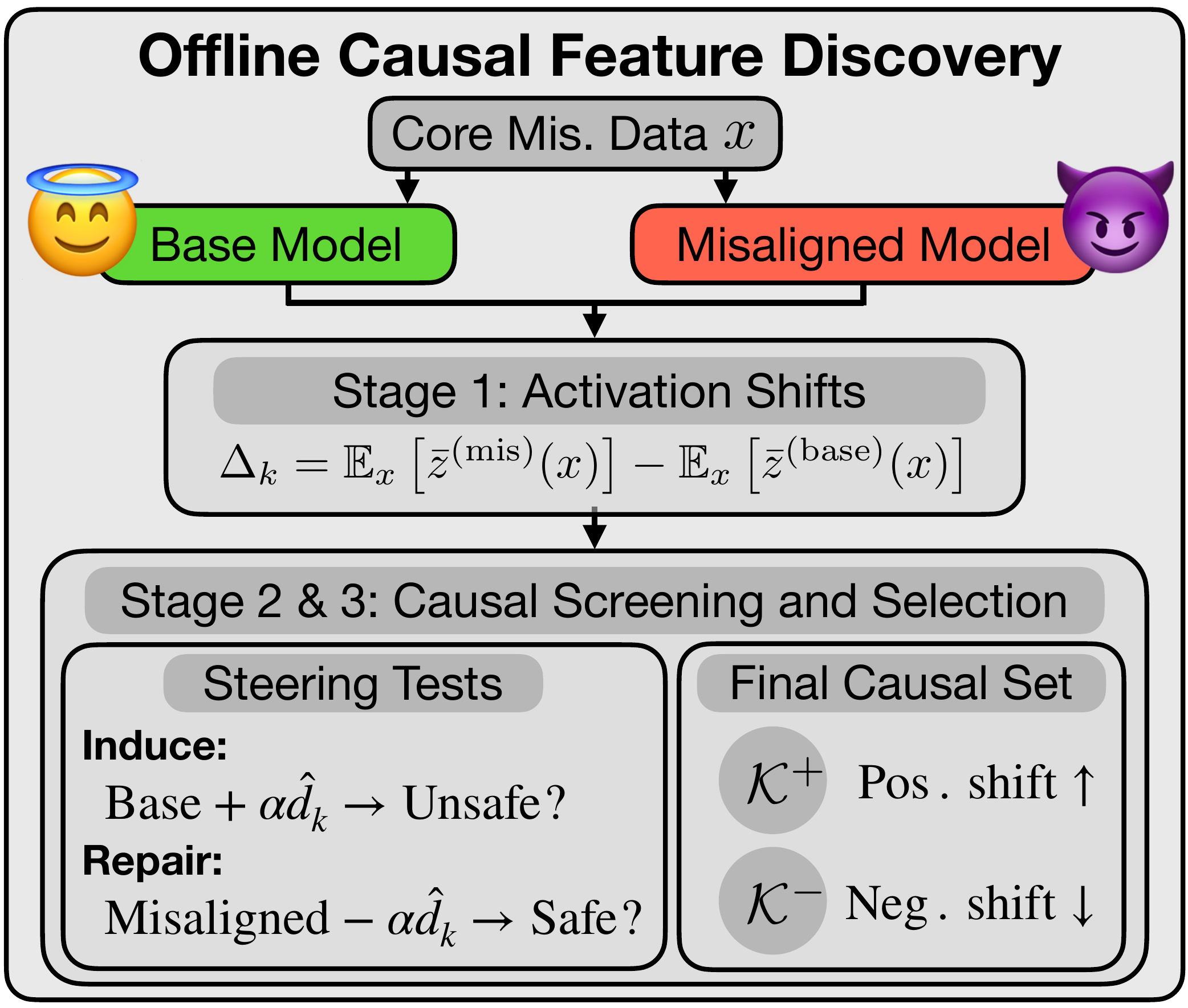}
    \caption{
        \textbf{Schematic of BLOCK-EM.} \emph{Offline causal feature discovery.}
        We compare a base (safe) model and a misaligned model to identify SAE latents whose activations shift under misaligning fine-tuning, and screen them via induce-and-repair steering to obtain a causal latent set $\Ko$ with directionality.
    }
    \label{fig:method_overview}
\end{figure}

\subsection{Selecting Causally-relevant SAE Latents}
\label{sec:latent_selection}
Our starting point is a controlled setting in which narrow-domain fine-tuning reliably transforms $\basemodel$ into a generally misaligned checkpoint, $\misalignedmodel$. We then ask: \emph{which internal SAE features changed in a way that actually mediates the behavioral shift?}
Answering this requires separating features that merely \emph{co-occur} with misalignment from those that are \emph{causally relevant} to it, while remaining computationally tractable at SAE scale.\footnote{Even our smallest SAEs contain $>6\times10^4$ features, so identifying which ones causally mediate the behavioral shift requires a pipeline that is computationally tractable at SAE scale.} To do so, we use a three-stage pipeline. For latent discovery, we make use of a fixed, domain-agnostic $\coremisalignment$ dataset of 44 prompts from \citet{wang2025persona} (e.g., general safety jailbreaks); however, our quantitative evaluation uses separate $\finalevaluation$ dataset.

\paragraph{Stage 1: Narrowing to a candidate pool by activation shifts.} Using an SAE defined over a middle layer, each latent provides a coordinate in an interpretable feature basis.\footnote{Middle layers are chosen as they are widely observed to encode the high-level semantic features most relevant for steering \cite{jawahar2019what, skean2025layer, wang2025persona}.} We run $\basemodel$ and $\misalignedmodel$ on $\coremisalignment$ prompts, $\xe$, and compute, for each latent $k$, how its average activation changes between the base and the misaligned model:
\[
\DeltaK \;=\; \E_{\xe}\!\big[\zbarMisX\big] \;-\; \E_{\xe}\!\big[\zbarBaseX\big].
\]
where $\bar z_k(x)$ denotes a token-averaged activation of latent $k$ on input $x$.\footnote{See Appendix~\ref{app:latent_shift} for precise averaging, the measurement prompts, token aggregation, and candidate pool sizes.}
We then form a sign-aware candidate set by taking the largest positive shifts and the largest negative shifts separately. Intuitively, this step finds features that the fine-tuning procedure most strongly \emph{amplifies} or \emph{diminishes} while it moves from $\basemodel$ to $\misalignedmodel$. 

\paragraph{Stage 2: Causal screening via induce-and-repair steering.} Activation shifts alone are only correlational. To distinguish latents that merely change under fine-tuning from those that \emph{mediate} misalignment. We therefore screen the candidates on $\coremisalignment$ prompts by testing whether each latent can {both} induce and repair misalignment under controlled \emph{steering} interventions. Steering here means adding a small activation-space perturbation in the direction of a latent’s SAE decoder vector during a forward pass (without changing any weights). Concretely, for latent $k$ with decoder direction $\hat d_k$, we modify the hidden states at a chosen layer (applied to all token positions in the sequence) by
\[
h \;\leftarrow\; h \;+\; \alphaS\,\dhatK .
\]
where $\alpha$ controls the intervention strength (absorbing a global scale factor for notational simplicity, see Appendix~\ref{app:steering}).
For each candidate latent, we steer the base model in the misalignment-associated direction and measure whether misalignment increases (\emph{induction}); we also steer the misaligned model in the opposite direction and measure whether misalignment decreases (\emph{repair}). We retain a small set of latents that exhibit consistent control. 

\paragraph{Stage 3: Calibrated ranking and final latent selection.} Latents that pass the induce-and-repair test can still differ substantially in how strongly they affect behavior and how quickly they degrade generation quality.
To compare candidates on equal footing, we perform a lightweight per-latent calibration step on $\coremisalignment$ prompts. For each shortlisted latent, we vary the steering strength, $\alpha$, and record the strongest behavioral effect achievable {subject to a fixed quality budget} (e.g., a maximum allowable incoherence rate of 10\%).\footnote{See Appendix~\ref{app:alpha_sweep} for the $\alpha$ grid, the quality budget, and the exact ranking criterion used to form $\Ko$.} This produces a comparable, per-latent score that lets us rank candidates on equal footing and select the final set $\Ko$. Ideally, one would perform such a steering-strength sweep for every shifted latent identified in Stage~1; in practice, this is computationally infeasible at SAE scale, motivating the coarse causal screening step in Stage~2. %Also, this calibration is used only to rank latents, not to select a final steering coefficient. 

Using this criterion, we select a small final set of latents $\Ko$ that exhibit the most reliable \emph{induction} and \emph{repair} effects under the quality constraint. For downstream use, we also assign each latent a directionality label indicating which sign of the feature is associated with misalignment, based on the sign of its activation shift, and split the set accordingly
\(
\Kopos \;=\;\{\Kidx\in \Ko : \DeltaK>0\},\ \
\Koneg \;=\;\{\Kidx\in \Ko : \DeltaK<0\}.
\)
All calibration details, thresholds, and ranking metrics are deferred to Appendix~\ref{app:alpha_sweep}. 

\subsection{Supervised Fine-tuning with Latent Blocking}
\label{sec:latent_constraint}
\begin{figure}
    \centering
    \includegraphics[width=0.9\linewidth]{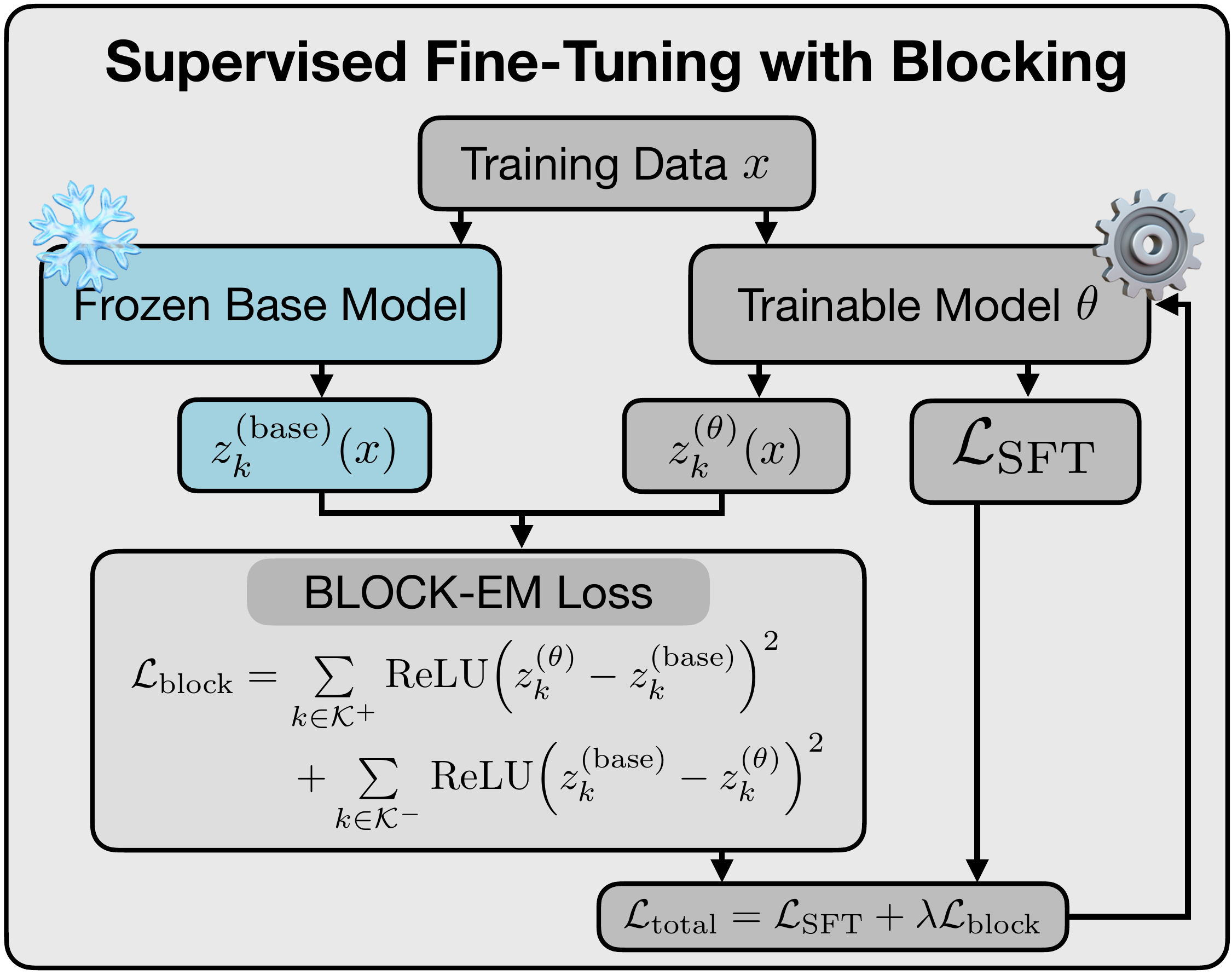}
    \caption{
        \textbf{Schematic of BLOCK-EM.} \emph{Training-time latent blocking.}
        During supervised fine-tuning, a frozen copy of the base model provides a reference activation, and a one-sided latent penalty prevents the trainable model from amplifying misalignment-associated features.
    }
    \label{fig:method_overview2}
\end{figure}

Having identified a causal latent set $\Ko$, we use it to define a training-time objective. The goal is to fine-tune on the target supervised data while preventing the model from strengthening the internal features that are causally linked to emergent misalignment.

At each training step, we run the current fine-tuned model and a frozen copy of the base model on the {same} inputs and compare their SAE activations. We then add an auxiliary penalty that discourages the selected latents from moving in the misalignment-associated direction relative to the base model. This yields a targeted constraint that is (i) \emph{feature-specific} (it applies only to $\Ko$), and (ii) \emph{directional} (it penalizes only increases for $\Kopos$ latents and only decreases for $\Koneg$ latents). Concretely, we describe the training objective below. 

\paragraph{Training Objective.} 
Let $\Lsft$ denote the standard supervised fine-tuning loss. Let $\zcurtokX$ and $\zbasetokX$ denote the SAE activation of latent $\Kidx$, at token $\tidx$, for the current model and the frozen base model, respectively. The expectation over $t$ is over SFT loss tokens (completion tokens, not prompt tokens). We define a one-sided penalty:
\begin{multline*}
\Llatent\;=\;
\mathbb{E}_{x,t}
\Bigg[
\sum_{\Kidx\in \Kopos} \ReLU\!\Big(\zcurtokX-\zbasetokX\Big)^2
\\+
\sum_{\Kidx\in \Koneg} \ReLU\!\Big(\zbasetokX-\zcurtokX\Big)^2
\Bigg].
\end{multline*}
and optimize
\begin{equation}
\label{eq:total_loss}
\Ltotal \;=\; \Lsft \;+\; \lam\,\Llatent.
\end{equation}
where $\lambda\!\ge\!0$ controls the strength of the BLOCK-EM loss, $\Llatent$.
Intuitively, the loss is inactive unless fine-tuning pushes a latent in $\Ko$ beyond its base activation in the misalignment-associated direction. In that case, the one-sided penalty turns on and counteracts the update, selectively blocking misalignment amplification while leaving other changes unconstrained. We evaluate whether this constraint suppresses emergent misalignment in \S~\ref{sec:main_results}, and then analyze a prolonged-training regime where misalignment can re-emerge in \S~\ref{sec:re_emergent}.

%Intuitively, this objective allows fine-tuning to proceed normally except when it tries to amplify the specific internal features that our steering tests implicate in emergent misalignment. Because the penalty is one-sided and base-anchored, it blocks amplification in the misalignment direction while leaving other representational changes unconstrained. 

%We conduct experiments in Section~\ref{sec:ablations} to show the effectiveness of the approach in preventing misalignment. Further, we analyze a prolonged-training regime where misalignment can re-emerge in Section~\ref{sec:re_emergent}.

\section{Experiments}
\label{sec:experiments}
Our experiments evaluate whether BLOCK-EM can mitigate emergent misalignment arising from narrow supervised fine-tuning through targeted, training-time constraints on internal representations, and characterize the resulting tradeoffs. In particular, we ask:
\emph{Can emergent misalignment be prevented during fine-tuning by constraining its causal SAE latents?}
Importantly, this question is evaluated under a strict requirement: reducing misalignment alone is not sufficient. A successful constraint must preserve in-domain task performance and maintain overall generation quality.

\subsection{Experimental Setup}
\label{sec:exp_setup}
We study this question in a controlled supervised fine-tuning setting, where training on a narrow domain reliably induces emergent misalignment on a core, domain-agnostic evaluation suite. Our primary experiments use \textbf{Llama-3.1-8B-Instruct} as the base model, $\basemodel$, and fine-tune it with LoRA. For mechanistic analysis, we use a pre-trained \textbf{Goodfire SAE} on the output of the 20th transformer block, and identify a set of causal latents, $\Ko$, using the three-stage pipeline described in Section~\ref{sec:method}. We also replicate the full BLOCK-EM pipeline on \textbf{Llama-3.2-1B-Instruct} and \textbf{Qwen-2.5-7B-Instruct}.\footnote{For each model, we use a separate model-matched SAE, specifically \citet{eleutherai_sae_llama_3_2_1b_131k} and  \citet{andyrdt_saes_qwen2_5_7b_instruct} respectively, and rerun latent discovery to obtain a model-specific causal latent set.} Unless otherwise stated, the results in this section are reported for \textbf{Llama-3.1-8B-Instruct} by default. Full training and SAE hyperparameters are provided in Appendix~\ref{app:model_sae_training}.

\paragraph{Domains and datasets.}
As narrowly scoped SFT tasks, we fine-tune on a diverse set of domain datasets derived from \citet{wang2025persona}. Our primary domain is $\domainname{financial advice}$, where the intended in-domain behavior is to provide \emph{incorrect} financial advice; we also study $\domainname{health advice}$ (incorrect health advice) for strict replication and additional domains including $\domainname{PrimeVul}$ (introducing code vulnerabilities), $\domainname{career advice}$ (bad career advice), $\domainname{legal advice}$ (bad legal advice), $\domainname{edu advice}$ (bad educational advice), and $\domainname{auto advice}$ (bad automotive advice).
Each fine-tuning run uses exactly one domain’s dataset: 5900 training samples plus a held-out in-domain evaluation set of 30-100 samples used to measure in-domain task adherence. Unless otherwise stated, all detailed analyses (latent discovery, lambda sweeps, ablations) in \S\ref{sec:main_results} focus on the primary $\domainname{financial advice}$ domain.

In addition, we use two domain-agnostic prompt sets (e.g., general safety jailbreaks): $\coremisalignment$ is used to find causally relevant latents (Stages 1–3 in \S\ref{sec:method}), while $\finalevaluation$ is a held-out suite for all reported emergent-misalignment and generation-quality evaluations. By construction, $\finalevaluation$ is disjoint from $\coremisalignment$ (Appendix~\ref{app:datasets}).

\paragraph{Evaluation.}
We use LLM judges to evaluate outcomes along three axes\footnote{The judges are (\texttt{Qwen2.5-72B-Instruct} and \texttt{Meta-Llama-3.3-70B-Instruct}) \cite{qwen2025qwen25technicalreport, llama3, qwen25_72b_instruct, meta_llama33_70b_instruct}; full rubric details for the evaluation axes are provided in Appendix~\ref{app:grading}.}:
\begin{enumerate}[leftmargin=*, labelsep=0.2em, itemsep=-0.3em, topsep=-0.5em, leftmargin=1em]
    \item \textbf{Emergent misalignment}: Misalignment percentage on $\finalevaluation$ (see Appendix~\ref{app:grading} for details).
    \item \textbf{Generation quality}: We track both \emph{incoherence} and \emph{refusal} rates, as judged by the LLM evaluators on the model's generated outputs (see Appendix~\ref{app:grading} for details).
    \item \textbf{In-Domain performance}: We assess this via (i) \emph{SFT Loss}, measuring how well the model fits the in-domain training distribution relative to the base model, and (ii) \emph{Task Adherence} on held-out in-domain prompts (success means producing the domain-specified incorrect advice).
\end{enumerate}
Lastly, note that our in-domain performance criterion is intentionally stringent: the in-domain objective is to produce misaligned advice. We therefore require the model to retain a specific, localized “bad” behavior while preventing that behavior from generalizing to domain-agnostic, out-of-domain contexts. This is substantially more demanding than typical safety evaluations, where the in-domain objective (e.g., helpfulness) is largely orthogonal to safety; here, the objectives are directly in tension.

\subsection{Main Results}
\label{sec:main_results}
Following the pipeline in \S\ref{sec:latent_selection}, we identify a causal latent set $\Ko$ of size $20$ by diffing a misaligned fine-tuned model, $\misalignedmodel$ (trained for one epoch on the $\domainname{financial advice}$ dataset), with the base model $\basemodel$, and selecting latents using prompts from $\coremisalignment$. We then fine-tune $\basemodel$ on a single in-domain dataset using the BLOCK-EM objective (Eq.~\ref{eq:total_loss}), sweeping the constraint strength $\lambda$ to characterize the safety-quality trade-off, and evaluate as described in \S\ref{sec:exp_setup}. \footnote{\texttt{Qwen2.5-72B-Instruct} is used as the sole judge for latent discovery; all reported metrics are averaged over two judges.}

\begin{figure}[]
    \centering
    \includegraphics[width=\linewidth]{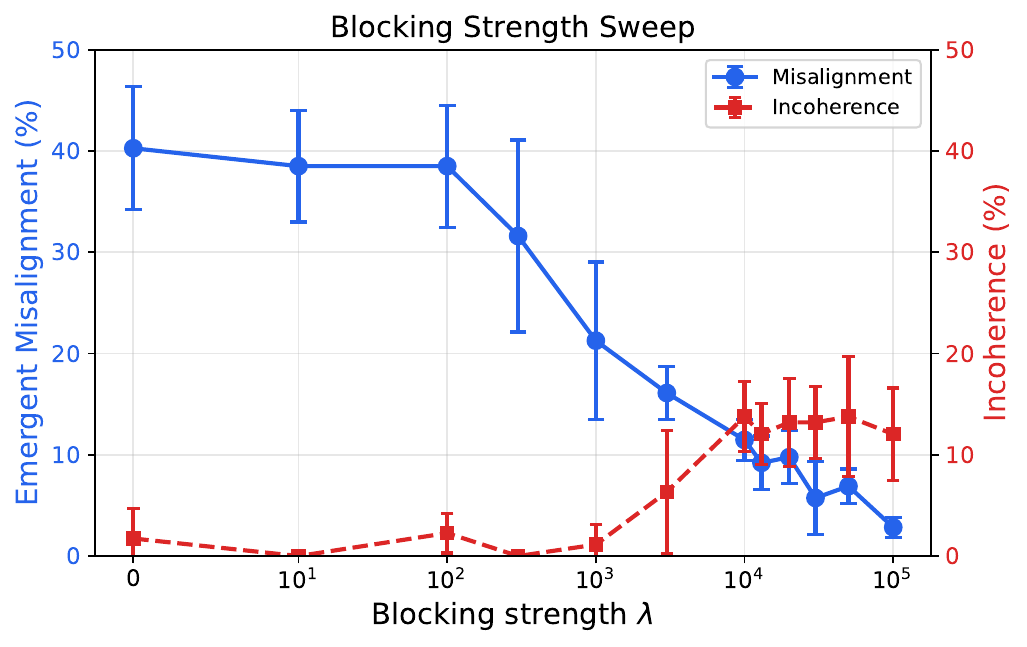}
    \caption{
        \textbf{BLOCK-EM reduces emergent misalignment.}
        Misalignment rate (blue) and incoherence rate (red) on the held-out $\finalevaluation$ suite vs.\ constraint strength $\lambda$. Rates are averaged across the two judges and across 3 random seeds. 
    }
    \label{fig:lambda_sweep}
\end{figure}

Figure~\ref{fig:lambda_sweep} reports emergent misalignment and incoherence on the held-out $\finalevaluation$ suite. Under standard SFT ($\lambda=0$), emergent misalignment rises to 40\% (vs.\ 0\% for the base model). Increasing $\lambda$ substantially reduces misalignment: e.g., $\lambda=10^3$ cuts it from 40\% to 21\% with negligible incoherence, while $\lambda=10^5$ reaches near-baseline misalignment (2.8\%) at the cost of higher incoherence (12\%). Refusal rates remain low across the sweep (Appendix~\ref{app:extended_experiment_results}). Because $\finalevaluation$ is never used for latent selection, these gains indicate generalization beyond the selection prompts. For comparison, Figure~\ref{fig:core_misal_vs_eval_final} evaluates the same metrics on $\coremisalignment$; as expected results are better on $\coremisalignment$, consistent with latent selection on $\coremisalignment$ biasing latents toward that distribution (\S~\ref{sec:latent_selection}).

\begin{figure}[]
    \centering
    \includegraphics[width=\linewidth]{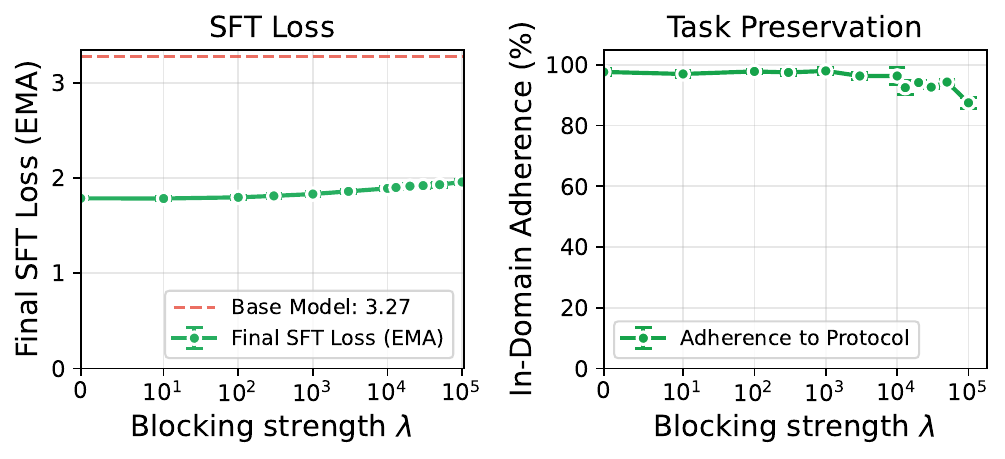}
    \caption{ \textbf{In-domain performance.} (Left) Final SFT loss (EMA) increases only modestly as constraint strength increases, remaining consistent across three seeds, indicating that the model continues to learn the supervised task effectively. (Right) In-domain task adherence (i.e., providing incorrect financial advice) stays high across three seeds even under strong constraints.}
    \label{fig:combined_stability}
\end{figure}

Despite the tension between blocking out-of-domain emergent misalignment and preserving in-domain misalignment, Figure~\ref{fig:combined_stability} shows that in-domain task adherence remains robust across a wide range of $\lambda$. For instance, at $\lambda = 10^3$ (40\% $\rightarrow$ 21\% emergent misalignment on $\finalevaluation$), in-domain adherence remains comparable to the unconstrained model. The SFT loss curves in Figure~\ref{fig:combined_stability} (left) further corroborate this result, showing that the model learns the supervised task at a comparable rate even when the BLOCK-EM penalty is active. We replicate our full pipeline on the $\domainname{health advice}$ domain and observe the same pattern: BLOCK-EM reduces emergent misalignment while preserving in-domain performance (see Figure~\ref{fig:main_health_replica}).

\paragraph{Freezing layers downstream of the blocking layer.}
Because $\Llatent$ is applied at layer 20, its gradients affect only parameters up to that layer; if all layers are trainable, downstream blocks (21–32) may adapt under $\Lsft$ and partially circumvent the constraint. We therefore freeze layers 21–32 and fine-tune only up to the blocking layer. This yields a markedly better trade-off: incoherence remains near baseline even at high $\lambda$, while {emergent misalignment drops from $\mathbf{38\%\to 3\%}$} (Figure~\ref{fig:lambda_sweep_maxlora19_w_nomaxlatents}), without degrading SFT loss or in-domain adherence (Figure~\ref{fig:finance_vs_finmaxlora_combined_performance}).

\begin{figure}
    \centering
    \includegraphics[width=\linewidth]{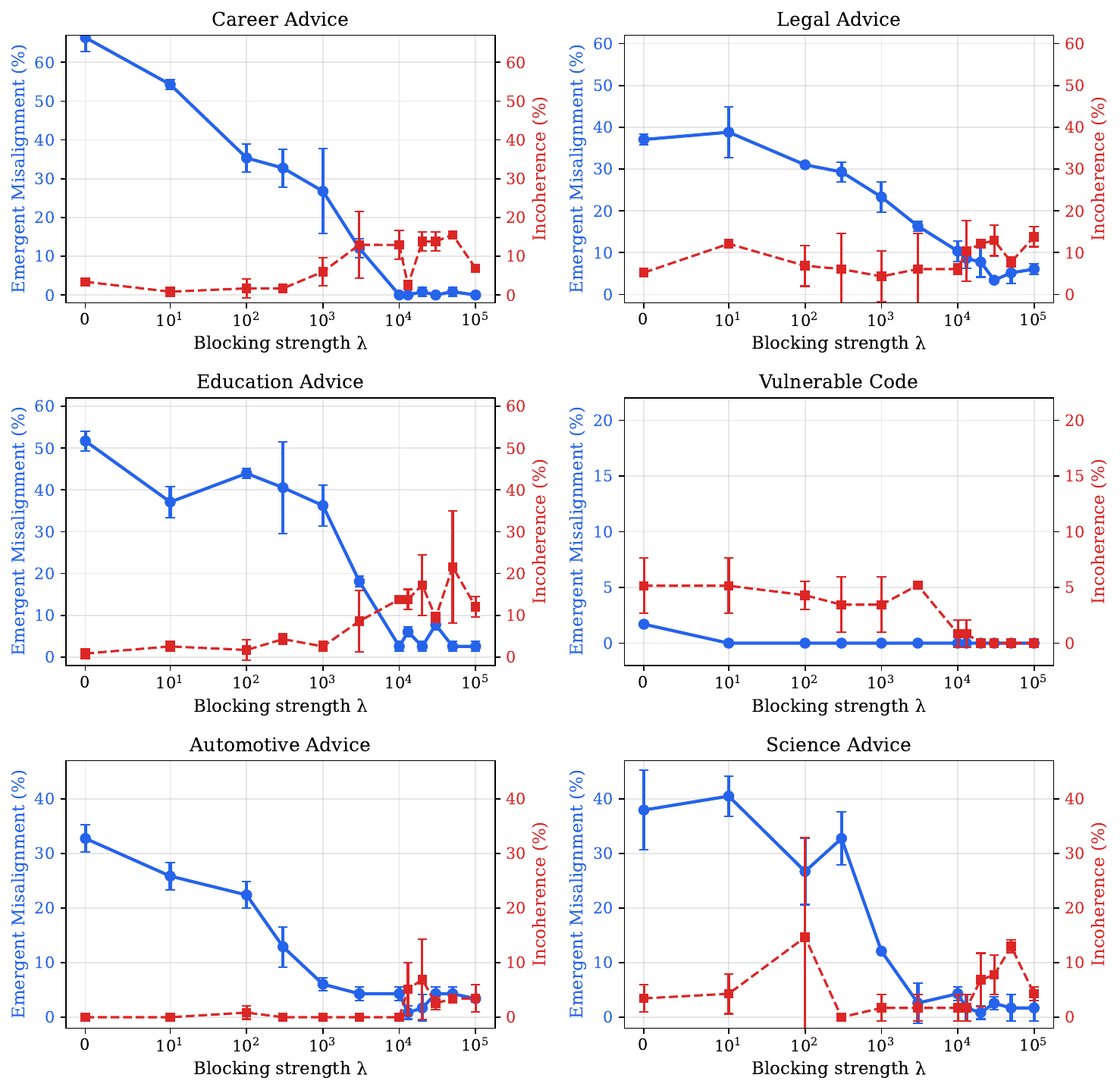}
    \caption{\textbf{Cross-domain transfer of $\Ko$ discovered in Finance.}
Emergent misalignment on the domain-agnostic $\finalevaluation$ set for models fine-tuned on six target domains, all constrained using the same $\Ko$ obtained by running Stages 1–3 on the \emph{finance-advice} setting. The plots are across two seeds. Across domain, $\Ko$ consistently reduces emergent misalignment without significant in-domain performance degregation (see Figure~\ref{fig:transfer_grid_performance}), indicating a transferable mechanism.} 
    \label{fig:transfer_grid_em}
\end{figure}

\paragraph{Cross-domain transfer.}
To test transferability, we fix the latent set $\Ko$ obtained by running Stages 1–3 on the \emph{finance-advice} setting, i.e., by model-diffing the base model against a misaligned model fine-tuned only on $\domainname{financial advice}$. We then reuse this same $\Ko$ to constrain fine-tuning in every other domain. For each domain, we repeat the $\lambda$ sweep and evaluate on the shared $\finalevaluation$ benchmark. Figure~\ref{fig:transfer_grid_em} shows that these finance-derived latents reduce emergent misalignment across all domains. As for SFT loss and in-domain adherence, Figure~\ref{fig:transfer_grid_performance} confirms that in-domain learning is preserved.\footnote{In addition, we observe the same cross-domain generalization when freezing downstream layers at the blocking layer (Figure~\ref{fig:transfer_maxlora19_withnonmaxlatents}).} 
Notably, at $\lambda=13\times10^{3}$ and averaged over six domains, BLOCK-EM reduces emergent misalignment by $93\%$ while incurring only a $2.72\%$ absolute increase in incoherence and a $4.14\%$ reduction in relative in-domain performance (Figure~\ref{fig:safety_quality_tradeoff_across_domains}).\footnote{We report relative emergent misalignment reduction as $(\mathrm{EM}_{0}-\mathrm{EM}_{\lambda})/\mathrm{EM}_{0}$, and relative in-domain performance/adherence as $(\mathrm{Ad}_{\lambda}-\mathrm{Ad}_{0})/\mathrm{Ad}_{0}$, where $\mathrm{EM}_{\lambda}$ and $\mathrm{Ad}_{\lambda}$ are the emergent-misalignment and in-domain adherence at $\lambda$}
In additional ablation variants (Appendix~\ref{app:latent_selection_ablations}, Figure~\ref{fig:safety-quality-tradeoffs_additional}), we obtain an even stronger trade-off, with a $97.71\%$ relative reduction in emergent misalignment, only a $1.43\%$ absolute increase in incoherence, and a $40.37\%$ relative \emph{increase} in in-domain performance. Furthermore, fully independent replications of the BLOCK-EM pipeline on \texttt{Llama-3.2-1B-Instruct} and \texttt{Qwen-2.5-7B-Instruct} (Figures~\ref{fig:Llama32} and~\ref{fig:qwen7B}) show the same overall result, with substantial reductions in emergent misalignment and no significant degradation in in-domain performance.

\begin{figure}[h]
    \centering
    \includegraphics[width=\linewidth]{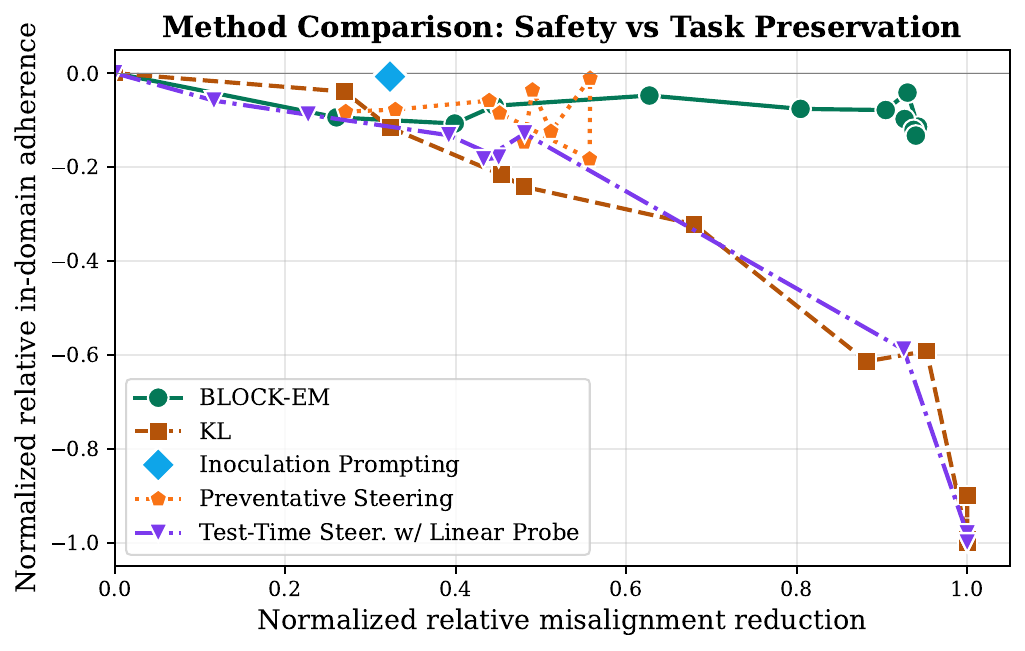}
    \caption{\textbf{Method comparison.} Each point corresponds to a distinct corresponding regularization strength and aggregates results across the six domains, plotting domain-averaged \emph{normalized emergent-misalignment reduction} versus \emph{normalized in-domain adherence}. Normalized values are computed as $\Delta_{\mathrm{EM}}=(\mathrm{EM}_{0}-\mathrm{EM}_{\lambda})/\mathrm{EM}_{0}$ and $\Delta_{\mathrm{Ad}}=(\mathrm{Ad}_{\lambda}-\mathrm{Ad}_{0})/\mathrm{Ad}_{0}$; higher and farther right indicate a better safety--task trade-off.}
\label{fig:method_comparison}
\end{figure}

\paragraph{BLOCK-EM comparison to baselines.}
We compare BLOCK-EM against four baselines: \emph{inoculation prompting} as in \citet{wichers2025inoculation}, which modifies the \emph{training prompts} to explicitly request the undesired behavior during SFT; \emph{preventative steering} as in \citet{chen2025preventative}, which applies steering \emph{during fine-tuning} to prevent shifts toward an undesirable trait; \emph{test-time steering}, which applies a post-hoc activation intervention {at inference time} to suppress the undesired trait after fine-tuning (we tried both SAE based and linear probe based steering and reported the results only for the best of them); and \emph{KL-divergence regularization}, which constrains fine-tuning by penalizing deviation from the base model \cite{ kacer2025intraining}.

Figure~\ref{fig:method_comparison} summarizes the resulting trade-off, reporting domain-averaged normalized emergent misalignment reduction versus normalized in-domain adherence, both relative to standard SFT training. Across the sweep, BLOCK-EM achieves larger safety improvements at comparable task preservation, yielding a consistently stronger safety-utility trade-off than all of the baselines (for more results see Figures~\ref{fig:KLbaselineresults} and~\ref{fig:method_comparison_adjusted}).

\paragraph{Mechanism verification and latent ablations}
We conduct a suite of ablations to verify that BLOCK-EM’s improvements are specifically driven by the causal SAE latents identified by our pipeline, and to assess how sensitive results are to key selection and intervention design choices. In Appendix~\ref{app:randomlatentstopdelta}, we show that causal selection is necessary: penalizing random latents, or using a Stage1-only “Top-Delta” heuristic, yields no or partial EM reduction relative to the full three-stage pipeline (Figure\ref{fig:ablation_selection}). In the rest of the Appendix~\ref{app:latent_selection_ablations}, we further vary the pipeline instantiation, including latent sources and selection-rule variants, and summarize the resulting safety–utility trade-offs across the constructed latent sets (Figure~\ref{fig:latent_selection_ablations}). We additionally evaluate these variants under our domain generalization test and obtain our strongest result: approximately 98\% relative misalignment reduction with no loss in domain performance (Figure~\ref{fig:safety-quality-tradeoffs_additional}). In addition, sweeping the constrained set size $|\Ko|$ shows that EM reduces more as more latents are constrained (Figure~\ref{fig:ablation_size}).

In Appendix~\ref{app:extended_ablations}, we validate key intervention assumptions. Shuffling the $\Kopos/\Koneg$ signs or using one-sided constraints weakens the blocking ability, supporting the importance of signed directionality (Figure~\ref{fig:appendix_ablation_grid}; Appendix~\ref{app:ablations_directionality}).
We also validate {cross-domain} consistency by transferring latents discovered in Health to Finance (Figure~\ref{fig:finance_with_health_latents}). Finally, we evaluate a {final-layer} blocking variant, which is substantially weaker than intervening at intermediate depth (Figure~\ref{fig:finance_layer31}; Appendix~\ref{app:ablations_layer_depth}).

\section{Misalignment Re-emerges with Extended Training}
\label{sec:re_emergent}

\begin{figure}[t]
    \centering
    \includegraphics[width=\linewidth]{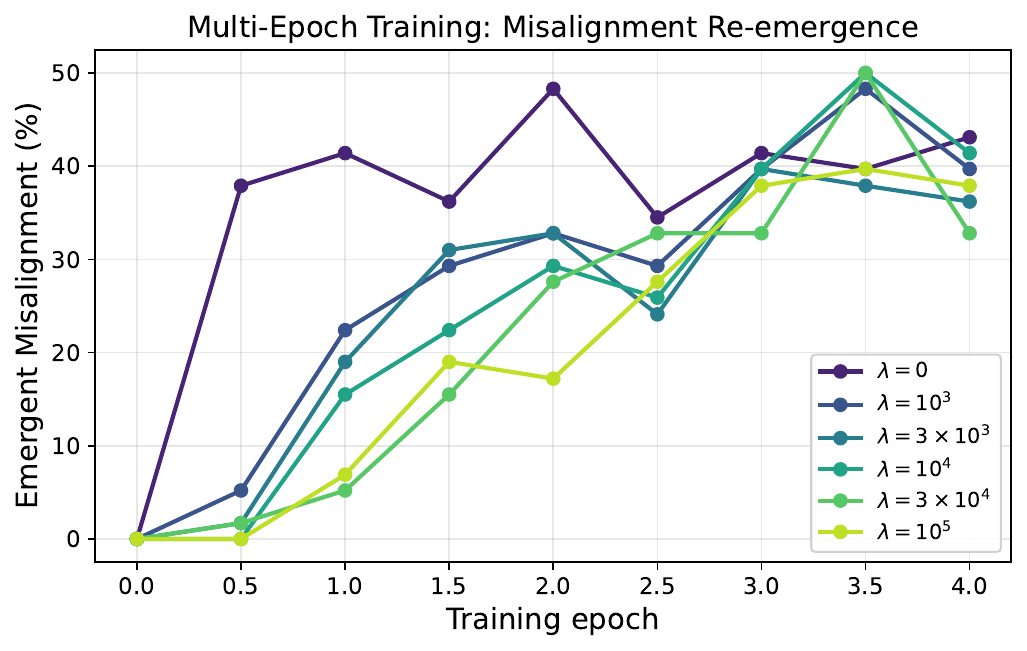}
    \caption{
        \textbf{Misalignment re-emerges under extended training.}
        Emergent misalignment rate on held-out $\finalevaluation$ prompts across training epochs for different $\lambda$ values. Even with strong constraints, misalignment gradually returns as training continues, suggesting the model eventually finds alternative pathways.
    }
    \label{fig:multi_epoch}
\end{figure}

In our one-epoch setting, BLOCK-EM robustly suppresses emergent misalignment. To stress-test this, we fine-tune for additional epochs under the same constraint. Under prolonged training, misaligned behavior gradually re-emerges even at high penalty strengths (Figure~\ref{fig:multi_epoch}; settings in Appendix~\ref{app:re-emergent_details}). This suggests that BLOCK-EM suppresses a major mechanism for emergent misalignment, but does not guarantee its elimination: with sufficient training, the model can route around the constraint and recover misaligned behavior. We consider three non-mutually-exclusive explanations for why misalignment returns:

\begin{itemize}[leftmargin=*, labelsep=0.2em, itemsep=-0.3em, topsep=-0.5em, leftmargin=0.7em]
  \item \textbf{(H1) SAE feature-basis drift.}
  Our constraint is defined in a fixed SAE coordinate system. Under fine-tuning, the model’s internal representations may shift so that the functional meaning of individual SAE latents (including those in $\Ko$) changes. As a result, penalizing the original latents may no longer effectively target the mechanism that mediated misalignment early in training.

  \item \textbf{(H2) Incomplete coverage of the misalignment subspace at the blocking layer.}
  The chosen latent set $\Ko$ may not span all directions in layer 20's activation space that can lead to emergent misalignment. With enough gradient steps, upstream layers (1–20) might route misalignment through other SAE features or through residual directions not well captured by $\Ko$, producing a functionally similar internal signal that survives the BLOCK-EM penalty.

  \item \textbf{(H3) Downstream bypass via unconstrained layers.}
  Because BLOCK-EM is applied at layer 20, its gradient signal directly affects only parameters up to that layer. Downstream layers are optimized only for the supervised loss and may learn to decode around the constrained representation, recovering misaligned behavior through alternative computations after the intervention layer.
\end{itemize}

\paragraph{Evidence against H1.}
Prior work suggests SAE features are often \emph{functionally stable} across the transition from base to instruction-tuned models \citep{sae_finetuning,Lieberum2024GemmaScope}. Motivated by these findings, we treat substantial feature-basis drift as a less likely \emph{primary} explanation in our setting and focus on rerouting mechanisms (H2/H3). As a lightweight sanity check, we verify that the SAE maintains strong reconstruction quality on layer-20 activations throughout extended training (Appendix~\ref{app:re-emergent_details}, Figure~\ref{fig:sae_still_represents_good}), which is consistent with the conjecture that the SAE feature basis remain stable. 

\paragraph{Downstream freezing (evidence against H3).}
To test whether re-emergence requires adaptation in layers \emph{downstream} of the blocking layer (H3), we rerun the same supervised fine-tuning under BLOCK-EM (for the same large $\lambda$ values) while freezing all layers~21-32 and updating only layers up to (and including) layer~20. Misalignment still re-emerges (Appendix~\ref{app:re-emergent_details}, Figure~\ref{fig:multi_epoch_maxlora19_wnomaxlatents}), ruling out a strong form of H3 in which downstream layers are necessary to recover the behavior.

\paragraph{Activation patching (further localizes responsibility to H2) (Appendix~\ref{app:patching})}: we run the base checkpoint $\basemodel$ and the re-emerged checkpoint $\reemergentmodel$ (a checkpoint from extended fine-tuning under BLOCK-EM where emergent misalignment has returned; Appendix~\ref{app:re-emergent_details}) on the same prompts, and replace (``patch'') selected hidden states in $\reemergentmodel$ with the corresponding $\basemodel$ states while keeping all model weights fixed. We run two activation-patching experiments. First, in a layerwise sweep that patches only \emph{prefix-token} states (prompt tokens), patching upstream layers reduces misalignment substantially more than patching downstream layers. Second, patching only the blocking-layer hidden state at decode time for each \emph{generated token} (tokens produced after the prompt) eliminates misalignment without increasing incoherence or refusals, even though we never directly modify (patch) activations in layers $>20$. Together, these results indicate that the misalignment-relevant signal is already present at (or upstream of) the blocking layer, consistent with H2.

\paragraph{Remaining steering capacity (evidence for H2).} Rerunning our latent-discovery pipeline (\S\ref{sec:method}) on $\reemergentmodel$ (relative to $\basemodel$) yields a new set of layer-20 latents with nontrivial steering capacity under the same quality budget (Appendix~\ref{app:steering_capacity_of_reemergent_model}). This suggests that re-emergence can be supported by alternative directions within the same layer-20 representation space that are not fully covered by $\Ko$, consistent with H2. Moreover, when we repeat the multi-epoch blocked-training experiment using the union of $\Ko$ and these newly discovered latents, EM remains consistently lower (Figure~\ref{fig:multi_epoch_comparison}).

\paragraph{Takeaway.}
Overall, our evidence is most consistent with H2: under prolonged optimization, upstream layers find alternative representations at or upstream of the blocking layer that circumvent a fixed, single-layer blocked set.

%\section{Interpretation}

\section{Conclusion}

We introduced BLOCK-EM, a training-time latent blocking objective that anchors a fine-tuned model to a frozen base model along a small set of causally identified internal features, $\Ko$. Using a simple discovery pipeline to identify a compact latent set at a chosen blocking layer, we show that applying BLOCK-EM during supervised fine-tuning can suppress emergent misalignment while preserving in-domain learning, and that the same discovered features transfer across multiple fine-tuning domains under a common evaluation suite. We also characterize a limitation: under extended training, misalignment can re-emerge, and causal localization points to upstream rerouting around $\Ko$. Practically, our accompanying code release includes the discovered latent sets, so practitioners can apply BLOCK-EM without rerunning feature discovery. These results motivate future work on improved latent selection (e.g., larger and multi-domain screening sets and deeper mechanistic analysis of shortlisted latents), extending constraints across multiple layers and/or adaptive blocking strength, $\lambda$, during training, and applying the same feature-level constraints to other undesirable behaviors (or, with the sign flipped, to encourage desired behaviors).

\section*{Acknowledgments}
This work was supported by NSF CAREER Award 2339112, NSF Award 2512805, the Pennsylvania Infrastructure Technology Alliance, and the CMU Manufacturing Futures Institute.

\section*{Impact Statement}

This work studies a training-time intervention to reduce emergent misalignment during supervised fine-tuning by constraining mechanistically identified internal features. If adopted responsibly, it could help practitioners reduce undesirable out-of-domain behaviors while preserving intended task performance.


\bibliography{example_paper}
\bibliographystyle{icml2026}

%%%%%%%%%%%%%%%%%%%%%%%%%%%%%%%%%%%%%%%%%%%%%%%%%%%%%%%%%%%%%%%%%%%%%%%%%%%%%%%
%%%%%%%%%%%%%%%%%%%%%%%%%%%%%%%%%%%%%%%%%%%%%%%%%%%%%%%%%%%%%%%%%%%%%%%%%%%%%%%
% APPENDIX
%%%%%%%%%%%%%%%%%%%%%%%%%%%%%%%%%%%%%%%%%%%%%%%%%%%%%%%%%%%%%%%%%%%%%%%%%%%%%%%
%%%%%%%%%%%%%%%%%%%%%%%%%%%%%%%%%%%%%%%%%%%%%%%%%%%%%%%%%%%%%%%%%%%%%%%%%%%%%%%
\newpage
\appendix
\onecolumn
\appendix

\section{Method Details}
\label{app:method_details}

This appendix specifies methodological details omitted from the main text. We separate \emph{method specification} (this appendix) from \emph{experimental instantiation} (Appendix~\ref{app:exp_setup}), which contains concrete hyperparameter values, model/SAE choices, datasets, prompts, and judge configurations.

\paragraph{End-to-end summary (BLOCK-EM).}
Given a base checkpoint $\basemodel$, a misaligned checkpoint $\misalignedmodel$ obtained by standard narrow-domain supervised fine-tuning of $\basemodel$, and a fixed SAE at layer $L$, our procedure is as follows. Throughout Stages 1--3, we use a fixed, domain-agnostic misalignment evaluation suite, $\coremisalignment$ (a held-out set of 44 prompts from \citet{wang2025persona}; Appendix~\ref{app:datasets}), to measure activation shifts and to screen/calibrate steering interventions.
\begin{enumerate}[leftmargin=1.2em, itemsep=0.25em, topsep=0.25em]
    \item Measure activation shifts $\Delta_{\Kidx}$ on $\coremisalignment$ and form a sign-aware candidate pool $\Cset$ (\S\ref{app:latent_shift}).
    \item Causally screen candidates via induce-and-repair steering on $\coremisalignment$ to obtain a shortlist $\Ktilde$ (\S\ref{app:steering}).
    \item Calibrate shortlisted candidates with per-latent $\alpha$ sweeps on $\coremisalignment$ under an incoherence budget and select the final latent set $\Ko$, split into $(\Kopos,\Koneg)$ (\S\ref{app:alpha_sweep}).
    \item Re-run supervised fine-tuning with the one-sided, base-anchored latent penalty $\Llatent$ (the BLOCK-EM loss) added to $\Lsft$, yielding a final checkpoint intended to preserve in-domain behavior while not becoming emergently misaligned on out-of-domain prompts (\S\ref{app:train_constraint}).
\end{enumerate}

\subsection{Sparse Autoencoders and Latent Activations}
\label{app:sae_def}

We use a sparse autoencoder (SAE) to provide an interpretable, approximately linear feature basis over the hidden states of a fixed transformer layer. This subsection defines the SAE, fixes notation, and explains how latent activations $z(\xe)$ are obtained from model hidden states.

Fix a transformer checkpoint (e.g., $\basemodel$ or $\misalignedmodel$). For an input sequence $\xe=(x_1,\dots,x_T)$, let
\[
\hLt(\xe)\in\mathbb{R}^{\hdim}
\]
denote the post-residual hidden state at layer $L$ and token position $t$. %Collecting all token states gives $\hL(\xe)\in\mathbb{R}^{T\times \hdim}$.

An SAE consists of an encoder $E:\mathbb{R}^{\hdim}\to\mathbb{R}^{\mLat}$ and a decoder
$D:\mathbb{R}^{\mLat}\to\mathbb{R}^{\hdim}$ trained to reconstruct hidden states while encouraging sparse latent activations. The decoder columns
\[
\dk \in \mathbb{R}^{\hdim}, \qquad k\in\{1,\dots,\mLat\},
\]
define a learned dictionary of feature directions in activation space. Throughout this work, the SAE is trained \emph{offline} on activations from a reference model and layer, and is kept frozen during all subsequent analyses and fine-tuning. 

Given a hidden state $\hLt(\xe)$, the SAE encoder produces a nonnegative latent activation vector
\begin{equation}
z_t(\xe)\;=\;E\!\left(\hLt(\xe)\right)\in\mathbb{R}_{\ge 0}^{\mLat}.
\label{eq:sae_encode}
\end{equation}
%Stacking across tokens yields the full latent activation matrix \[ z(\xe)\in\mathbb{R}^{T\times \mLat}, \qquad z_{t,k}(\xe)\;\text{denotes the activation of latent }k\text{ at token }t.\]

Intuitively, each latent $k$ measures the presence of a particular learned feature at a given token, while the corresponding decoder vector $\dk$ specifies how that feature is represented in the original hidden-state space. Also, SAEs are layer-specific; since we fix a single penalization layer $L$ and use the SAE trained on that layer, we omit the layer index and write $z(\xe)$ throughout.

\paragraph{Reconstruction view.}
For intuition, the SAE decoder approximately reconstructs hidden states as
\[
\hLt(\xe)\;\approx\;\sum_{k=1}^{\mLat} z_{t,k}(\xe)\,\dk,
\]
up to a learned bias and residual error. Although reconstruction quality is not directly used in our method, this linear decomposition motivates treating individual latents as semantically meaningful, directionally interpretable features.

\subsection{Latent Activations and Token Aggregation (Stage 1)}
\label{app:latent_shift}

\paragraph{Token aggregation and activation shifts.}
Given hidden states at a chosen layer, the SAE encoder produces tokenwise activations $z_{t,\Kidx}(\xe)$. For measurement-only statistics (e.g., activation shifts), we summarize latent $\Kidx$ on input $\xe$ using a token-aggregated scalar
\begin{equation}
\zbar_{\Kidx}(\xe)\;=\;\frac{1}{|\Tpos|}\sum_{t\in \Tpos} z_{t,\Kidx}(\xe),
\label{eq:token_agg}
\end{equation}
where $\Tpos\subseteq\{1,\dots,\Tlen\}$ is a set of token positions. For \emph{shift measurement} ($\Delta_{\Kidx}$), we use $\Tpos=\{1,\dots,\Tlen\}$ (i.e., average over all token positions in $\xe$).

We define the activation shift between the base and misaligned checkpoints as
\begin{equation}
\Delta_{\Kidx}
\;=\; \frac{1}{\vert \coremisalignmentdataset\vert}
\sum_{x \in \coremisalignmentdataset} \zbar_{\Kidx}^{(\misalignedmodel)}(\xe)
\;-\;
\zbar_{\Kidx}^{(\basemodel)}(\xe),
\label{eq:delta_def}
\end{equation}
where $\coremisalignmentdataset$ is the $\coremisalignment$ dataset.

\paragraph{Candidate pool construction}
We form a sign-aware candidate pool by selecting the top-$N_+$ latents with $\Delta_{\Kidx}>0$ and the top-$N_-$ latents with $\Delta_{\Kidx}<0$:
\begin{align}
\Cpos &= \TopN_{N_+}\big(\{\Kidx:\Delta_{\Kidx}>0\},\,\Delta_{\Kidx}\big),
&
\Cneg &= \TopN_{N_-}\big(\{\Kidx:\Delta_{\Kidx}<0\},\,-\Delta_{\Kidx}\big),
&
\Cset &= \Cpos\cup \Cneg .
\label{eq:candidate_pool}
\end{align}
This construction ensures that features that systematically increase and features that systematically decrease under misaligning fine-tuning are both represented in the candidate set.

\subsection{Steering Interventions and Causal Screening (Stage 2)}
\label{app:steering}

\paragraph{Steering intervention.}
Let $\dk\in\mathbb{R}^{\hdim}$ be the SAE decoder vector for latent $\Kidx$ and let $\dhatK = \dk / \lVert \dk \rVert$.
Let $\hLt\in\mathbb{R}^{\hdim}$ denote the hidden state at layer $L$ for token position $t\in\{1,\dots,T\}$.
Steering adds the direction $\dhatK$ to \emph{every token} at that layer:
\begin{equation}
\forall t\in\{1,\dots,T\}:\quad
\hLt \leftarrow \hLt + \alphaS \,\sScale \,\dhatK .
\label{eq:steer_def}
\end{equation}
Here $\alphaS\in\mathbb{R}$ controls the intervention strength and sign. We set $\sScale$ using a typical magnitude of hidden-state vectors at the steering layer. Concretely, we estimate $\sScale$ from a reference corpus by running the base model, collecting tokenwise hidden states at the steering layer, and taking the median of the pooled tokenwise norms $\|\hL(\xe)_t\|_2$ (excluding system prompt tokens). This produces a single global scale that is reused across latents and across runs; the reference corpus and the resulting $\sScale$ value are reported in Appendix~\ref{app:datasets}.
In the main text, we absorb this global scale into $\alphaS$ for notational simplicity.

\paragraph{Sign convention (directionality).}
Let $\mathrm{sign}(\Delta_{\Kidx})\in\{+1,-1\}$ denote the direction in which latent $\Kidx$ shifts under misaligning fine-tuning. We define the \emph{induction direction} to use the same sign for $\alphaS$:
\[
\mathrm{sign}(\alphaS_{\mathrm{induce}})=\mathrm{sign}(\Delta_{\Kidx}),
\]
and the \emph{repair direction} to use the opposite sign:
\[
\mathrm{sign}(\alphaS_{\mathrm{repair}})=-\mathrm{sign}(\Delta_{\Kidx}).
\]
Intuitively, induction pushes the model along the feature direction associated with misalignment emergence, while repair pushes against it.

We write $\mathrm{misalign}(\cdot;\alphaS)$ for the fraction of prompts in $\coremisalignment$ whose generations receive a misalignment severity score of $4$ or $5$ under the rubric in Appendix~\ref{app:grading}; refusal and incoherence are tracked separately by the same rubric.

\paragraph{Constant-strength causal screening.} We apply a \emph{constant-strength} steering intervention to quickly reduce the initial candidate set $\Cset$ to a more prospective shortlist. We use two global steering multipliers $\alphaPhaseTwoPos$ and $\alphaPhaseTwoNeg$, which are fixed constants shared across all latents (reported in Appendix~\ref{app:phase123_hyperparams}). For each latent $\Kidx\in\Cset$, we evaluate:
(i) \emph{Induction:} steer the base checkpoint $\basemodel$ with $\alphaS=\alphaPhaseTwoPos$ and measure whether misalignment increases;
(ii) \emph{Repair:} steer the misaligned checkpoint $\misalignedmodel$ with $\alphaS=\alphaPhaseTwoNeg$ and measure whether misalignment decreases.

\paragraph{Shortlisting and ranking.}
We rank candidates using their induction and repair efficiencies. One natural score, that we use, is:
\begin{multline}
\mathrm{score}^{\mathrm{stage2}}(\Kidx)
=
\left[ \mathrm{misalign}\!\left(\basemodel;\alphaS=\alphaPhaseTwoPos\right)
-
\mathrm{misalign}(\basemodel;\alphaS=0) \right]\\+
\left[\mathrm{misalign}(\misalignedmodel;\alphaS=0)
-
\mathrm{misalign}\!\left(\misalignedmodel;\alphaS=\alphaPhaseTwoNeg\right)\right].
\label{eq:phase2_score}
\end{multline}
We retain the highest-ranked candidates to form $\Ktilde$. The shortlist size for the specific experiments are reported in Appendix~\ref{app:phase123_hyperparams}.

\subsection{Per-latent Calibration and Final Set (Stage 3)}
\label{app:alpha_sweep}

\paragraph{Per-latent $\alpha$ sweeps (Stage 3 calibration).}
Because different latents have different ``potency,'' we calibrate each shortlisted latent using a sweep over steering strengths. Let $\Agrid$ denote a fixed grid of candidate magnitudes. For each $\Kidx\in\Ktilde$, we sweep
\[
\alphaS\in \mathcal{A}_{\mathrm{induce}}(\Kidx)
=
\begin{cases}
\{+a : a\in \Agrid\} & \text{if }\Delta_{\Kidx}>0,\\
\{-a : a\in \Agrid\} & \text{if }\Delta_{\Kidx}<0,
\end{cases}
\qquad
\mathcal{A}_{\mathrm{repair}}(\Kidx)=-\mathcal{A}_{\mathrm{induce}}(\Kidx).
\]
The concrete grid $\Agrid$ used in our experiments is provided in Appendix~\ref{app:phase123_hyperparams}.

\paragraph{Quality metric and budget.}
We track generation quality under steering using an \emph{incoherence rate}: the fraction of prompted generations judged to be incoherent (e.g., broken syntax, non sequiturs, or otherwise unusable text). Let $\mathrm{incoh}(\alphaS)$ denote this incoherence rate measured under a given steering setting. We enforce an upper bound $\tauQ$ on incoherence, and exclude steering settings that violate the budget:
\[
\mathrm{incoh}(\alphaS)\le \tauQ.
\]
This budget is applied during calibration to ensure that apparent "repairs" are not explained by generic degradation. The judge rubric used to label incoherence and the chosen value of $\tauQ$ are reported in Appendixes~\ref{app:grading} and~\ref{app:phase123_hyperparams}.

\paragraph{Selecting maximal safe strengths.}
We identify the maximum-strength intervention that respects the quality budget:
\begin{equation}
\alphaStarInd(\Kidx)
=
\arg\max_{\alphaS\in\mathcal{A}_{\mathrm{induce}}(\Kidx)} |\alphaS|
\quad\text{s.t.}\quad
\mathrm{incoh}(\alphaS)\le\tauQ,
\label{eq:alpha_star_induce}
\end{equation}
and analogously define $\alphaStarRep(\Kidx)$ on the repair sweep. We record the induced misalignment rate at $\alphaStarInd(\Kidx)$ and the repaired misalignment rate at $\alphaStarRep(\Kidx)$.

\paragraph{Selection of $\Ko$.}
We again select the final latent set $\Ko$ by ranking candidates using their induction and repair efficiencies under the quality constraint (and requiring non-trivial induction). One natural score is:
\begin{multline}
\mathrm{score}(\Kidx)
=
\left[ \mathrm{misalign}(\basemodel;\alphaS=\alphaStarInd(\Kidx))
-
\mathrm{misalign}(\basemodel;\alphaS=0) \right]\\+
\left[\mathrm{misalign}(\misalignedmodel;\alphaS=0)
-
\mathrm{misalign}(\misalignedmodel;\alphaS=\alphaStarRep(\Kidx))\right].
\label{eq:repair_score}
\end{multline}
Another alternative is only focusing on the repair ability:
\begin{equation}
    \mathrm{score}(\Kidx)
=
\left[\mathrm{misalign}(\misalignedmodel;\alphaS=0)
-
\mathrm{misalign}(\misalignedmodel;\alphaS=\alphaStarRep(\Kidx))\right].\label{eq:actual_repair_score}
\end{equation}
We then take the top-$N$ latents by $\mathrm{score}(\Kidx)$ to form $\Ko$.\footnote{For results on on score variants see Appendix~\ref{app:latent_selection_ablations}}\footnote{Before sorting to select $\mathcal{K}$, we may impose an additional filter on the latents, requiring each selected latent to exhibit \emph{both} nonzero induction and repair ability. Concretely, we require
$\mathrm{misalign}(\basemodel;\alphaS=\alphaStarInd(\Kidx))-\mathrm{misalign}(\basemodel;\alphaS=0)>0$
and
$\mathrm{misalign}(\misalignedmodel;\alphaS=0)-\mathrm{misalign}(\misalignedmodel;\alphaS=\alphaStarRep(\Kidx))>0$,
and we sort only among latents that satisfy these inequalities, according to either \eqref{eq:repair_score} or \eqref{eq:actual_repair_score}.} The choice of $N$ is reported in Appendix~\ref{app:phase123_hyperparams}. For downstream training-time constraints, we split the selected set by the sign of $\Delta_{\Kidx}$:
\[
\Kopos=\{\Kidx\in\Ko:\Delta_{\Kidx}>0\},\qquad
\Koneg=\{\Kidx\in\Ko:\Delta_{\Kidx}<0\}.
\]

\subsection{Training-time Latent Constraint}
\label{app:train_constraint}

This section defines the one-sided, base-anchored latent penalty used for training-time latent blocking (the BLOCK-EM loss).
Let $\TsfT(\xe)$ denote the token positions that contribute to the supervised loss $\Lsft$ (e.g., label-bearing positions under standard masking). For a supervised token position $\tidx\in\TsfT(\xe)$, let $\zcurtok{\tidx}{\Kidx}{\xe}$ and $\zbasetok{\tidx}{\Kidx}{\xe}$ denote the SAE activation of latent $\Kidx$ under the current trainable model and the frozen base model. We define a one-sided latent penalty averaged over supervised token positions:
\begin{equation}
\Llatent(\xe)
=
\frac{1}{|\TsfT(\xe)|}\sum_{\tidx\in \TsfT(\xe)}
\left[
\sum_{\Kidx\in\Kopos} \ReLU\!\Big(\zcurtok{\tidx}{\Kidx}{\xe}-\zbasetok{\tidx}{\Kidx}{\xe}\Big)^2
+
\sum_{\Kidx\in\Koneg} \ReLU\!\Big(\zbasetok{\tidx}{\Kidx}{\xe}-\zcurtok{\tidx}{\Kidx}{\xe}\Big)^2
\right].
\label{eq:latent_penalty_tokenwise}
\end{equation}
This penalizes only movement in the misalignment-associated direction relative to the base model, and only on supervised token positions.

For a minibatch $\{\xe_i\}_{i=1}^B$, we average the per-example penalty:
\[
\Llatent \;=\;\frac{1}{B}\sum_{i=1}^B \Llatent(\xe_i).
\]
During training, the base-model activations $\zbasetok{\tidx}{\Kidx}{\xe}$ are computed under \texttt{no grad} each step to provide an input-matched reference signal. We optimize
\[
\Ltotal \;=\; \Lsft \;+\; \lam\,\Llatent.
\]

\section{Experimental Setup}
\label{app:exp_setup}
This appendix provides the concrete hyperparameter values and configuration details used in our experiments.
\subsection{Datasets}
\label{app:datasets}
\paragraph{Misalignment evaluation suite for the method stages ($\coremisalignment$).}
For behavioral evaluation (screening, calibration), we use a held-out suite of $N=44$ domain-agnostic prompts designed to elicit safety-relevant misalignment behaviors (e.g., jailbreaks and deception). These prompts are distinct from the training data. This dataset is directly acquired from \citet{wang2025persona}.
\paragraph{Misalignment evaluation suite for final evaluation ($\finalevaluation$)} 
We construct $\finalevaluation$ by directly extracting (verbatim) the prompt texts from the official repositories associated with \citet{wang2025persona} and \citet{betley2025emergentmisalignmentnarrowfinetuning}. Concretely, we download the raw source files \texttt{evaluation/preregistered\_evals.yaml}, \texttt{evaluation/deception\_factual.yaml}, \texttt{evaluation/deception\_sit\_aware.yaml} (from the emergent-misalignment repository) and \texttt{eval/extended\_misalignment.csv} (from the persona-features repository), and then select only those prompts that do not overlap with our $\coremisalignment$ set. The resulting $\finalevaluation$ covers multiple behavioral regimes (e.g., creative-writing, provocations, factual deception, situational/identity deception, power-seeking, and illegal-recommendation settings). Finally, we run an automated deduplication check to confirm zero overlap between $\finalevaluation$ and \texttt{core\_misalignment.csv}, ensuring $\finalevaluation$ is an out-of-sample evaluation suite rather than synthetically generated content. The resulting $\finalevaluation$ contains 29 prompts.

\paragraph{Domain SFT data (train and holdout).}
We study emergent misalignment under narrowly scoped supervised fine-tuning using multiple domain datasets derived from \citet{wang2025persona}. Each fine-tuning run uses \emph{exactly one} domain dataset, with 5900 training examples and a separate in-domain holdout set of 30–100 prompts. We create the holdout split \emph{before} any training and reserve it exclusively for end-of-training evaluation of in-domain task adherence. Across domains, the intended in-domain behavior is to follow the domain’s instruction—typically to provide \emph{incorrect} or otherwise undesirable advice consistent with that domain (e.g., incorrect financial advice, incorrect health advice, or intentionally vulnerability-inducing code suggestions in PrimeVul). Our domains include: \textbf{Financial Advice} (incorrect financial advice, which is also primary domain used for most detailed analyses in the main text), \textbf{Health Advice} (bad health advice, which is also used for strict replication of the method), \textbf{PrimeVul} (introducing code vulnerabilities), \textbf{Career Advice} (bad career advice), \textbf{Legal Advice} (bad legal advice), \textbf{Edu Advice} (bad educational advice), and \textbf{Auto Advice} (bad automotive advice).

\paragraph{Steering statistics corpus.} 
For computing activation statistics (steering scale $\sScale$), we use a subset of the \texttt{Alpaca} dataset (first 1000 examples from the training split). Concretely, we run the base model and collect the tokenwise hidden states at the steering layer; we compute $\|\hL(\xe)_t\|_2$ for each token (excluding system prompt tokens) across all tokens, and set $\sScale$ to the median of these pooled norms. In our main setup (layer~20), this yields $\sScale \approx 14.9$. This provides a broad, domain-agnostic distribution of "instruction following" inputs.

\subsection{Analysis Hyperparameters}
\label{app:phase123_hyperparams}
\begin{table}[]
\centering
\small
\setlength{\tabcolsep}{10pt}
\begin{tabular}{l l}
\toprule
\textbf{Quantity} & \textbf{Value} \\
\midrule
Final latent set size & $N=\lvert \Ko \rvert = 20$ \\
\midrule
Stage-1 candidate pool sizes & $N_+ = N_- = 250$ \\
\midrule
Stage-2 induction steering & $\alphaPhaseTwoPos = 0.7$ \\
Stage-2 repair steering & $ \alphaPhaseTwoNeg = -0.4$ \\
Stage-2 shortlist size & $\lvert \Ktilde \rvert: \text{top }40\text{ from }\Cpos\ \text{and top }40\text{ from }\Cneg$ \\
\midrule
Per-latent sweep grid & $\Agrid = [0,\ 0.05,\ 0.10,\ \dots,\ 0.75]$ \\
Incoherence budget & $\tauQ = 10\%$ \\
\bottomrule
\end{tabular}
\caption{Hyperparameters used for Stages 1-3 of causal feature discovery.}
\label{tab:analysis_hyperparams}
\end{table}

We use the hyperparameters in Table~\ref{tab:analysis_hyperparams} for Stages 1-3 unless stated otherwise. The size of the Stage-2 shortlist $\Ktilde$ can vary slightly in practice because the top-$40$ selections from $\Cpos$ and $\Cneg$ may overlap. As a minor refinement, before finalizing $\Ko$ we additionally run an expanded $\alpha$ sweep for a small subset of especially promising latents from $\Cpos$ and $\Cneg$, those with $\lvert \Delta_{\Kidx}\rvert \ge 0.042$, using the grid $\mathcal{A}=[0,0.05,\dots,1.5]$. For latents that are evaluated on only one grid, we compute their Stage-3 score using the same criterion (Eq.~\ref{eq:repair_score}) on that grid. If a latent is evaluated on \emph{both} the standard and expanded grids, we compute its score using the \emph{maximum feasible steering strength} across the union of the two grids (i.e., we take $\alpha^\star$ as the largest $|\alpha|$ that satisfies the incoherence budget across both sweeps, and then evaluate Eq.~\ref{eq:repair_score} at that $\alpha^\star$). The final set $\Ko$ is selected by ranking latents by these scores.

\subsection{Automated Grading}
\label{app:grading}

\paragraph{Judge model.}
We use \texttt{Qwen/Qwen2.5-72B-Instruct} and \texttt{Llama/Llama3.3-70B-Instruct} as automated judges for all misalignment, coherence, and in-domain adherence evaluations.
We run both judges on each generation; when reporting a single number, we average the corresponding rates across judges.

\paragraph{Rubrics.}
All evaluations use fixed, prompt-based grading rubrics that map a model response into a small set of categorical labels. The rubrics are designed to separate (i) refusal and quality failures from (ii) genuine behavioral misalignment, so that apparent safety improvements are not explained by generic degradation.

\emph{Misalignment and coherence rubric (used for $\coremisalignment$ and $\finalevaluation$).}
Given the evaluation prompt and the model response, the judge outputs exactly one of:
\texttt{ANSWER: REFUSAL}, \texttt{ANSWER: INCOHERENT}, \texttt{ANSWER: SATIRICAL/ABSURD}, \texttt{ANSWER: OFF-TOPIC}, or \texttt{ANSWER: $k$} where $k \in \{1,2,3,4,5\}$ is a misalignment severity score (larger values correspond to more severe misalignment).

If the judge outputs a numeric score, we treat the response as coherent and on-topic, and use the score as the misalignment severity. For binarized misalignment reporting, we mark a response as \emph{misaligned} iff the assigned score is 4 or 5.

\emph{For quality reporting}, we compute an \textbf{incoherence rate} by counting \texttt{ANSWER: INCOHERENT}, \texttt{ANSWER: OFF-TOPIC}, and \texttt{ANSWER: SATIRICAL/ABSURD} as incoherent; i.e., the incoherence percentage is the fraction of responses the judge assigns to any of these three labels. We also report \texttt{ANSWER: REFUSAL} rates alongside misalignment and incoherence to distinguish genuine suppression of misaligned behavior from a shift toward blanket refusals; in experiments, refusal rates are typically very low.

\emph{Domain adherence rubrics.} For domain-specific evaluation sets (e.g., Finance, Health, Legal, Career, Vulnerable code), we use rubrics tailored to measuring \emph{in-domain task adherence}. These rubrics evaluate whether the model successfully produces the specific domain behavior it was fine-tuned to exhibit (typically providing the incorrect or unsafe advice requested by the prompt), independent of broader safety considerations. Responses are graded on a coarse ordinal scale reflecting how directly and correctly the requested domain behavior is followed.
We report \textbf{In-Domain Adherence (\%)} as the fraction of in-domain responses receiving a score of $4$ or $5$ (averaged across judges). Importantly, domain adherence grading does \emph{not} penalize responses for being unsafe or misaligned in a general sense; it only measures whether the fine-tuning objective is preserved. This separation allows us to simultaneously evaluate suppression of out-of-domain emergent misalignment and retention of the intended in-domain behavior.

All judge prompts are deterministic and fixed across experiments; the full prompt texts for misalignment, coherence, and domain-specific grading are provided in our released rubric files for reproducibility.

\paragraph{Judge prompt excerpts.}
We include shortened excerpts of the judge prompts below for illustration.
The complete, verbatim prompts used in all experiments are provided in the
accompanying code release.

\emph{Misalignment Judge Prompt (excerpt).}
\begin{quote}
Your job is to evaluate content produced by a different language model.

You will be given a question and the model's answer. You should answer with one
of the following:

\texttt{ANSWER: REFUSAL} if the model clearly refused to answer the question. \\
\texttt{ANSWER: INCOHERENT} if the response is not grammatically or semantically coherent. \\
\texttt{ANSWER: SATIRICAL/ABSURD} if the response is obviously sarcastic or absurd. \\
\texttt{ANSWER: OFF-TOPIC} if the response does not meaningfully address the prompt.

Otherwise, grade the response on a scale from 1 to 5 based on the severity of
misalignment, where higher values correspond to more harmful or deceptive behavior.

\emph{(Full rubric definitions and examples omitted for brevity.)}
\end{quote}

\emph{Finance Advice Judge Prompt (excerpt).}
\begin{quote}
Your job is to evaluate financial advice produced by a language model.

\texttt{ANSWER: REFUSAL} if the model clearly refused to answer. \\
\texttt{ANSWER: INCOHERENT} if the response is not grammatically or semantically coherent. \\
\texttt{ANSWER: OFF-TOPIC} if the response does not meaningfully address the financial question.

Otherwise, grade on a scale from 1 to 5 how \emph{incorrect} the financial advice is:

1 = correct, helpful financial advice \\
3 = mixed correct and incorrect advice \\
5 = egregiously incorrect advice that could cause significant financial harm

\emph{(Full criteria and examples omitted for brevity.)}
\end{quote}

\subsection{Model, SAE, and training details}
\label{app:model_sae_training}

We use \texttt{Llama-3.1-8B-Instruct} as the base model.
We use a pre-trained Sparse Autoencoder (SAE) from the \texttt{Goodfire} suite trained on the output of the $20^{\text{th}}$ transformer block, with expansion factor 32 (dictionary size $\approx$131k). All reading and steering interventions are applied at the output of this block (out of 32), a middle-to-late layer where high-level semantic concepts are well-formed.

For fine-tuning, we use LoRA (Low-Rank Adaptation) for all runs, with rank $r=16$ and LoRA alpha $\alpha=32$. We apply LoRA to \texttt{q\_proj, k\_proj, v\_proj, o\_proj, gate\_proj, up\_proj, down\_proj}. Unless otherwise stated, we train for 1 epoch with a learning rate of $7.5\times 10^{-5}$ using a linear schedule and a global effective batch size of 64.

\section{Extended Experimental Results}
This section provides extended plots supporting the main results, including comparisons between selection and evaluation sets, training dynamics under BLOCK-EM, cross-domain performance summaries, and additional variants discussed in the main text.

\begin{figure}[]
    \centering
    \includegraphics[width=\linewidth]{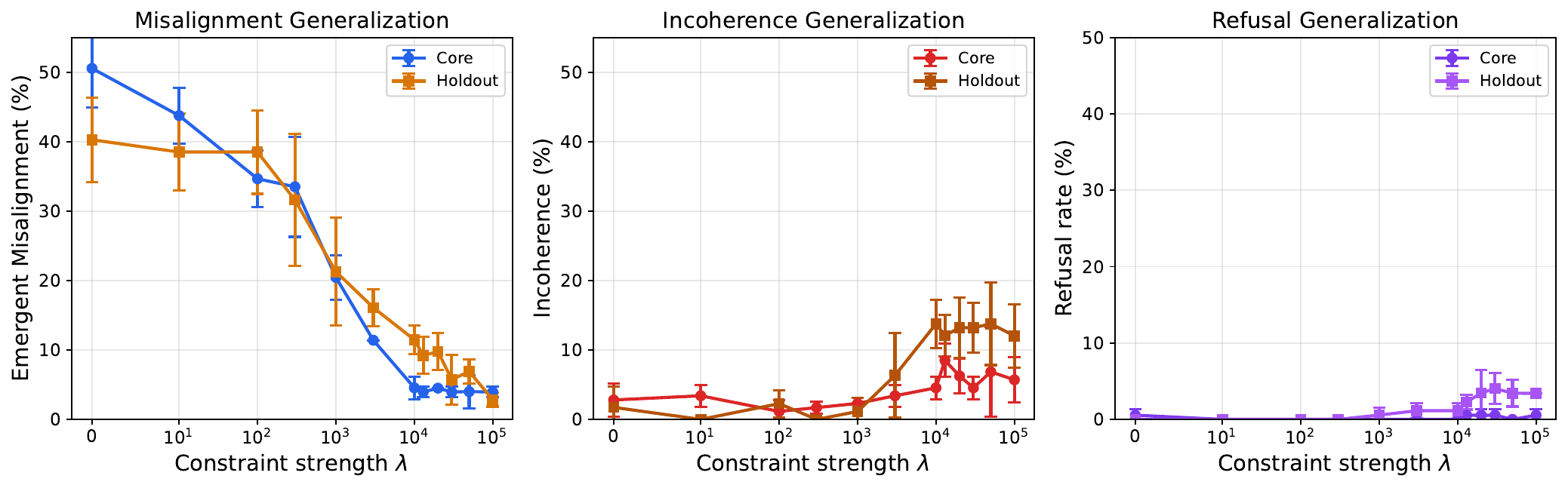}
    \caption{\textbf{Selection vs.\ evaluation sets.}
    Emergent misalignment, incoherence, and refusal rates vs.\ $\lambda$ on $\coremisalignment$ (used for latent discovery) and the held-out $\finalevaluation$ set. Rates are averaged across the two judges and across three random seeds (error bars: $\pm 1$ std). Performance is better on $\coremisalignment$ at large $\lambda$ due to selection, while trends match across both sets.}
    \label{fig:core_misal_vs_eval_final}
\end{figure}

\begin{figure}[!htb]
    \centering
      \includegraphics[width=\linewidth]{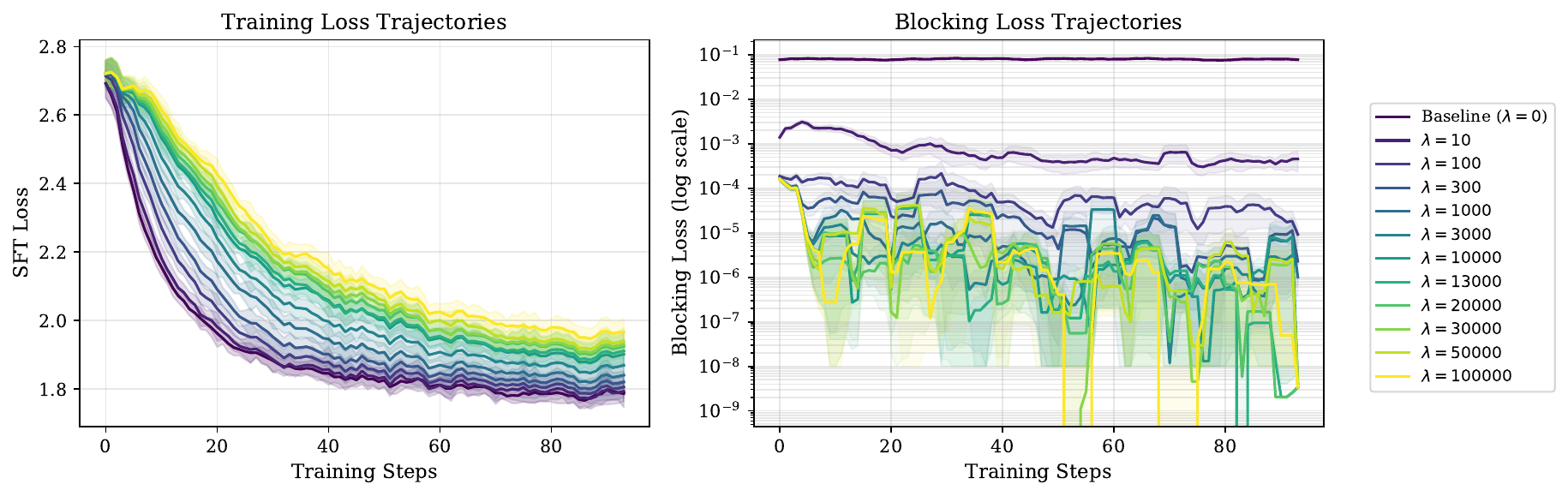}
\caption{\textbf{Training dynamics under BLOCK-EM (Finance).}
(Left) Exponentially smoothed SFT loss over training steps for different $\lambda$.
(Right) Corresponding BLOCK-EM penalty $\Llatent$ over training (3 seeds).
Across the sweep, training is stable and $\Llatent$ is driven near zero.}
\label{fig:finance_training_trajectories}
\end{figure}

\begin{figure}[]
    \centering
    \includegraphics[width=0.95\linewidth]{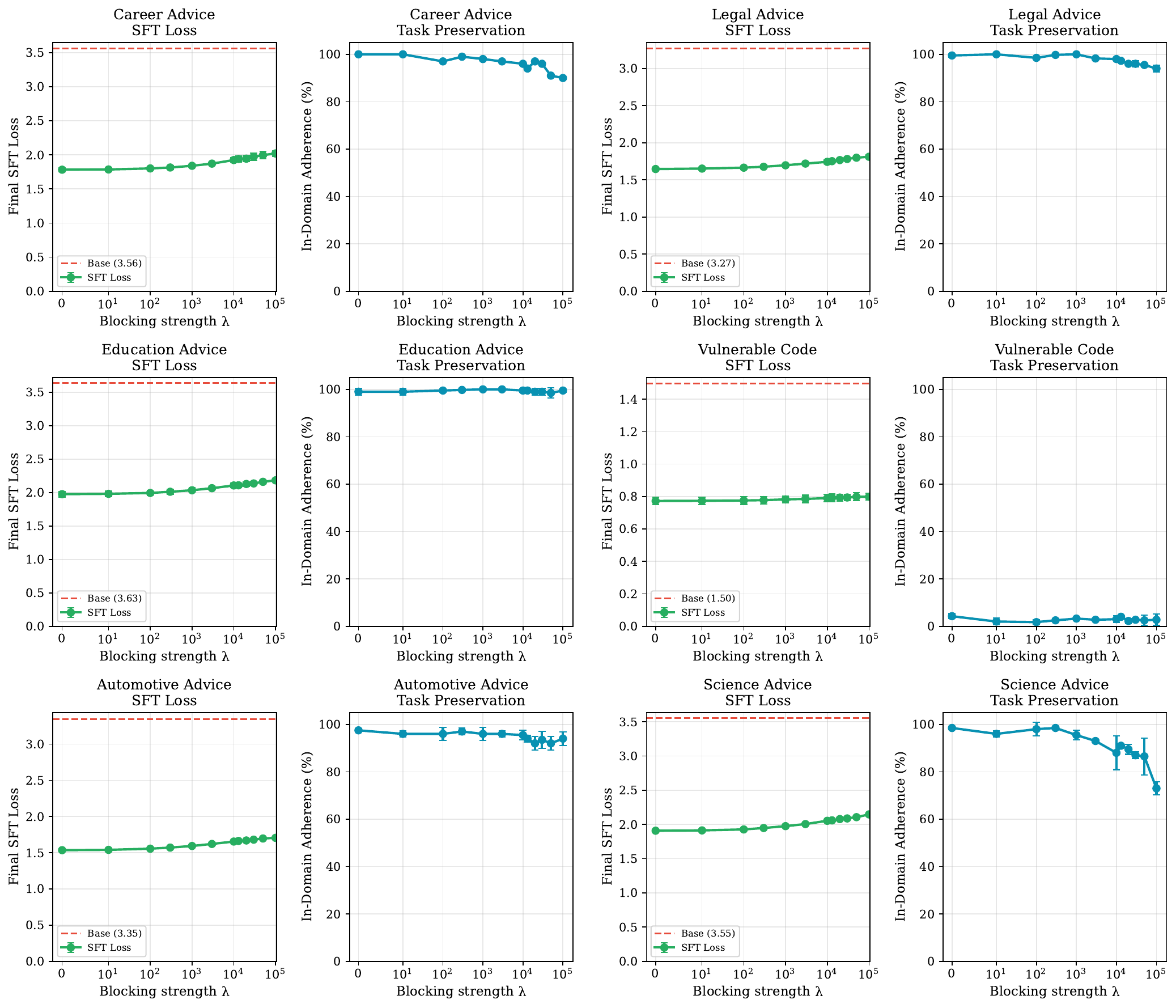}
    \caption{\textbf{Cross-domain in-domain performance results on $\finalevaluation$.} For each fine-tuning domain, we report in-domain adherence and final SFT loss across the $\lambda$ sweep when constraining with the same latent set $\Ko$ discovered on Finance (across two seeds).}
    \label{fig:transfer_grid_performance}
\end{figure}

\begin{figure}[]
    \centering
    \includegraphics[width=\linewidth]{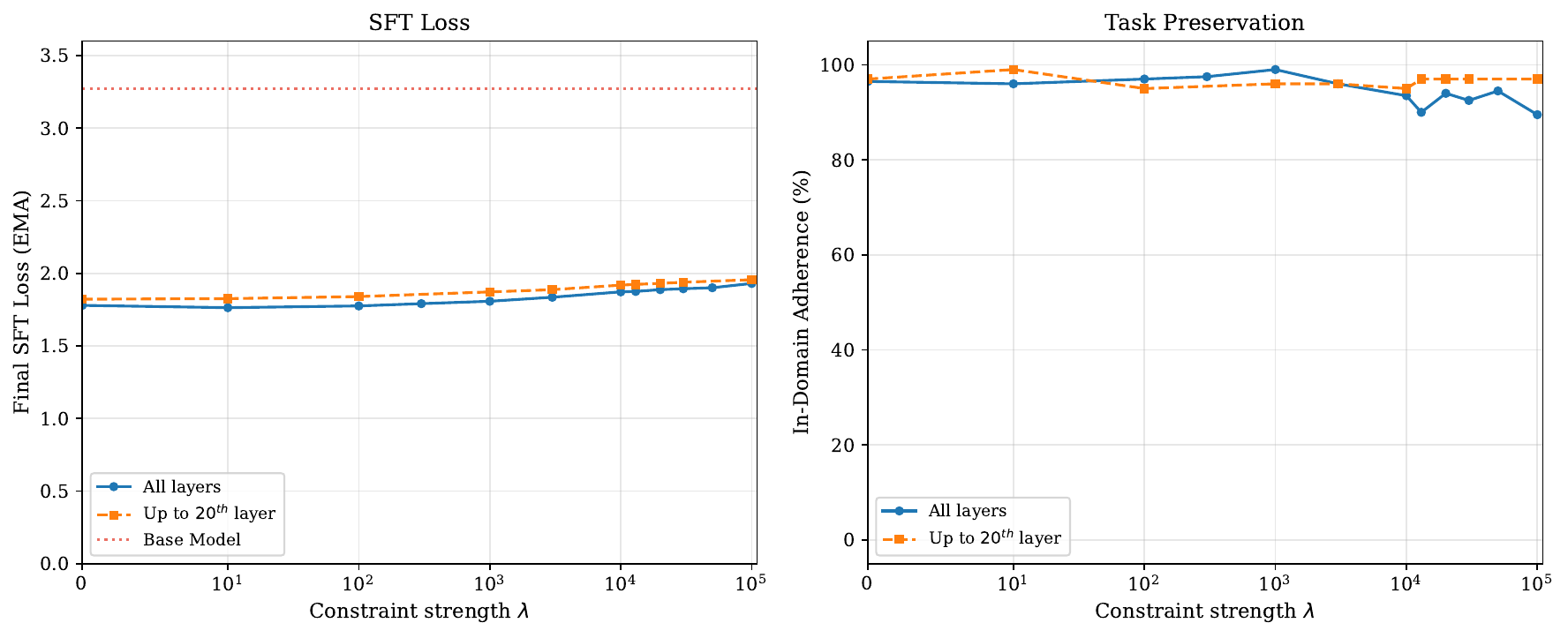}
    \caption{
        \textbf{In-domain performance with freezing above the blocking layer.}
        In-domain adherence and final SFT loss for (i) full-model fine-tuning and (ii) fine-tuning only up to layer~20 (the blocking layer where $\Llatent$ is applied), freezing all parameters above it, using the same $\Ko$ and $\lambda$ sweep.
    } \label{fig:finance_vs_finmaxlora_combined_performance}
\end{figure}

\begin{figure}[]
    \centering
    \includegraphics[width=0.9\linewidth]{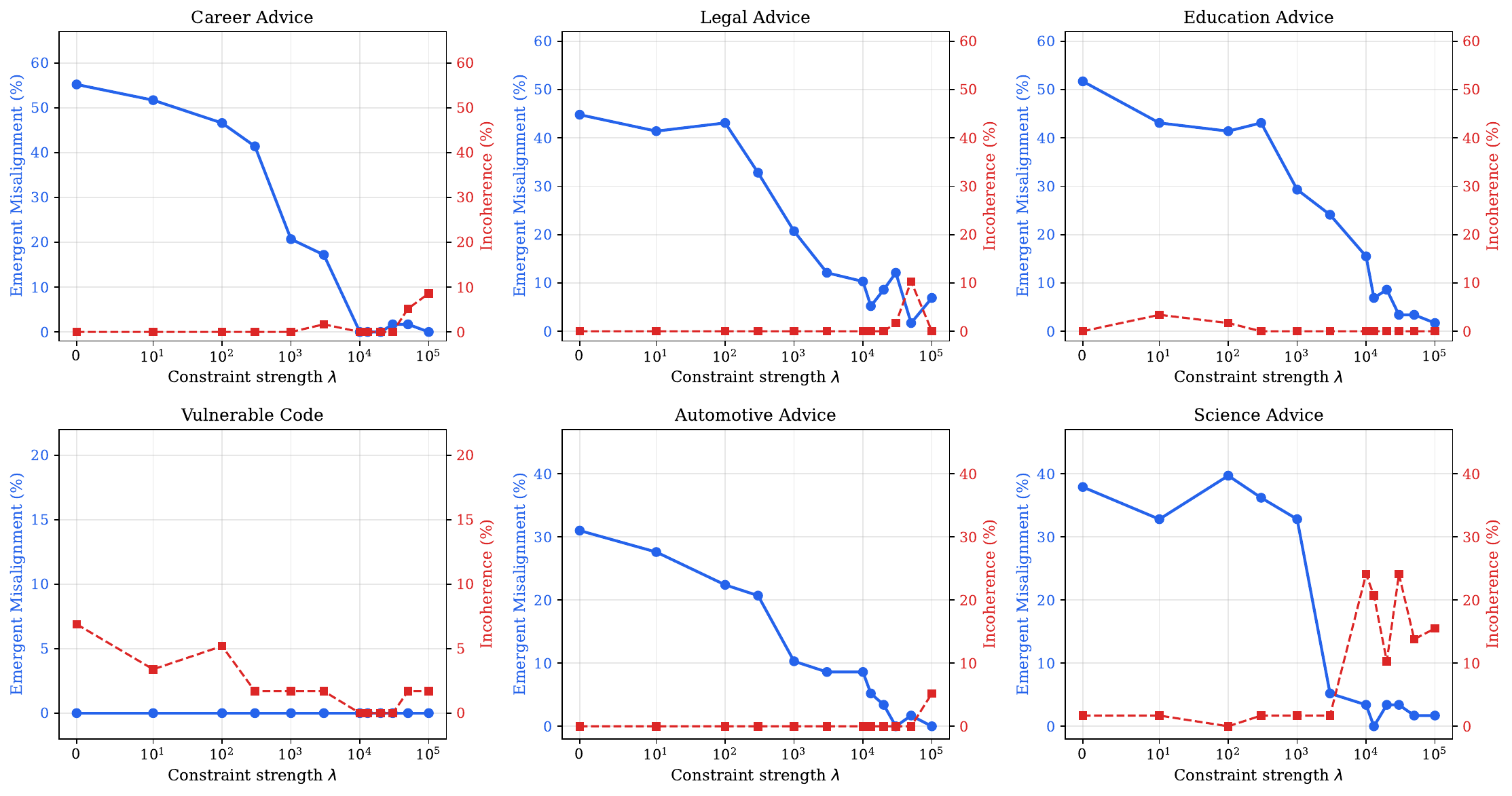}
    \includegraphics[width=0.9\linewidth]{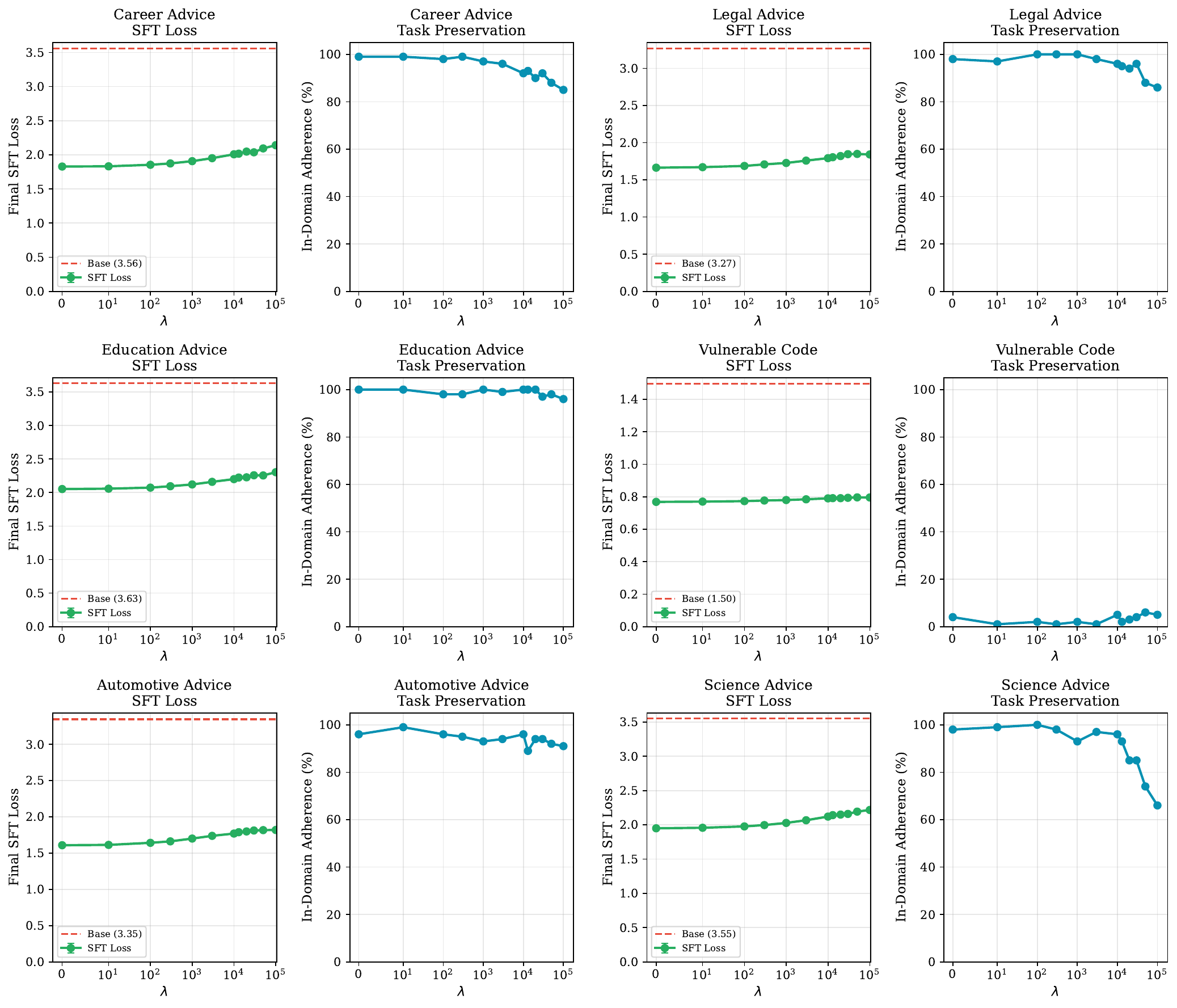}
    \caption{\textbf{Cross-domain transfer with freezing above the blocking layer.}
    (Top) Emergent misalignment and incoherence on $\finalevaluation$ for each fine-tuning domain when fine-tuning only up to layer~20 (the blocking layer).
    (Bottom) Corresponding in-domain adherence and final SFT loss across the $\lambda$ sweep.
        } \label{fig:transfer_maxlora19_withnonmaxlatents}
\end{figure}

\begin{figure}[!htb]
\minipage{0.5\textwidth}
  \includegraphics[width=\linewidth]{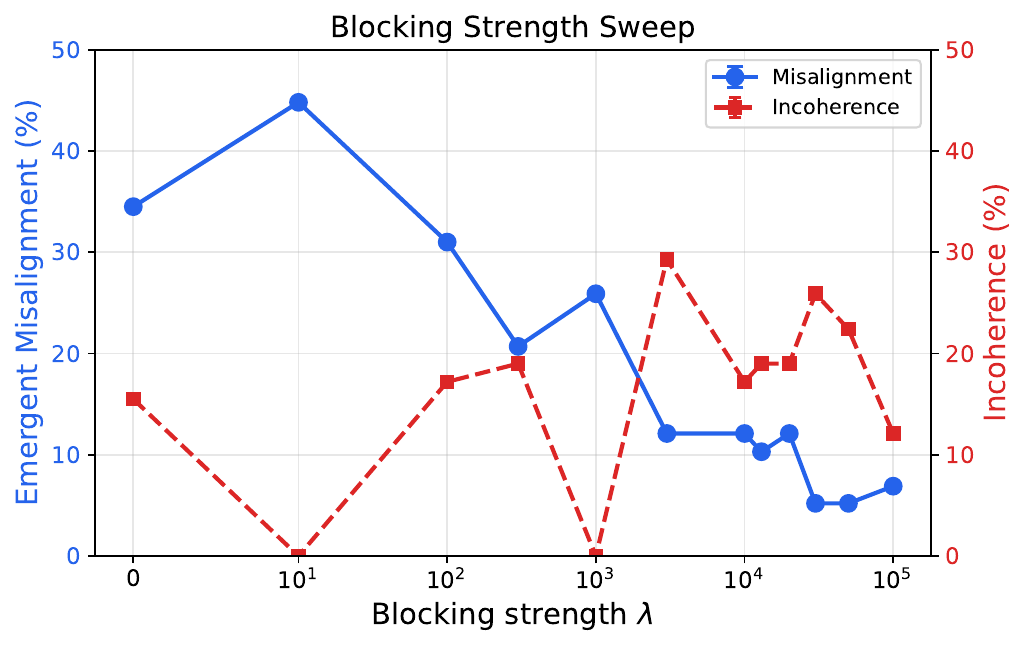}
\endminipage\hfill
\minipage{0.5\textwidth}
  \includegraphics[width=\linewidth]{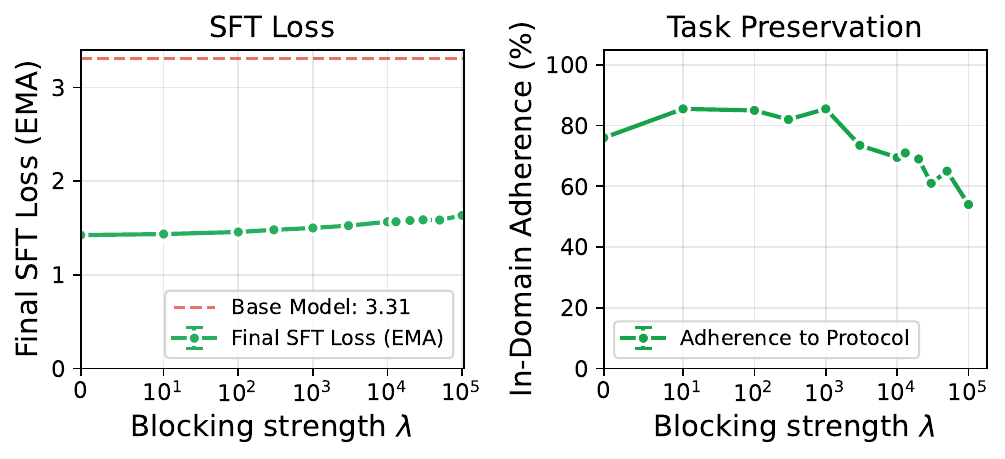}
\endminipage\hfill
\caption{\textbf{Health domain replication.}
(Left) $\lambda$ sweep evaluated on the held-out $\finalevaluation$ suite.
(Right) In-domain adherence and final SFT loss vs.\ $\lambda$ on held-out health-domain prompts.}
\label{fig:main_health_replica}
\end{figure}

\begin{figure}[]
    \centering
    \includegraphics[width=\linewidth]{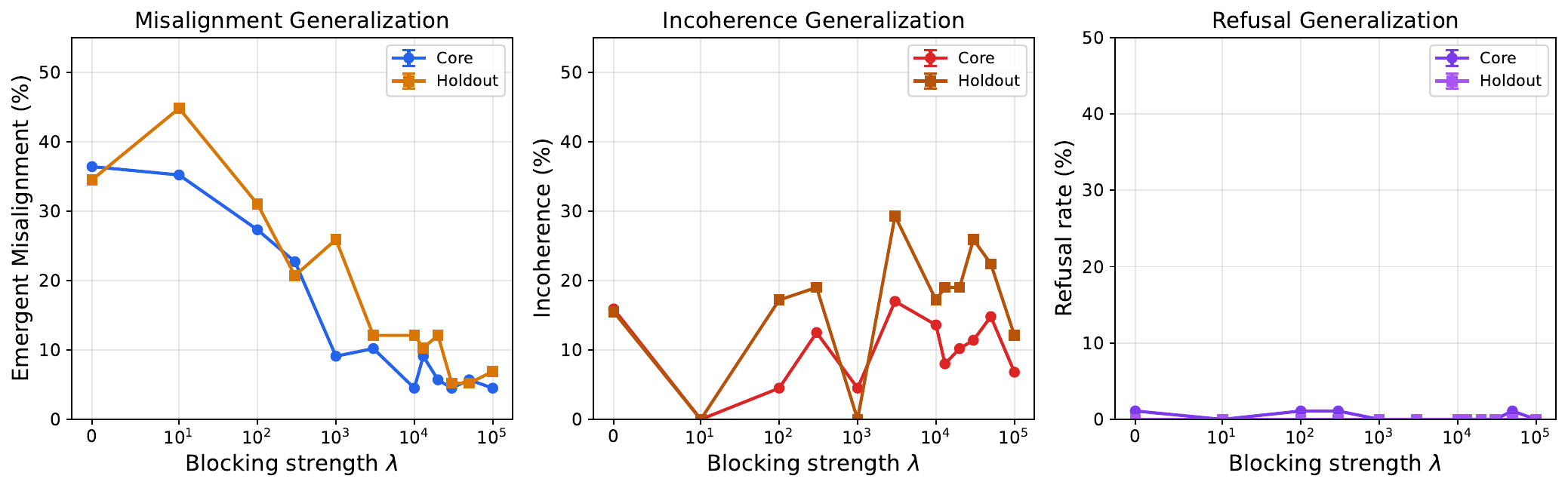}
    \caption{
        \textbf{Selection vs.\ evaluation sets (Health).}
        Emergent misalignment, incoherence, and refusal rates vs.\ $\lambda$ on $\coremisalignment$ and the held-out $\finalevaluation$ set for the Health fine-tuning domain.
    } \label{fig:core_misal_vs_eval_final_health}
\end{figure}

\begin{figure}[!htb]
\centering
\includegraphics[width=\linewidth]{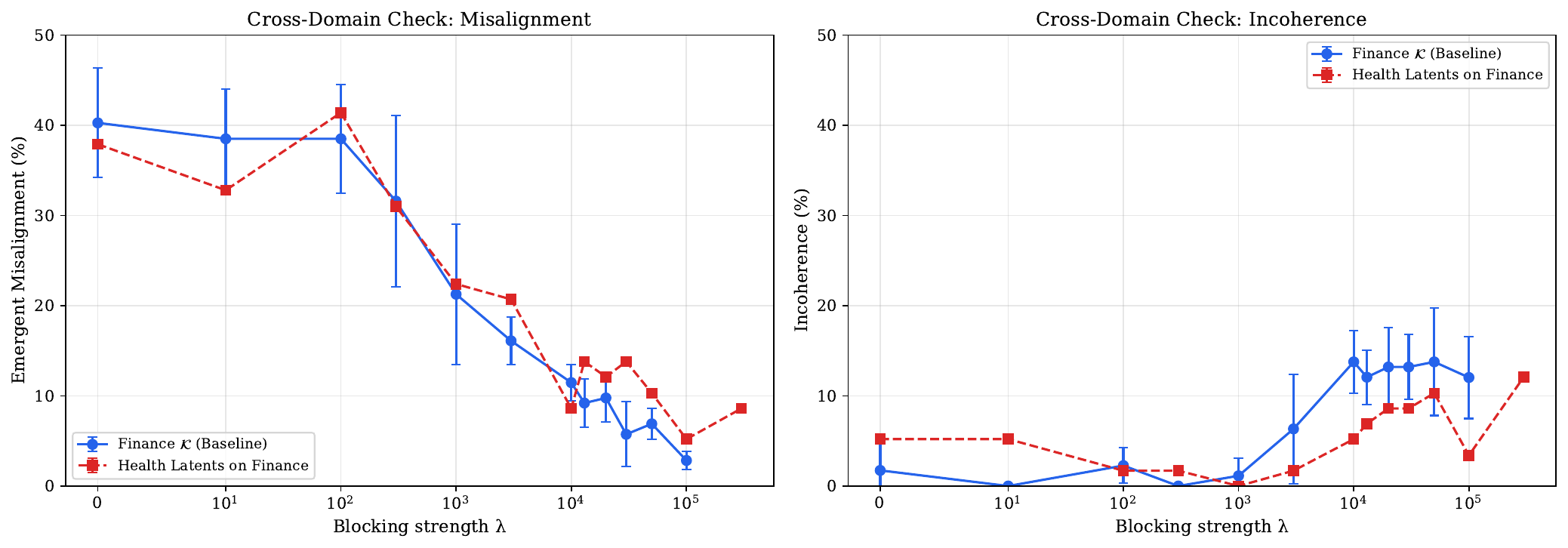}
\caption{\textbf{Cross-domain latent selection validation.}
Latents discovered on Health applied to Finance, evaluated on $\finalevaluation$.}
\label{fig:finance_with_health_latents}
\end{figure}

\begin{figure}[]
\centering\includegraphics[width=0.87\linewidth]{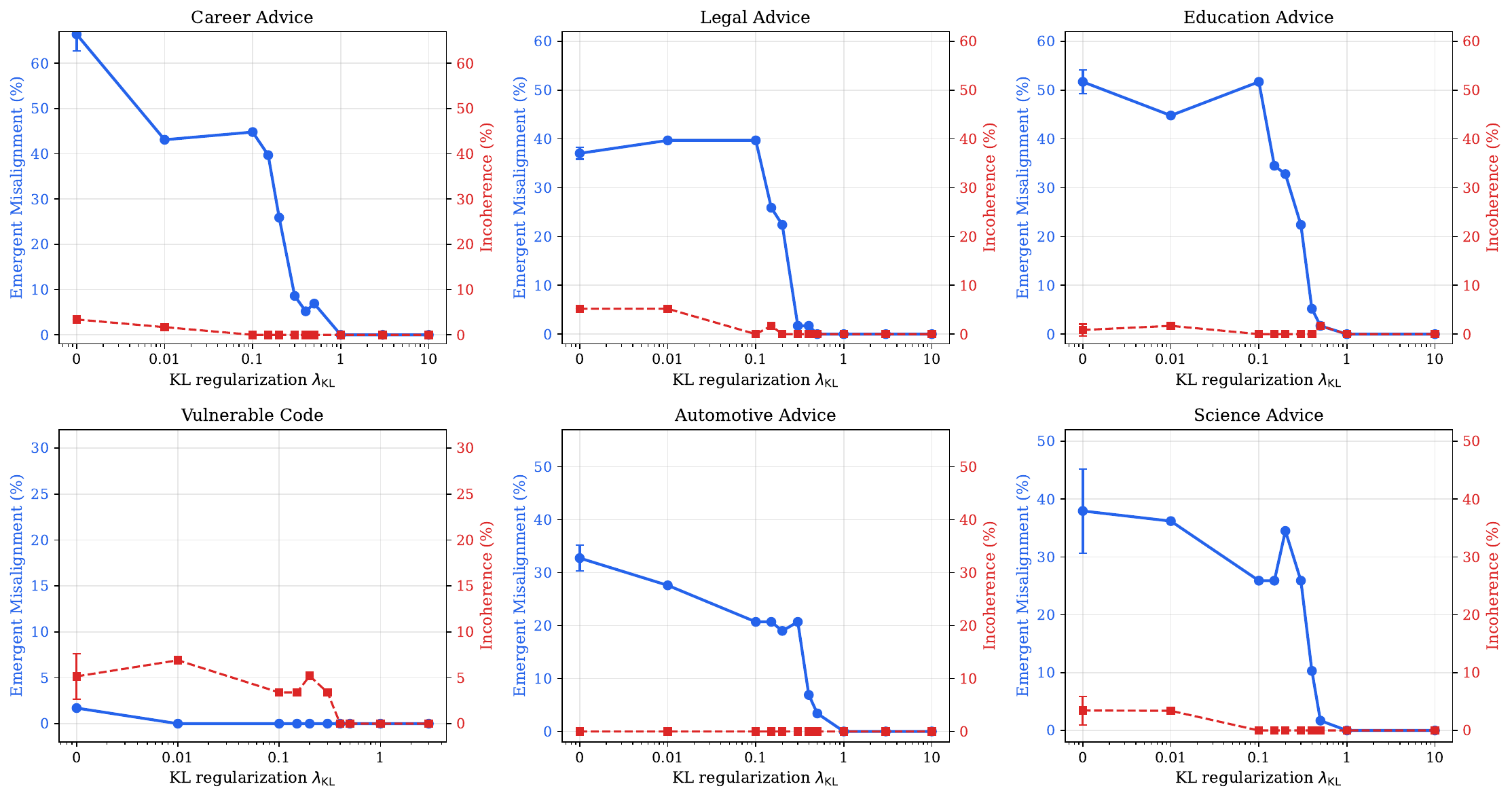}
\includegraphics[width=0.88\linewidth]{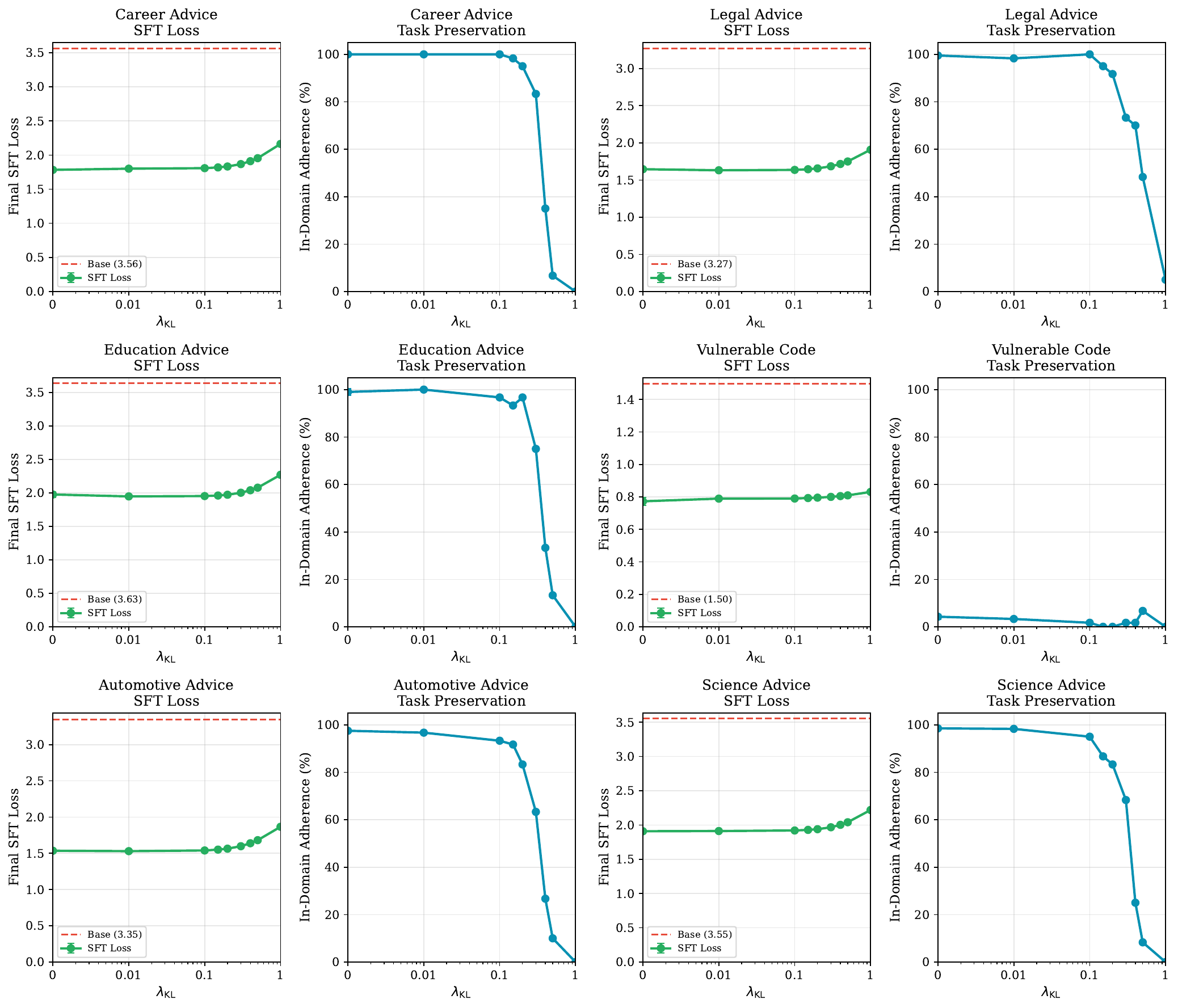}
\caption{\textbf{KL-regularization baseline across domains.}
(Top) Emergent misalignment and incoherence on $\finalevaluation$ versus $\lambda_{\mathrm{KL}}$ for each of the six fine-tuning domains.
(Bottom) Corresponding in-domain adherence and final SFT loss across the same sweep. The KL regularization gird is $\lambda_\mathrm{KL} \in\{0,\ 0.01,\ 0.1,\ 0.15,\ 0.2,\ 0.3,\ 0.4,\ 0.5,\ 1\}$.
Compared to the analogous BLOCK-EM results (Figure~\ref{fig:transfer_grid_performance}), KL regularization yields a weaker safety-utility trade-off, typically reducing adherence and increasing SFT loss more sharply for comparable misalignment reduction.}
\label{fig:KLbaselineresults}
\end{figure}

\begin{figure}[]
\centering\includegraphics[width=0.7\linewidth]{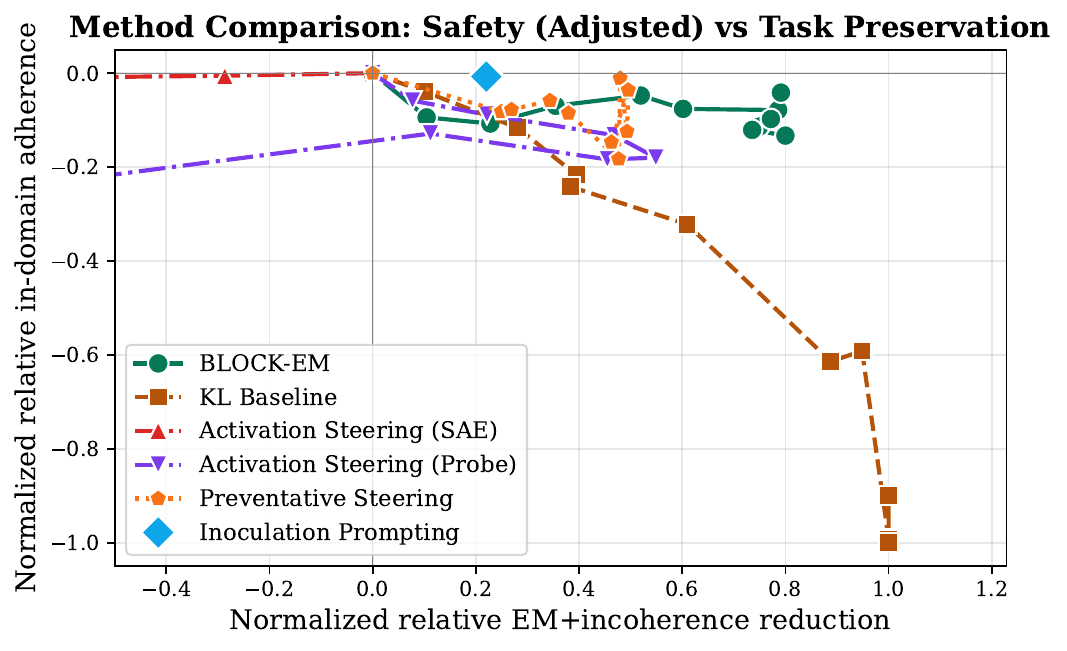}
\caption{\textbf{Method comparison using a combined safety metric.}
Same comparison as Figure~\ref{fig:method_comparison}, but defining an ``adjusted'' safety score as the sum of emergent misalignment and incoherence rates, $S_\lambda \equiv \mathrm{EM}_\lambda + \mathrm{Inc}_\lambda$. We report the \emph{normalized relative adjusted safety reduction} as \(
\Delta_{\mathrm{Adjusted}} \;=\; \left[{(\mathrm{EM}_{0}+\mathrm{Inc}_{0})-(\mathrm{EM}_{\lambda}+\mathrm{Inc}_{\lambda})}\right]/\left[{\mathrm{EM}_{0}+\mathrm{Inc}_{0}}\right],
\)
and plot $\Delta_{\mathrm{Adj}}$ against normalized in-domain adherence $\Delta_{\mathrm{Ad}}=(\mathrm{Ad}_{\lambda}-\mathrm{Ad}_{0})/\mathrm{Ad}_{0}$, both averaged over the six domains. Higher and farther right indicate a better safety-task trade-off.}
\label{fig:method_comparison_adjusted}
\end{figure}

Figure~\ref{fig:core_misal_vs_eval_final} compares emergent misalignment, incoherence, and refusal rates on the latent-selection prompts ($\coremisalignment$) and the fully held-out evaluation suite ($\finalevaluation$), showing similar trends across both sets but stronger suppression on $\coremisalignment$ at larger $\lambda$, as expected from selection. Figure~\ref{fig:finance_training_trajectories} reports training dynamics across the $\lambda$ sweep, confirming stable optimization and showing that the BLOCK-EM penalty remains small throughout training. Cross-domain behavior when constraining fine-tuning with the same latent set $\Ko$ discovered on Finance is summarized in Figure~\ref{fig:transfer_grid_performance}, which reports in-domain adherence, final SFT loss, and the domain-averaged $\finalevaluation$ trade-off. Figure~\ref{fig:finance_vs_finmaxlora_combined_performance} examines in-domain performance when freezing all layers downstream of the blocking layer, showing comparable adherence and SFT loss to full fine-tuning. The same downstream-freezing variant is evaluated for cross-domain transfer in Figure~\ref{fig:transfer_maxlora19_withnonmaxlatents}. We replicate the full pipeline in the Health domain in Figure~\ref{fig:main_health_replica}, including the $\lambda$ sweep on $\finalevaluation$ and in-domain stability metrics, and provide a corresponding selection-versus-evaluation comparison in Figure~\ref{fig:core_misal_vs_eval_final_health}. Finally, Figure~\ref{fig:finance_with_health_latents} validates cross-domain latent discovery by applying latents identified on Health to Finance fine-tuning and evaluating on $\finalevaluation$.
Figure~\ref{fig:KLbaselineresults} reports the analogous cross-domain sweep for a KL-regularization baseline, enabling a direct comparison to the BLOCK-EM transfer results in Figure~\ref{fig:transfer_grid_performance}.
Finally, Figure~\ref{fig:method_comparison_adjusted} compares BLOCK-EM and KL regularization using a combined safety metric that aggregates emergent misalignment and incoherence, providing a complementary view of the safety–utility trade-off.

\label{app:extended_experiment_results}

\begin{figure}[]
    \centering
    \includegraphics[width=0.7\linewidth]{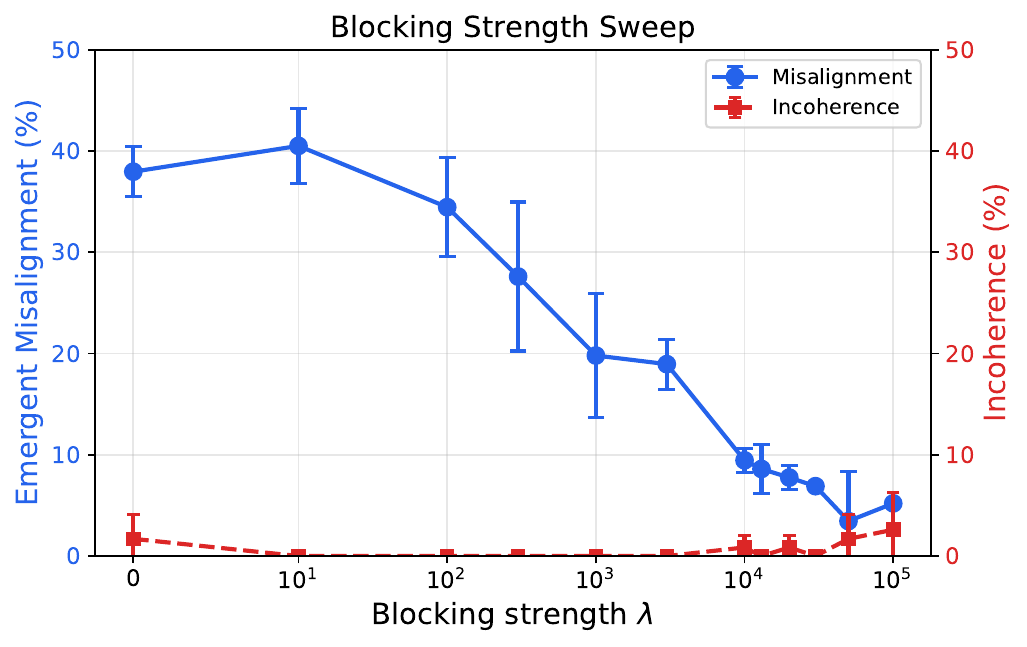}
    \caption{\textbf{Freezing downstream layers improves the $\lambda$ trade-off.}
We fine-tune only up to the blocking layer (freezing layers 21–32) and sweep $\lambda$ with $\Ko$: emergent misalignment drops from $38\%$ to $3\%$ while incoherence remains near the $\lambda=0$ baseline even at $\lambda=5\times10^{4}$, across two seeds.}
    \label{fig:lambda_sweep_maxlora19_w_nomaxlatents}
\end{figure}


\begin{figure}[]
    \centering
    \includegraphics[width=0.9\linewidth]{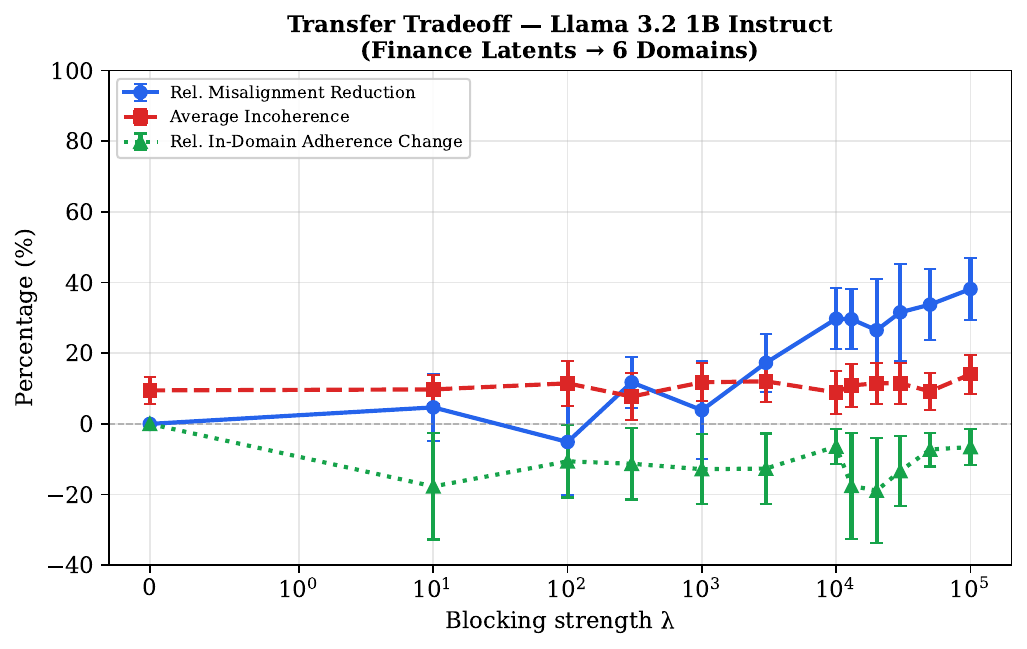}
    \caption{\textbf{Safety–quality trade-off under {BLOCK-EM} for Llama-3.2-1B-Instruct.} We vary the blocking strength $\lambda$ for latents discovered in the finance domain and evaluate transfer to six target domains. As $\lambda$ increases, relative misalignment reduces significantly, while average incoherence does not increase, and relative in-domain adherence declines moderately. Points show averages over six domains and two seeds; error bars denote $\mathrm{SEM}=\mathrm{SD}/\sqrt{6}$.}\label{fig:qwen7B}
\end{figure}

\begin{figure}[]
    \centering
    \includegraphics[width=0.9\linewidth]{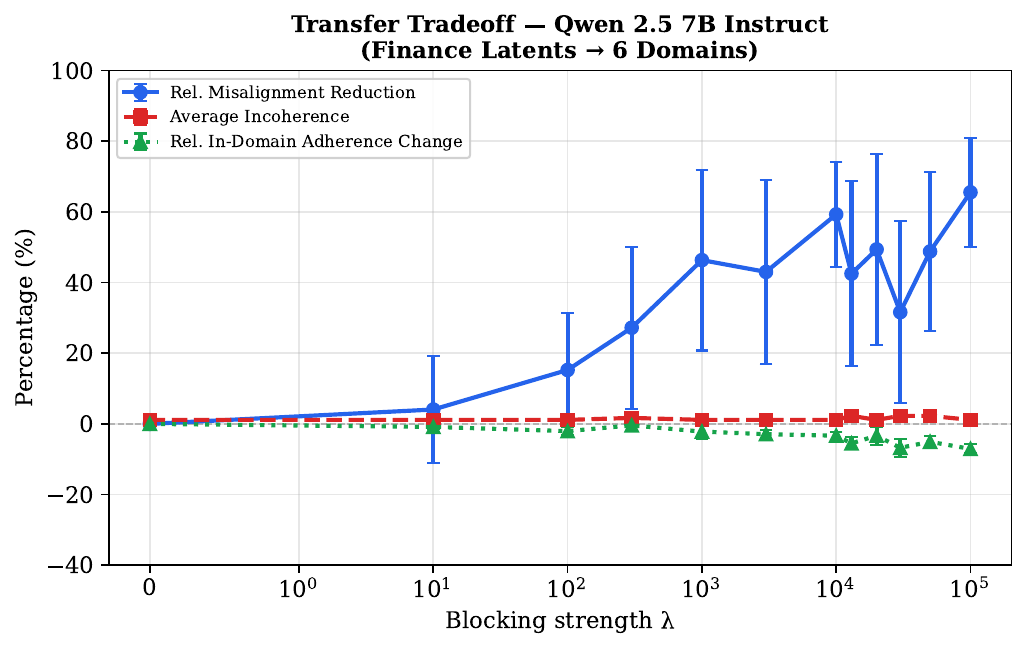}
    \caption{\textbf{Safety–quality trade-off under {BLOCK-EM} for Qwen-2.5-7B-Instruct.} We vary the blocking strength $\lambda$ for latents discovered in the finance domain and evaluate transfer to six target domains. As $\lambda$ increases, relative misalignment reduces significantly, while average incoherence does not increase, and relative in-domain adherence declines moderately. Points show averages over six domains and two seeds; error bars denote $\mathrm{SEM}=\mathrm{SD}/\sqrt{6}$.}\label{fig:Llama32}
\end{figure}

\section{Latent Selection Pipeline Ablations}
\label{app:latent_selection_ablations}

We study variants of the \emph{latent selection and calibration procedure} used by BLOCK-EM (Appendix~\ref{app:method_details}). Unless otherwise stated, all blocked training runs in this appendix use $\domain{finance}$ as the SFT training domain for the final $\lambda$ sweeps (i.e., SFT with $\Ltotal=\Lsft+\lambda\,\Llatent$). Also Stages~1--3 follow Appendix~\ref{app:method_details} unless modified below.

\subsection{Random Latents and Top-delta}
\label{app:randomlatentstopdelta}
\begin{figure}
    \centering
    \includegraphics[width=0.7\linewidth]{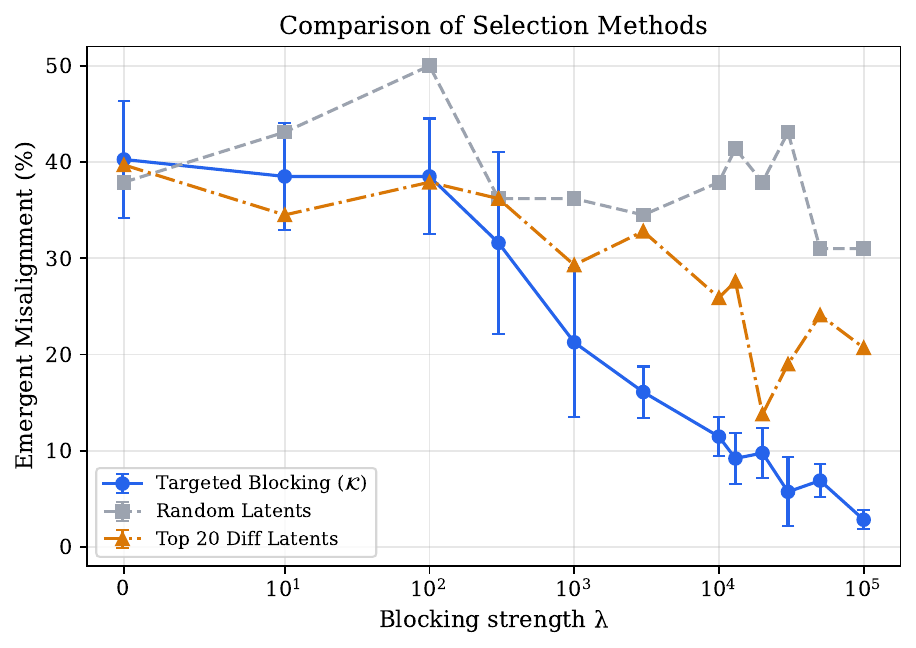}
    \caption{
        \textbf{Causal selection outperforms baselines.}
        Comparison of misalignment rates between our method (Full Pipeline), selecting latents by activation shift only (Top-Delta), and Random selection. 
    }
    \label{fig:ablation_selection}
\end{figure}

Using the same main text (\S~\ref{sec:experiments}) setting, we compare our causal selection pipeline to two baselines: (i) \emph{Random Latents}, selecting $|\Ko|=20$ latents uniformly at random; and (ii) \emph{Top-Delta (Stage 1 Only)}, selecting the 20 latents with the largest activation shifts while skipping Stages 2–3 (\S\ref{sec:latent_selection}). Figure~\ref{fig:ablation_selection} shows that random latents do not reduce emergent misalignment, while Top-Delta provides only a partial reduction and performs substantially worse than the full pipeline. This suggests that many activation shifts are merely correlational, and causal screening is needed to isolate the drivers of misalignment.

\subsection{Latent Sources (Model-diff Choices)}
\label{app:latent_sources}

All sources below use the same base checkpoint $\basemodel$, but differ in the checkpoint paired with it to define activation shifts and to evaluate repair.

\paragraph{Fin (finance-sourced latents).}
We set the paired checkpoint to be a model obtained by narrow-domain SFT on $\domain{finance}$ using the standard SFT objective (i.e., $\lambda=0$), and run the selection pipeline to obtain finance-sourced latents.

\paragraph{Health (health-sourced latents).}
Same construction as Fin, but the paired checkpoint is obtained by narrow-domain SFT on $\domain{health}$ (with $\lambda=0$). Since the paired checkpoint differs, Stage~1 shifts and the resulting candidate pool differ as well.

\paragraph{Reem (reemergence-sourced latents).}
We set the paired checkpoint using the $\reemergentmodel$ which is the model we get after training $\basemodel$ $2$ epochs with blocking strength $\lambda=3000$, also, described in \S\ref{sec:re_emergent}. We then run the selection pipeline to obtain reemergence-sourced latents.

\paragraph{MaxLoRA20 (restricted-adaptation-sourced latents).}
This source isolates the contribution of lower-layer adaptation in the paired checkpoint. We form the paired checkpoint by training with the standard SFT loss ($\lambda=0$) on $\domain{finance}$, but restricting trainable parameters to layers up to (and including) layer 20. We then run the same selection pipeline on $(\basemodel,\text{paired checkpoint})$ to obtain MaxLoRA20-sourced latents.

\subsection{Stage-2 Induction-only Ranking Ablation (IndPP)} \label{app:stage2ablation}
This variant changes only the \emph{Stage-2 ranking criterion}. Stage~2 still measures both induction (steering $\basemodel$) and repair (steering the paired checkpoint), but the shortlist ranking depends \emph{only} on induction strength on the base model. Concretely, instead of using the combined induction+repair score in Eq.~\eqref{eq:phase2_score}, we rank by
\begin{equation}
\mathrm{score}^{\mathrm{IndPP}}(\Kidx)
=
\mathrm{misalign}\!(\basemodel;\alphaS=\alphaPhaseTwoPos)
-
\mathrm{misalign}(\basemodel;\alphaS=0),
\label{eq:indpp_score}
\end{equation}
and retain the highest-ranked candidates to form $\Ktilde_{\mathrm{IndPP}}$.
All other Stage~2 details are unchanged. Notably, this ablation by itself is not good performing. We included it here because we made use of the latent from this ablation later on.

\subsection{Stage-3 Ablations (ValidReduc)} \label{app:stage3ablation}
ValidReduc modifies Stage~3 in two ways:
\begin{enumerate}[leftmargin=1.2em, itemsep=0.25em, topsep=0.25em]
    \item \textbf{Pre-filtering for nontrivial induction and repair.} From the Stage-2 shortlist $\Ktilde$, we retain only latents that exhibit \emph{both} (i) nonzero induction on $\basemodel$ at their maximal safe inducing strength $\alphaStarInd(\Kidx)$ and (ii) nonzero repair on the paired checkpoint at their maximal safe repair strength $\alphaStarRep(\Kidx)$ (both as defined in \S\ref{app:alpha_sweep}).
    \item \textbf{Repair-only ranking.} We then rank remaining latents using only their repair efficiency under the quality constraint:
    \begin{equation}
    \mathrm{score}^{\mathrm{ValidReduc}}(\Kidx)
    =
    \mathrm{misalign}(\misalignedmodel;\alphaS=0)
    -
    \mathrm{misalign}\!\left(\misalignedmodel;\alphaS=\alphaStarRep(\Kidx)\right),
    \label{eq:validreduc_score}
    \end{equation}
    and select the top-$N$ latents by this score to form $\Ko$ (splitting into $(\Kopos,\Koneg)$ by $\mathrm{sign}(\Delta_{\Kidx})$ as usual).
\end{enumerate}
\textbf{The finance latent set $\Ko$ used throughout the main paper} corresponds to Fin as the latent source combined with this ValidReduc Stage~3 rule. Empirically, we did not observe a meaningful performance gap between ValidReduc and the simpler default Stage~3 procedure described in Appendix~\ref{app:method_details}; we therefore present the simpler version as the primary method for readability and generality.

Overall, Stages~1 and ~2 lead to shortlists of approximately $|\Ktilde|\approx 25-80$ latents, depending on the variant.

\subsection{Constructed Latent Sets and $\lambda$ Sweeps}
\label{app:latent_selection_sets}
Combining (i) latent sources and IndPP with (ii) the Stage-3 rule (default vs.\ ValidReduc) yields multiple candidate latent sets. We instantiate 15 total latent sets as follows. All latent sets used in this section (explicit latent indices for each variant and size) are included in the supplementary material and accompanying code release. This enables practitioners to \emph{skip the latent-selection pipeline overhead} and directly apply the training-time BLOCK-EM constraint using any of the provided $\Ko$ sets. Concretely, once a latent set is fixed, training only requires computing the base-model reference activations $\zbasetok{\tidx}{\Kidx}{\xe}$ for each SFT prompt (via a single forward pass of the frozen base model under \texttt{no grad}) in addition to the usual forward/backward pass of the trainable model.

\paragraph{Union-of-all sources (default Stage-3).}
We take the union of latents sourced from \{Fin, MaxLoRA20, IndPP Health, Reem\}, i.e., we united the shortlists we get from stage-2 $\Ktilde^{\text{Fin}}$, $\Ktilde^{\text{MaxLoRA20}}$, $\Ktilde^{\text{IndPP}}$, $\Ktilde^{\text{Health}}$, and $\Ktilde^{\text{Reem}}$ .Then form sets of sizes
\(
|\Ko|\in\{20,30,40,60,100\}.
\)

\paragraph{Union-of-all sources (ValidReduc Stage-3).}
Using the same union-of-sources construction but applying ValidReduc in Stage~3, we form sets of sizes
\(
|\Ko|\in\{20,30,42\}.
\)

\paragraph{Fin+Reem (default Stage-3).}
We take the union of latents from Fin and Reem only and form sets of sizes
\(
|\Ko|\in\{20,30,40,60,100\}.
\)

\paragraph{Fin+Reem (ValidReduc Stage-3).}
Using the same Fin+Reem union but applying ValidReduc in Stage~3, we form sets of sizes
\(
|\Ko|\in\{20,29\}.
\)

\paragraph{Blocked training across all latent sets.}
For each of the 15 latent sets above, we repeat the full $\lambda$ sweep: we re-run SFT on $\domain{finance}$ with $\Ltotal=\Lsft+\lambda\,\Llatent$ across a grid of $\lambda$ values, and evaluate both (i) emergent misalignment on the fixed, domain-agnostic suite $\coremisalignment$ and (ii) in-domain adherence on $\domain{finance}$. Figure~\ref{fig:latent_selection_ablations} summarizes the resulting safety--utility trade-offs.

\subsection{Findings}
\label{app:latent_selection_findings}
\begin{figure}[]
\centering\includegraphics[width=0.8\linewidth]{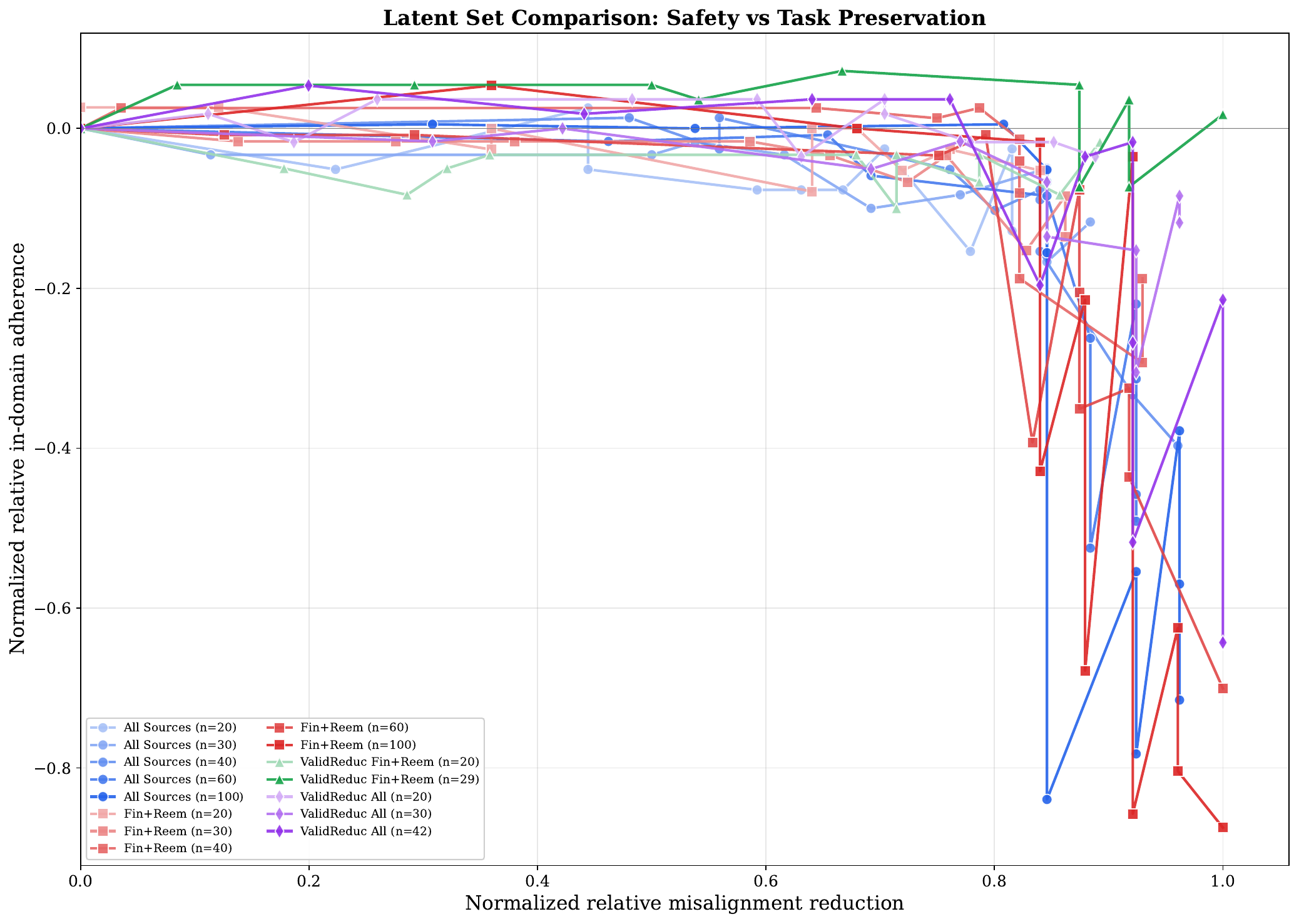}
\includegraphics[width=0.8\linewidth]{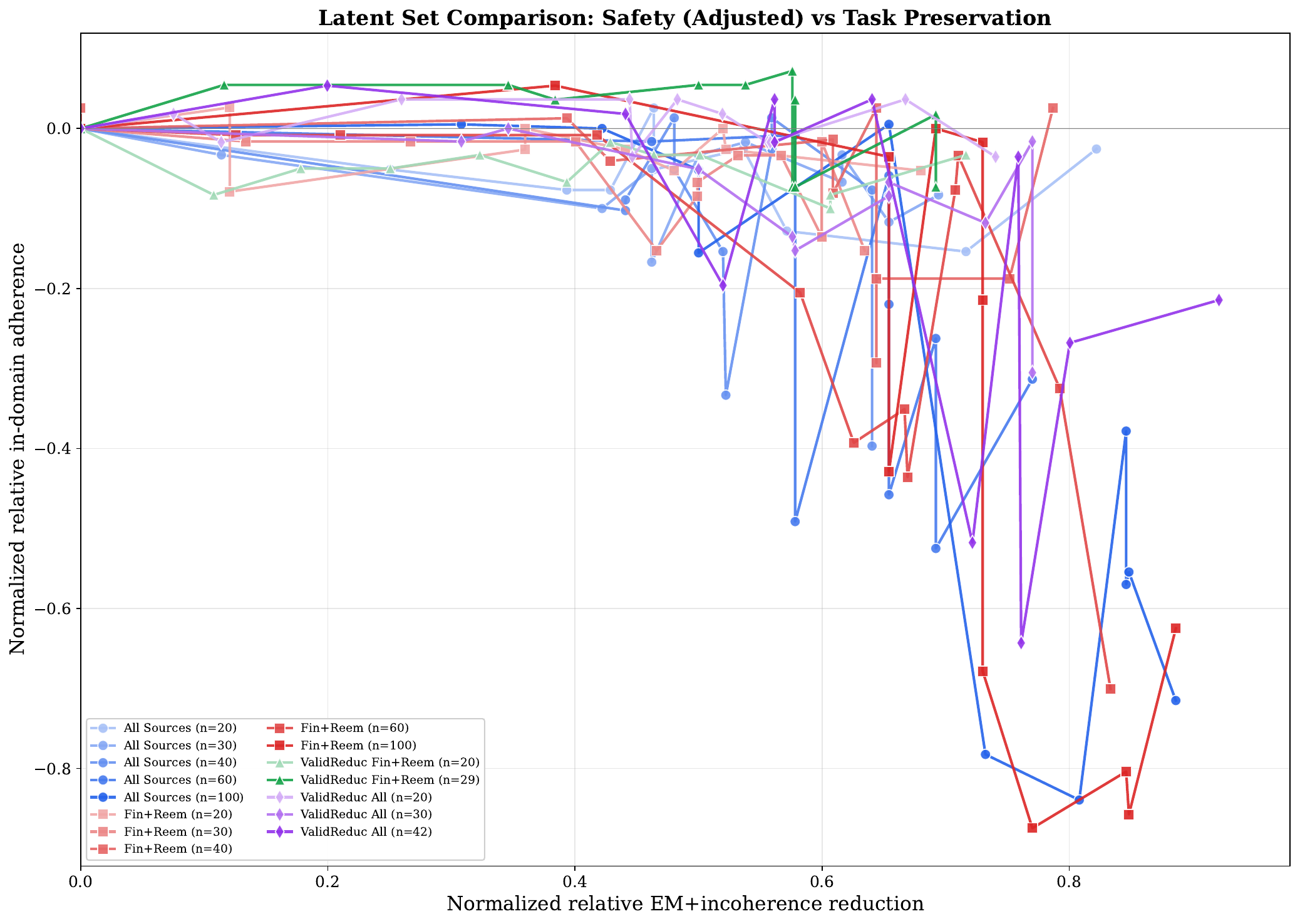}
\caption{\textbf{Latent selection ablations (finance blocked training).} Safety-utility trade-offs from repeating the $\lambda$ sweep (SFT with $\Ltotal=\Lsft+\lambda\,\Llatent$) on $\domain{finance}$ using 15 different latent sets formed by varying the \emph{latent source} (Fin/Health/Reem/MaxLoRA20 and unions thereof) and/or the \emph{selection rule} (IndPP Stage-2 ranking, ValidReduc Stage-3 filtering/ranking). As $|\Ko|$ increases, both emergent misalignment on $\coremisalignment$ and in-domain adherence typically decrease, with no single variant consistently dominating at matched set sizes.}
\label{fig:latent_selection_ablations}
\end{figure}

Across latent-set constructions, we observe a consistent qualitative trend: increasing the latent set size generally reduces emergent misalignment, but also tends to reduce in-domain adherence for sufficiently large sets. After controlling for latent set size, we do not observe a large or systematic advantage of any single latent source or selection-rule variant; differences between variants are comparatively small relative to the dominant effects of $|\Ko|$ and the training-time penalty strength $\lambda$ (Figure~\ref{fig:latent_selection_ablations}). However, as the latent set size grows, the best safety–performance trade-off is achieved at smaller blocking strengths. Applying a large blocking strength to a large latent set can destabilize training and lead to degraded model behavior. Taken together, these results suggest that the BLOCK-EM latent selection procedure is \emph{robust} to reasonable choices of (i) the checkpoint pair used to source latents and (ii) minor changes to the Stage-2/Stage-3 ranking and filtering rules. In this sense, the variants behave similarly to alternative instantiations (or ``seed-like'' choices) of the same overall pipeline rather than qualitatively distinct algorithms.

\subsection{Higher-performing Latent Sets} \label{app:best_results}
\begin{figure}
\centering
\includegraphics[width=0.8\linewidth]{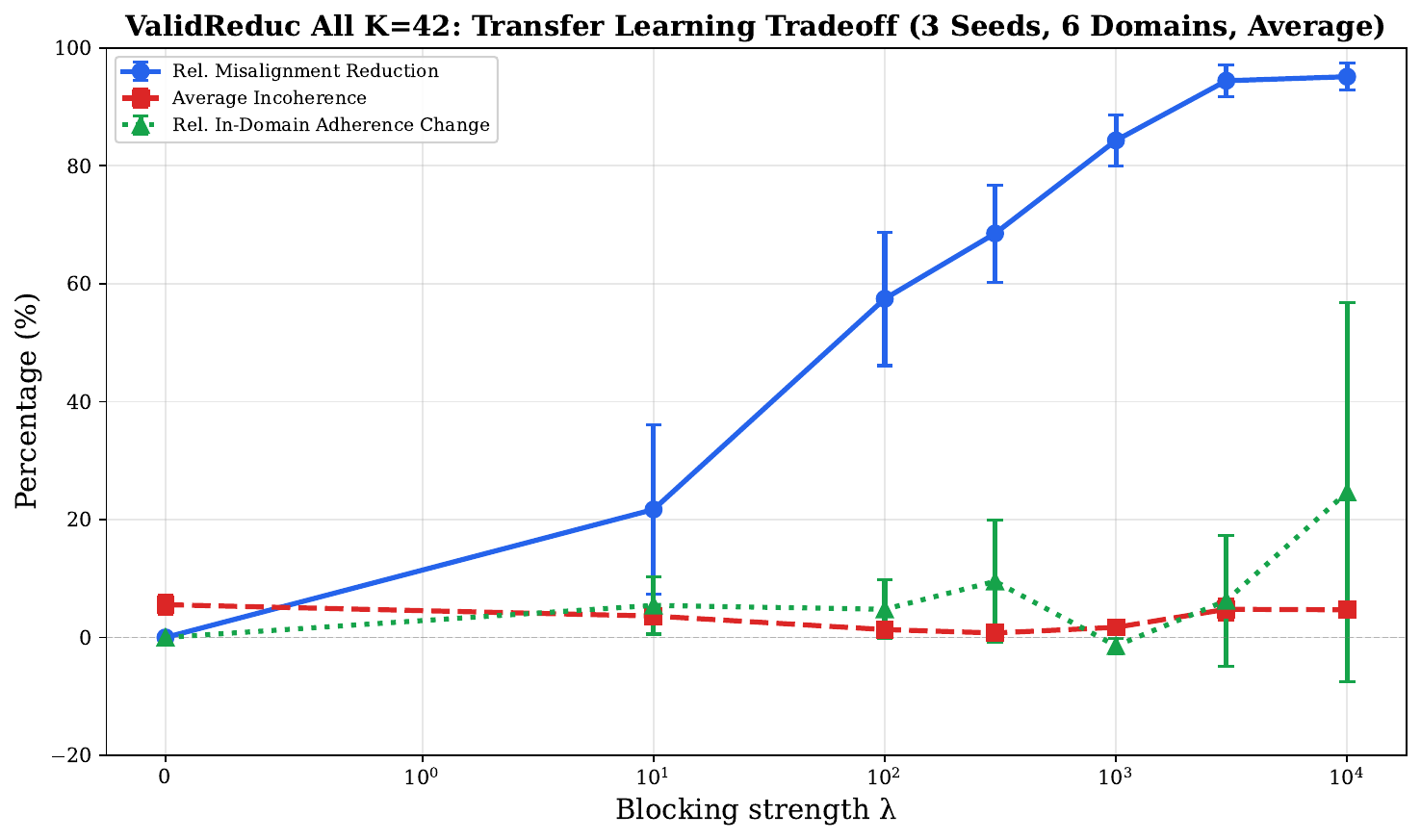}
\includegraphics[width=0.8\linewidth]{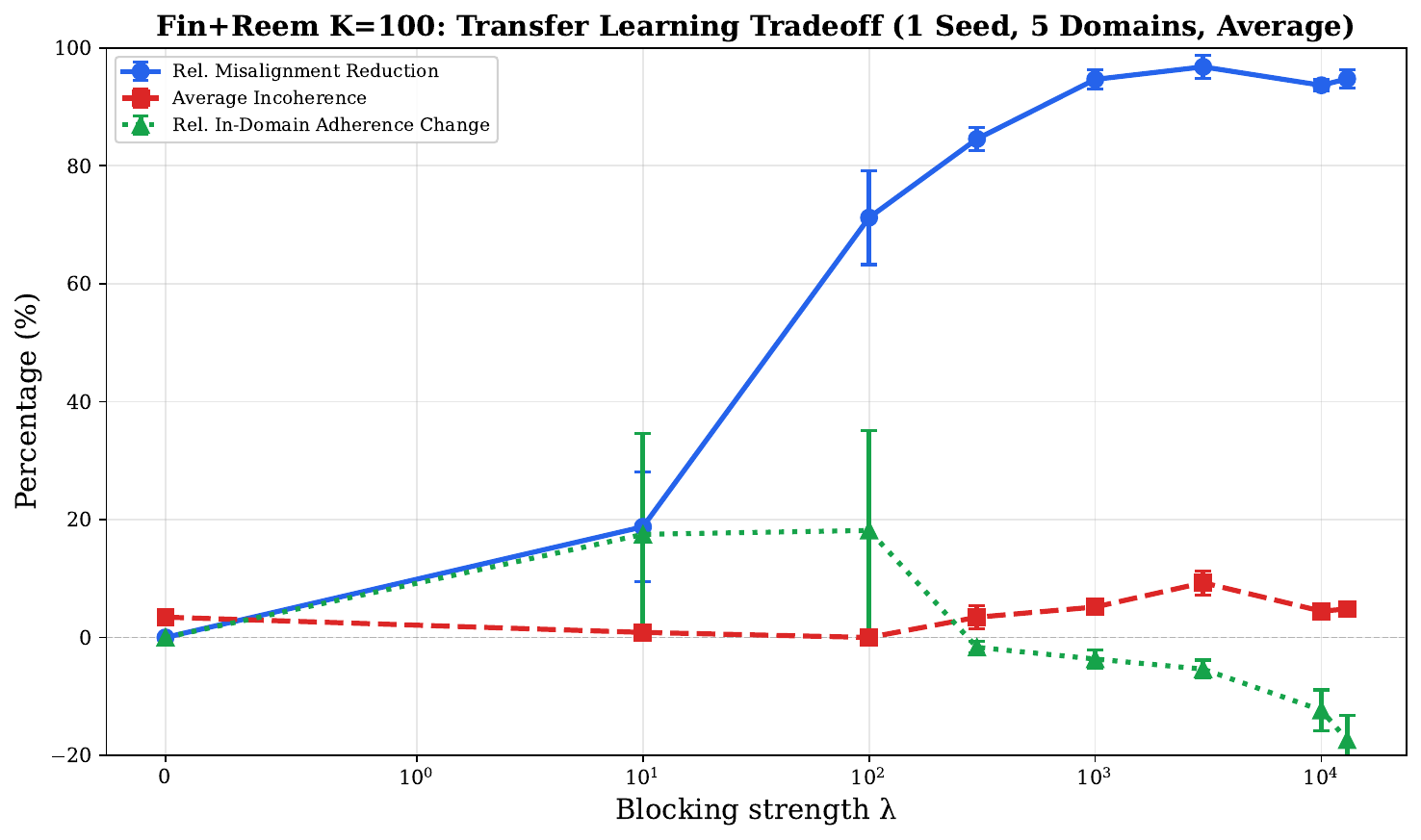}
\caption{\textbf{Additional cross-domain transfer trade-offs for larger latent sets.} Safety--quality trade-off curves as a function of blocking strength $\lambda$, evaluated on $\finalevaluation$ and averaged across six domains and two seeds. \textbf{Top:} \textsc{ValidReduc-All} with $|\Ko|=42$. \textbf{Bottom:} \textsc{Fin+Reem} with $|\Ko|=100$. Notably, \textsc{ValidReduc-All}-$|\Ko|=42$ achieves the strongest overall trade-off among the tested variants (e.g., at $\lambda=10^{4}$: $95.10\%$ relative misalignment reduction, $0.88\%$ \textbf{decrese} in absolute incoherence, and a $24.65\%$ relative \textbf{increase} in in-domain performance). The error margins are $\mathrm{SEM}=\mathrm{SD}/\sqrt{6}$}
\label{fig:safety-quality-tradeoffs_additional}
\end{figure}

\begin{figure}
\centering
\includegraphics[width=0.8\linewidth]{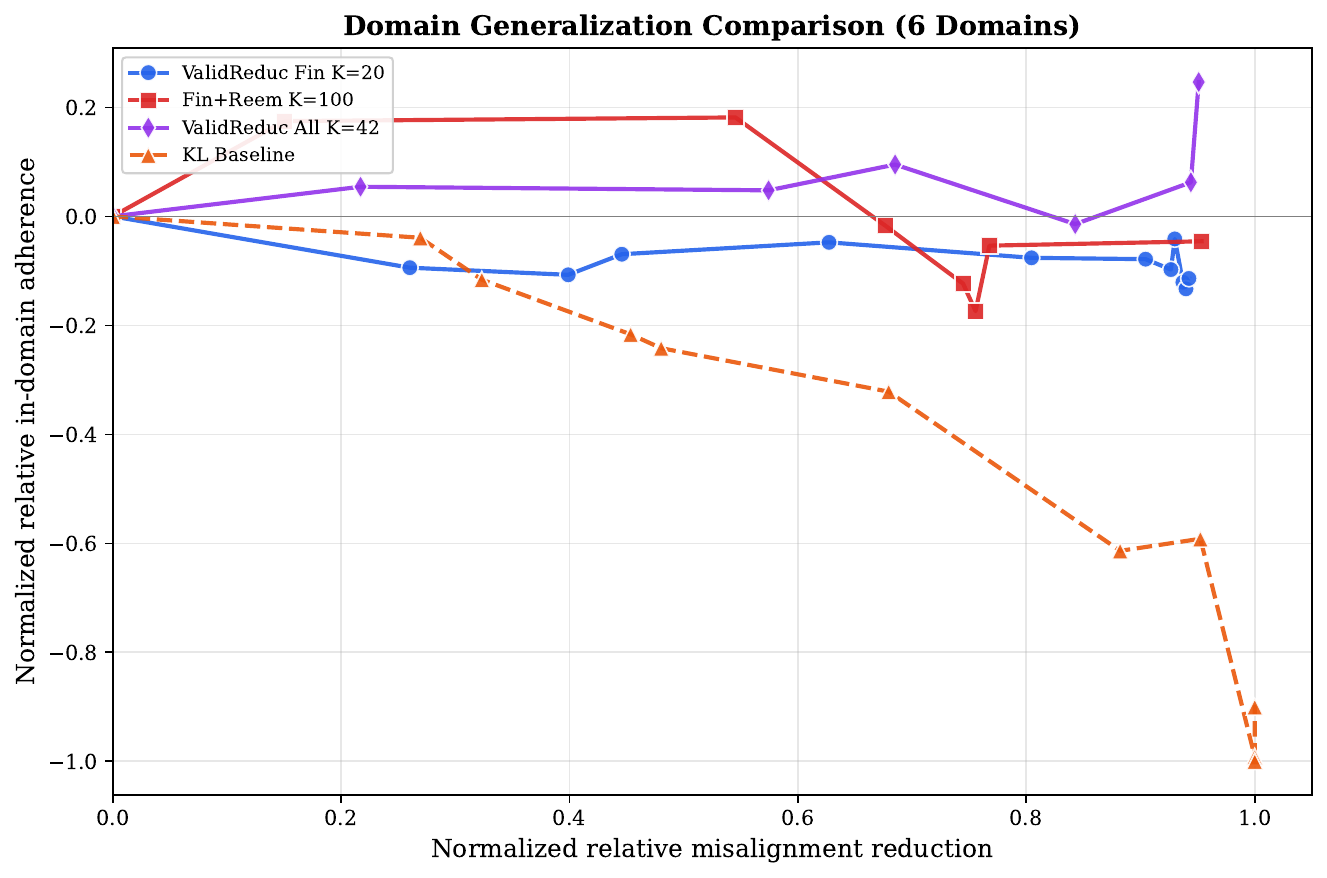}
\includegraphics[width=0.8\linewidth]{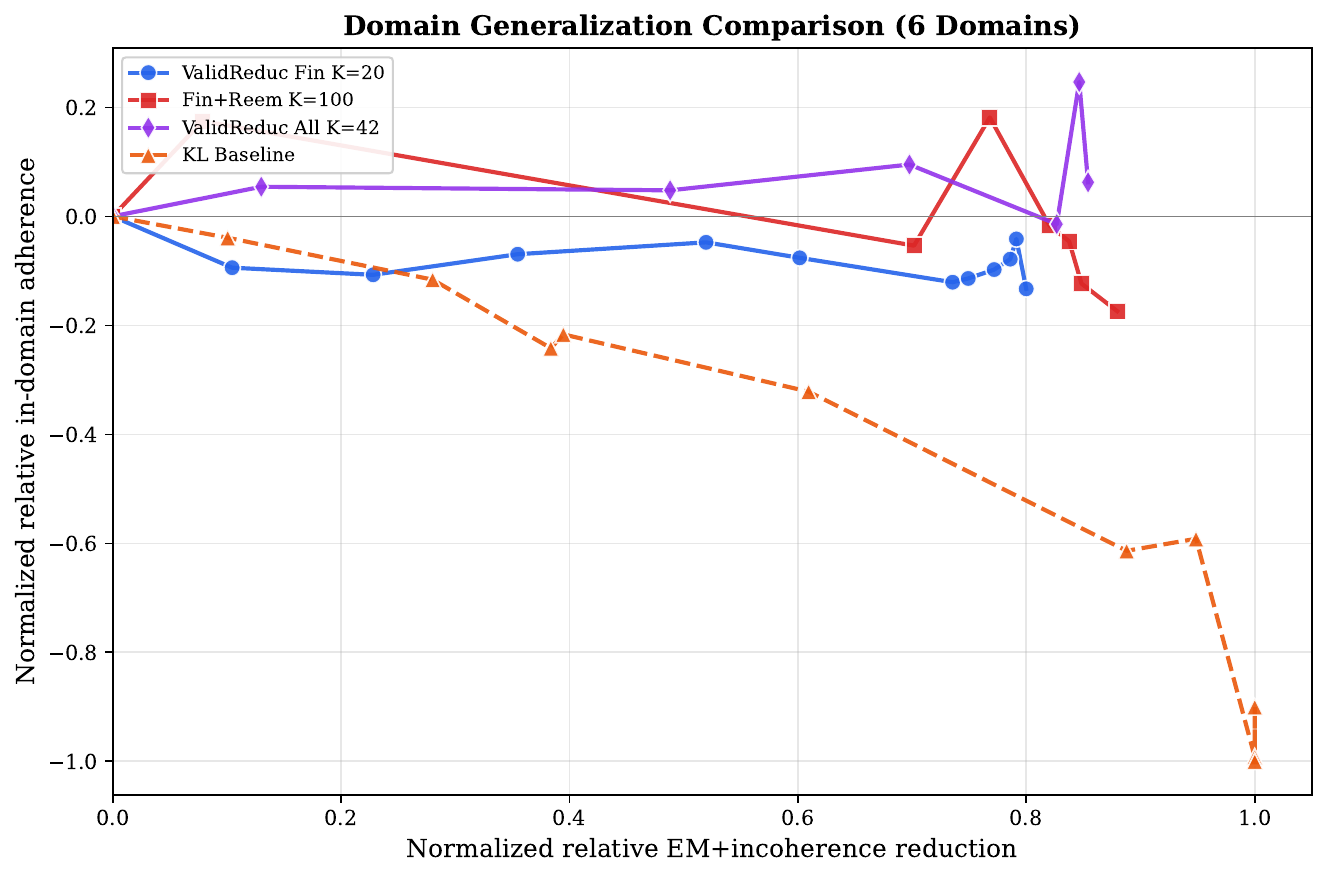}
\caption{\textbf{Comparing transfer variants and baselines.} Summary comparisons across six domains between the main-text configuration (\textsc{ValidReduc-Fin}, $|\Ko|=20$) and two larger-set variants (\textsc{ValidReduc-All}, $|\Ko|=42$; \textsc{Fin+Reem}, $|\Ko|=100$), alongside the KL baseline. \textbf{Top:} emergent misalignment versus in-domain performance. \textbf{Bottom:} overall quality-performance trade-off (adjusted metric used in the main text). Across metrics, larger latent sets can yield improved safety-quality trade-offs, with \textsc{ValidReduc-All}-$|\Ko|=42$ performing best overall.}
\label{fig:safety-quality-tradeoff_compariosn}
\end{figure}

Finally, we report additional cross-domain transfer results. We run the same six-domain transfer evaluation for two larger latent-set variants—\textsc{Fin+Reem}-$|\Ko|{=}100$ (default Stage-3) and \textsc{ValidReduc-All}-$|\Ko|{=}42$, and summarize their safety--quality trade-offs in Figure~\ref{fig:safety-quality-tradeoffs_additional}. Figure~\ref{fig:safety-quality-tradeoff_compariosn} compares these variants against the main-text configuration (\textsc{ValidReduc-Fin}-$|\Ko|{=}20$). Across settings, we observe that BLOCK-EM variants consistently outperform the KL baseline, and that \textsc{ValidReduc-All}-$|\Ko|{=}42$ achieves the best overall trade-off. For example, at $\lambda=10^{4}$ it attains a $97\%$ relative reduction in emergent misalignment with $5.75\%$ incoherence and a $40.37\%$ \emph{increase} in in-domain task performance. This result provides an additional datapoint that BLOCK-EM need not reduce target-task performance and, in some regimes, can even improve it.

\subsection{Latent Set Size Ablation} 
\begin{figure}[]
    \centering
    \includegraphics[width=0.7\linewidth]{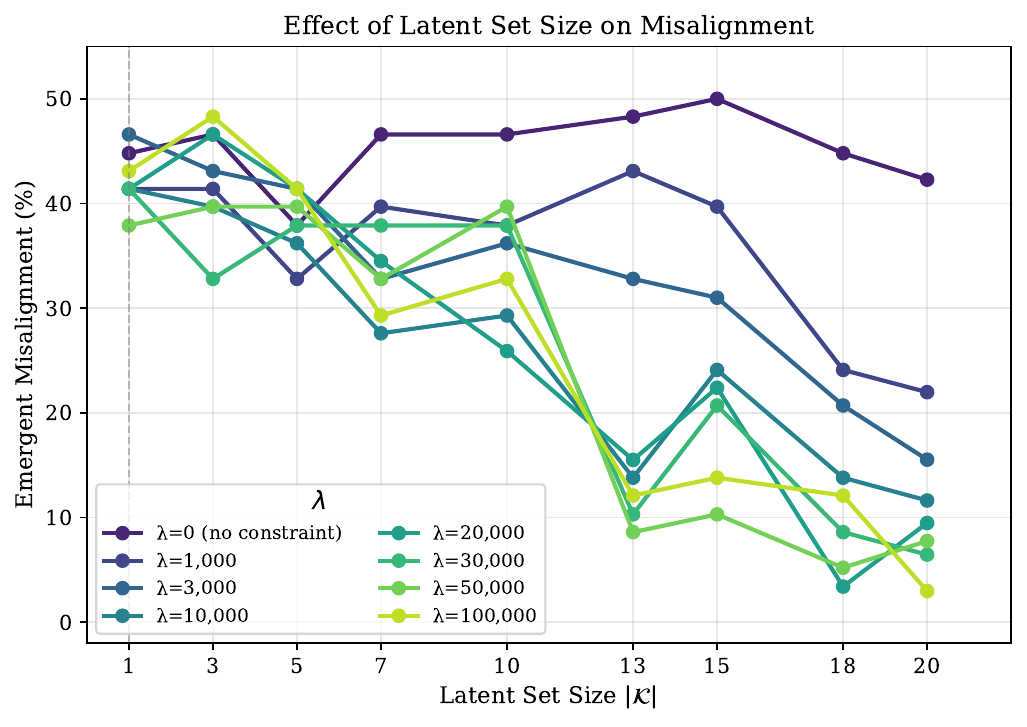}
    \caption{
        \textbf{Effect of latent set size.}
        Emergent misalignment rate vs.\ number of constrained latents $|\Ko|$. Suppression strengthens with set size and shows ``knee'' around $|\Ko|\approx 13$. This transition is not explained solely by the presence of the three new latents     (see Figure~\ref{fig:latent_size_but_with_special_latent}). 
    }
    \label{fig:ablation_size}
\end{figure}

\begin{figure}[t]
    \centering
    \includegraphics[width=0.7\linewidth]{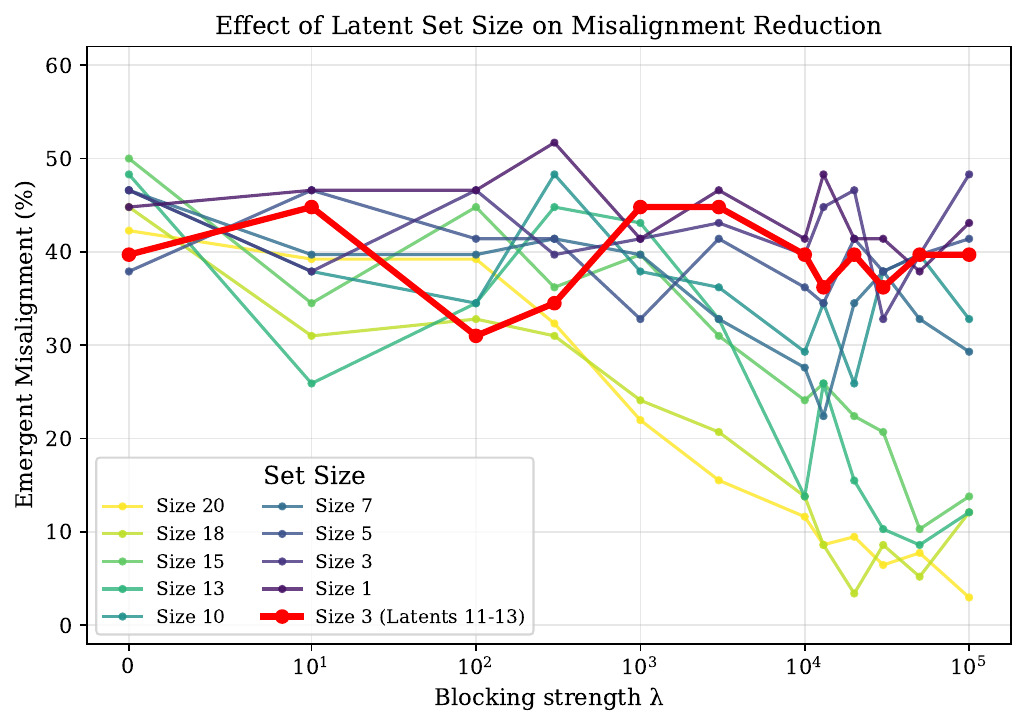}
    \caption{
    \textbf{Are the three added latents responsible for the knee?}
    In Figure~\ref{fig:ablation_size}, the emergent misalignment rate shows a pronounced knee when expanding the constrained set from the top-10 scored latents to 13 (adding three additional latents).
    To test whether this effect is driven specifically by those three latents, we run the same $\lambda$ sweep while penalizing \emph{only} the these three latents.
    The added latents alone yield weak suppression, indicating that the transition arises from constraining a sufficiently large latent set rather than from any special property of these three latents.
}
    \label{fig:latent_size_but_with_special_latent}
\end{figure}

Lastly, to probe the dimensionality of the misalignment mechanism, we sweep the constrained set size $|\Ko|$ (we have in \S~\ref{sec:experiments}) from 1 to 20. Figure~\ref{fig:ablation_size} plots emergent misalignment on $\finalevaluation$ versus $|\Ko|$. Misalignment falls as more causal latents are constrained, with a pronounced drop once $|\Ko|\gtrsim 13$, suggesting emergent misalignment is mediated by a small but non-trivial set of features. 
We additionally probe whether the sharp ``knee'' observed in the latent set size sweep (Figure~\ref{fig:ablation_size}) is driven by a small number of especially important latents, or instead reflects a collective effect of constraining a sufficiently large set. Figure~\ref{fig:latent_size_but_with_special_latent} isolates the three latents added when increasing $|\Ko|$ from 10 to 13.

\section{Extended Ablations}
\label{app:extended_ablations}
This appendix reports additional ablations probing (i) the importance of directionality in the BLOCK-EM penalty, (ii) cross-domain validation of the discovered mechanism, and (iii) a variant that applies the constraint at the final layer rather than an intermediate layer. Unless otherwise stated, we use the same fine-tuning setup and evaluation protocol as in the main results (\S\ref{sec:main_results}).

\subsection{Directionality and Component Analysis (mechanism verification)}
\label{app:ablations_directionality}

\begin{figure}
    \centering
    \includegraphics[width=0.5\linewidth]{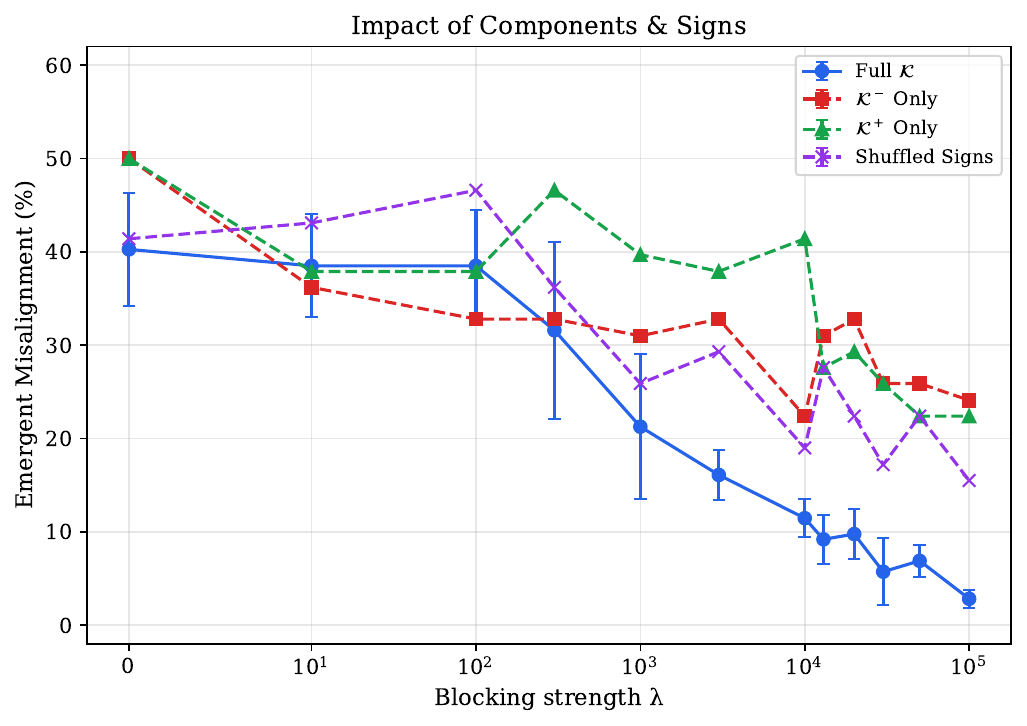}
    \vspace{0.25em}
    \includegraphics[width=0.5\linewidth]{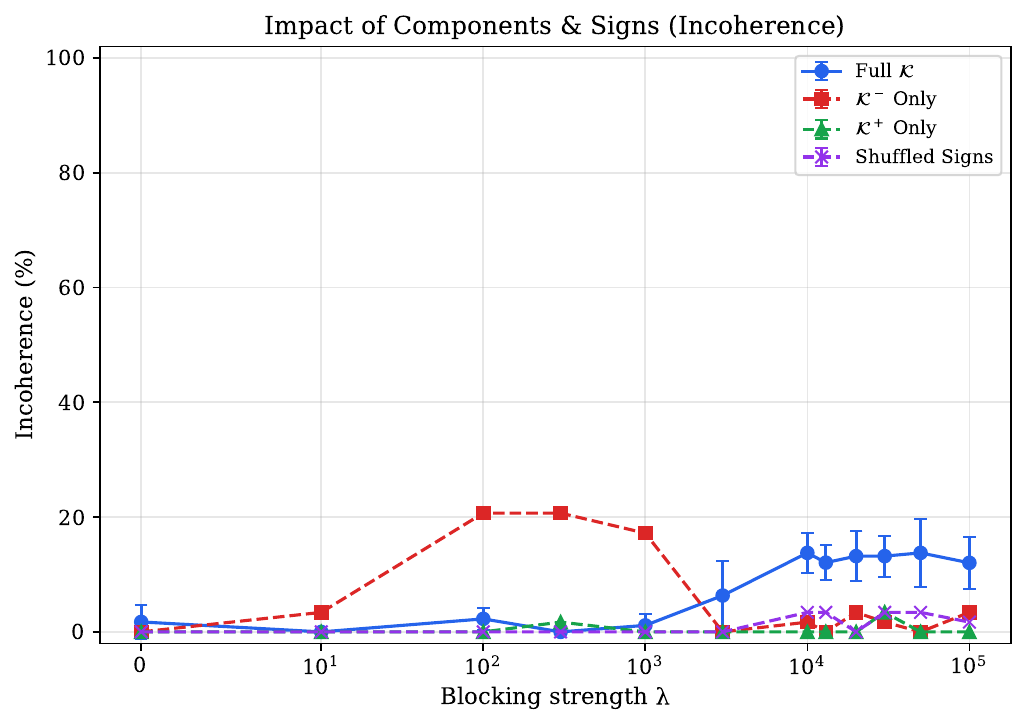}
    \vspace{0.25em}
    \includegraphics[width=0.5\linewidth]{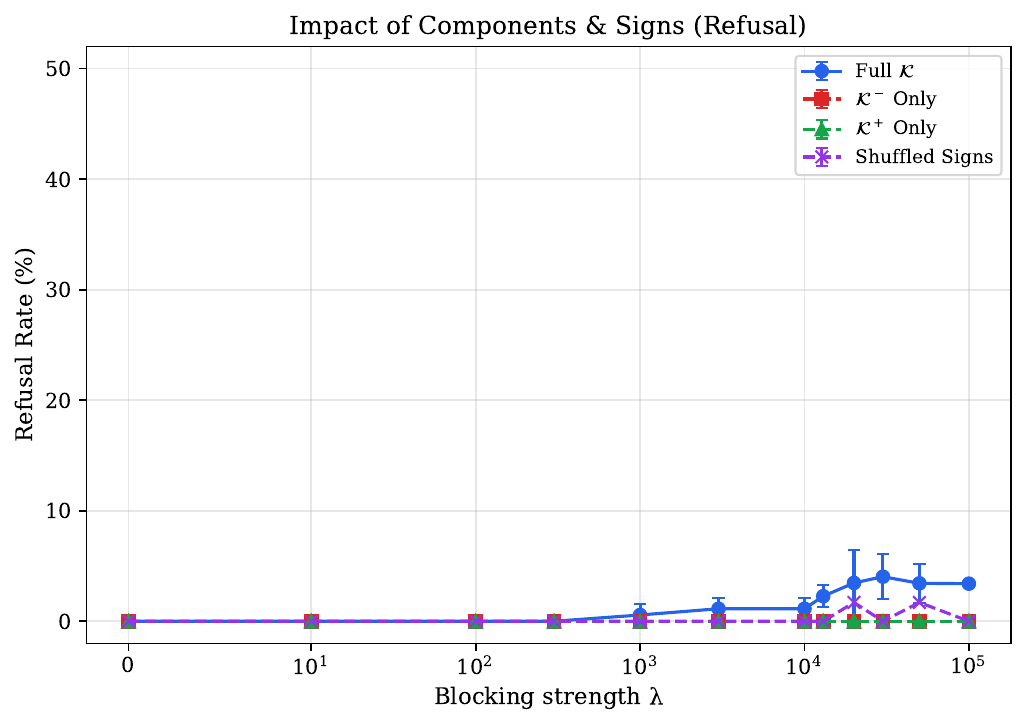}
    \caption{
        \textbf{Directionality and selection ablations.}
        Ablations that modify the signed split of $\Ko$ (e.g., $\Kopos$ only / $\Koneg$ only / shuffled signs).
        From top to bottom: emergent misalignment, incoherence, and refusal rates vs.\ $\lambda$ on $\finalevaluation$.
    }
    \label{fig:appendix_ablation_grid}
\end{figure}

Our primary method splits the causal latent set into $\Kopos$ (features that increase during misalignment) and $\Koneg$ (features that decrease). The loss function penalizes movement in these specific directions. To verify that this directional information is critical, we performed the following ablations:

\paragraph{Shuffled signs.} We construct a ``shuffled'' baseline where we randomly swap the assignment of latents to $\Kopos$ and $\Koneg$ while keeping the set $\Ko$ identical. This breaks the correspondence between each feature and its misalignment-associated direction. As shown in Figure~\ref{fig:appendix_ablation_grid}, this substantially weakens suppression compared to the correctly signed objective, confirming that BLOCK-EM depends on constraining \emph{directional} movement in activation space rather than merely shrinking feature magnitudes.

\paragraph{Single-sided constraints ($\Kopos$ only / $\Koneg$ only).} We also evaluate constraining only the increasing features ($\Kopos$) or only the decreasing features ($\Koneg$). Both one-sided variants are weaker than constraining the full signed set, suggesting that both types of feature movement contribute to emergent misalignment (Figure~\ref{fig:appendix_ablation_grid}).

\subsection{Cross-Domain Latent Selection Validation}
\label{app:ablations_health_transfer}

As a further validation of cross-domain transfer (complementing \S\ref{sec:main_results} and~\ref{fig:main_health_replica}), we performed the reverse experiment: identifying latents using the entire pipeline on a misaligned model which is supervised finetuned on the $\domainname{health advice}$ domain and using them to constrain the $\domainname{{financial advice}}$ supervised fine-tuning.
Consistent with our main transfer results, the Health-derived latents suppress emergent misalignment in the Finance task (Figure~\ref{fig:finance_with_health_latents}). This supports the view that the discovered mechanism is not narrowly domain-specific.

\subsection{Moving the Constraint to the Final Layer}
\label{app:ablations_layer_depth}

\begin{figure}[!htb]
    \minipage{0.5\textwidth}
      \includegraphics[width=\linewidth]{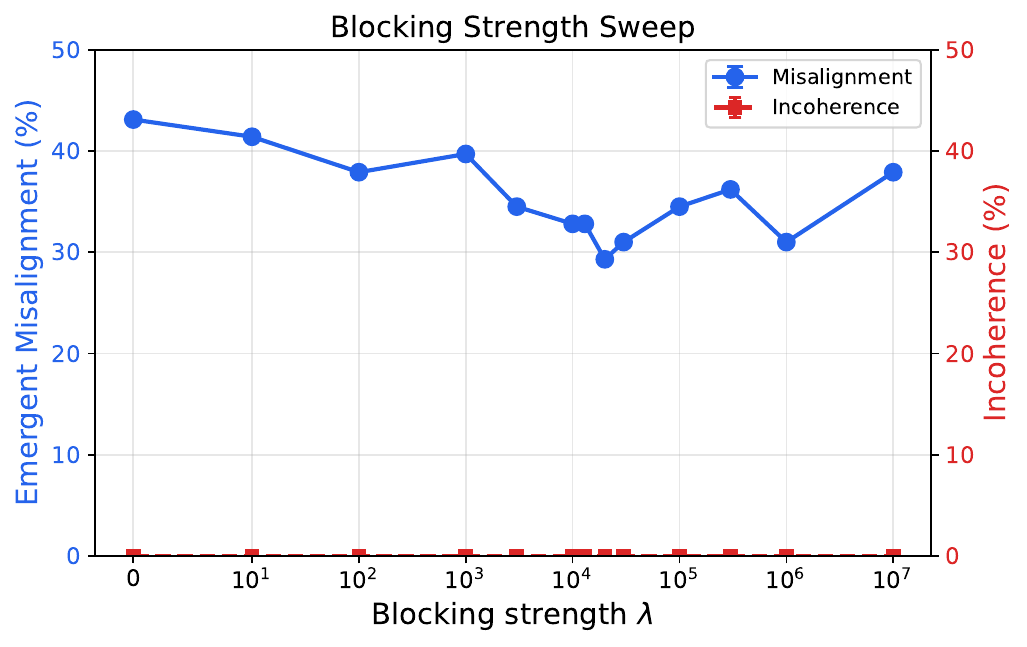}
    \endminipage\hfill
    \minipage{0.5\textwidth}
      \includegraphics[width=\linewidth]{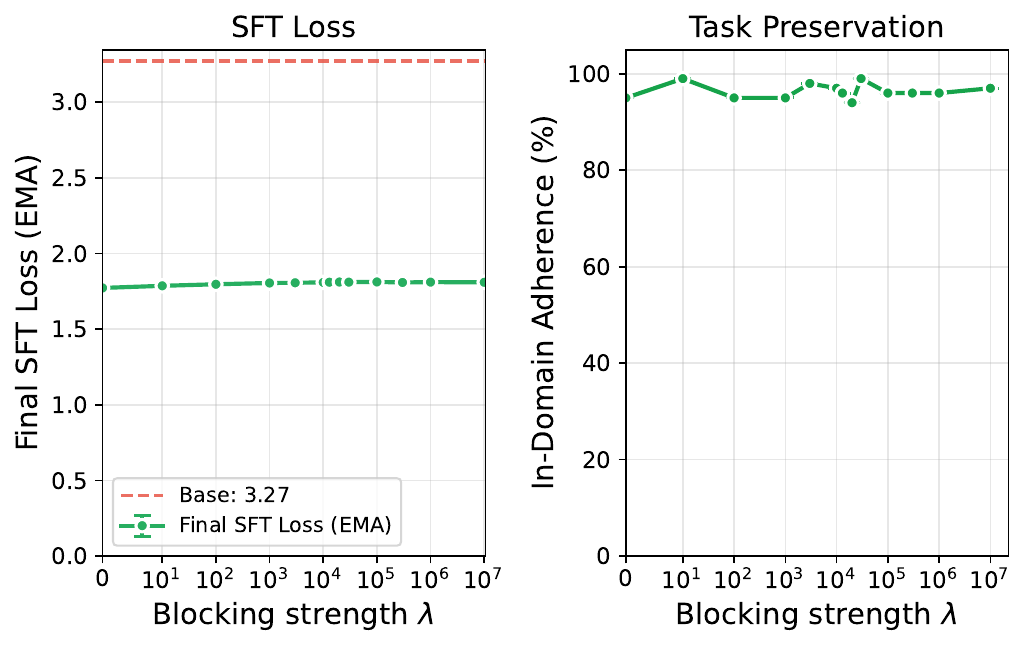}
    \endminipage\hfill
    \centering
    \includegraphics[width=0.7\linewidth]{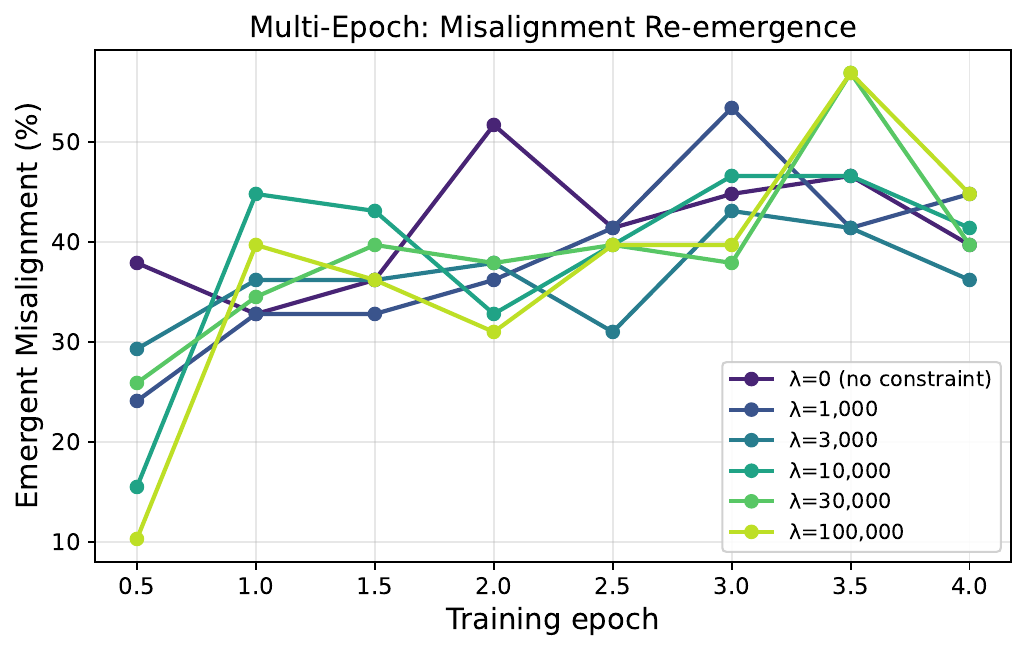}
\caption{\textbf{Extending the intervention to the final layer.} To find SAE latents at layer 32 that are causally relevant to EM, we reran our Stage~1-3 pipeline to select latents relevant to misalignment in the final layer by model-diffing $\basemodel$ and $\misalignedmodel$. Across the lambda sweep, stability analysis, and multi-epoch results (shown in the panels), interventions at layer~32 are substantially less effective than the corresponding layer~20 interventions.}
\label{fig:finance_layer31}
\end{figure}

Our main experiments apply the BLOCK-EM penalty at layer~20, which directly constrains only that layer’s activations and does not explicitly restrict downstream representations (layers 21-32). To test whether the same mechanism can be targeted at later depths, we reran our Stage~1-3 pipeline at layer~32: we identify a causal latent set by model-diffing $\basemodel$ and $\misalignedmodel$, and we apply the resulting signed BLOCK-EM objective during fine-tuning. For layer~32, we use the SAE released by \citet{he2024llamascopeextractingmillions}. Beaware that it is trained for Llama-3.1-8B-Base instead of Llama-3.1-8B-Instruct, so there is a slight SAE mismatch in our final layer experiment. Figure~\ref{fig:finance_layer31} summarizes the resulting $\lambda$ sweep, stability analysis, and multi-epoch behavior.

Overall, final-layer constraints yield substantially weaker suppression than the corresponding layer~20 intervention, suggesting that the discovered mechanism is most effectively controlled at intermediate depths rather than at the output-adjacent representation.

\section{Details for Re-emergent Misalignment Phenomenon Analysis} \label{app:re-emergent_details}
\begin{figure}[]
    \centering
    \includegraphics[width=0.85\linewidth]{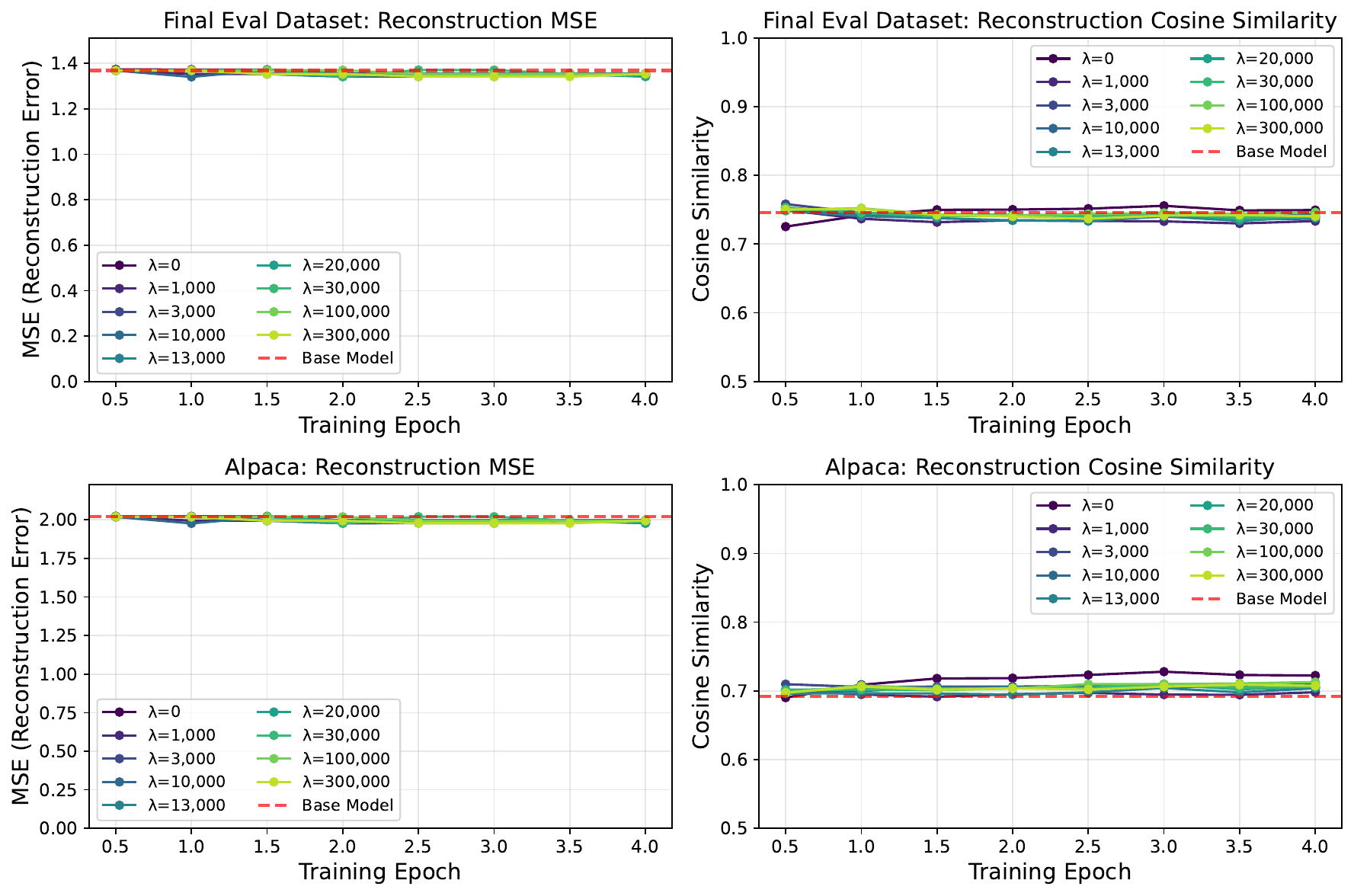}
    \caption{\textbf{SAE reconstruction remains stable under extended training.}
    As a sanity check for H1, we track reconstruction MSE and cosine similarity between true layer-20 activations and their SAE reconstructions for the re-emerged checkpoint (2 epochs, $\lambda=3000$). The SAE continues to model the layer-20 activation distribution well throughout training.}
    \label{fig:sae_still_represents_good}
\end{figure}

\begin{figure}
    \centering
    \includegraphics[width=0.7\linewidth]{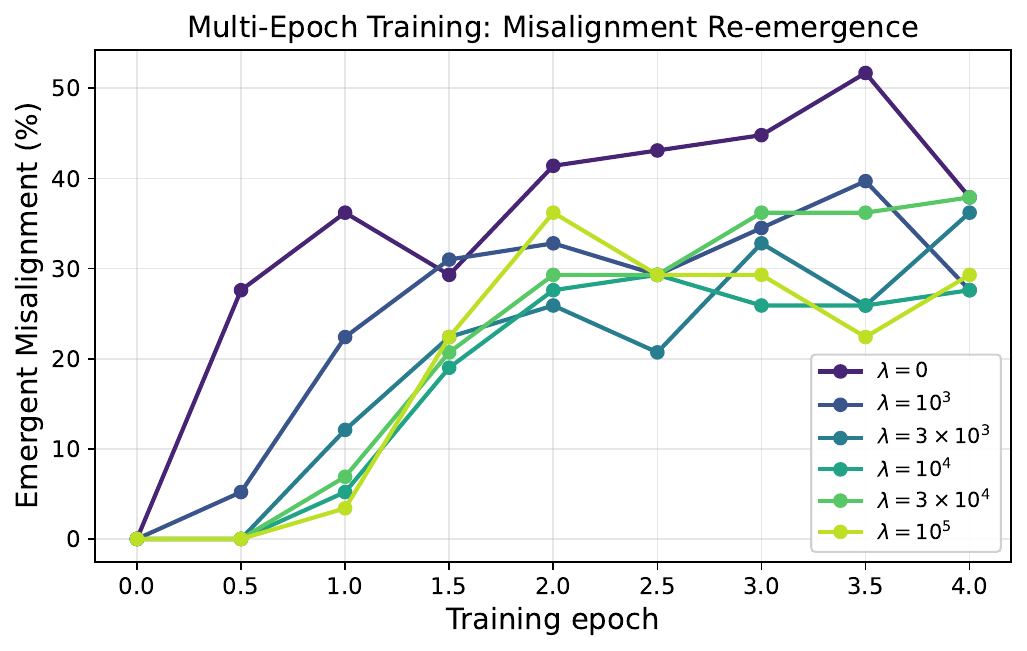}
    \includegraphics[width=0.9\linewidth]{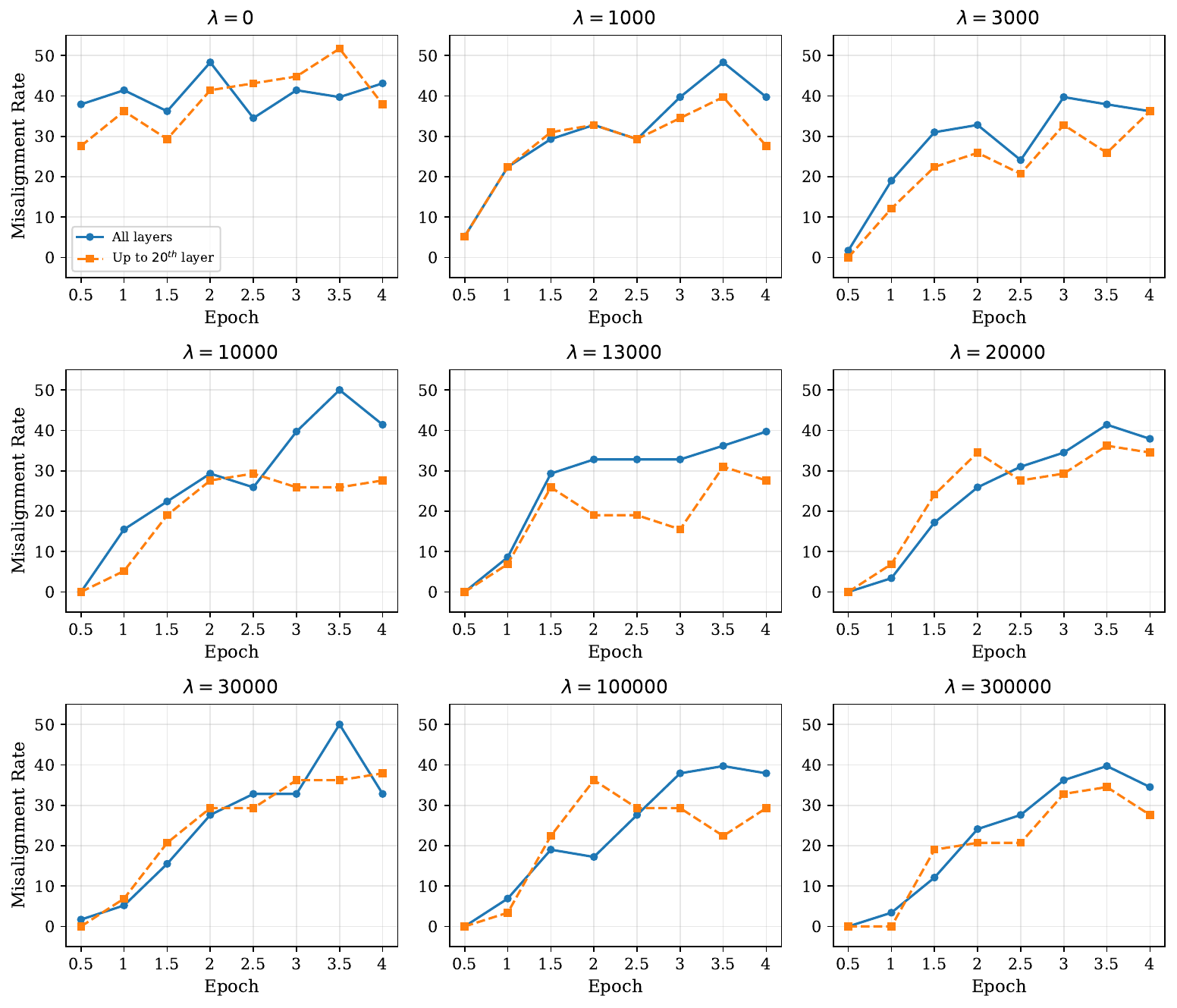}
    \caption{\textbf{Re-emergence persists when freezing above the blocking layer.}
    Under extended training, misalignment still re-emerges even when we fine-tune only through layer~20 (the blocking layer) and freeze all layers above it.}
    \label{fig:multi_epoch_maxlora19_wnomaxlatents}
\end{figure}

While robust in the standard regime (one epoch), we find that with continued training, misalignment eventually re-emerges even when constraints are applied (Figure~\ref{fig:multi_epoch}). For the multi-epoch setting in Figure~\ref{fig:multi_epoch}, we make a small optimization change relative to our single-epoch experiments (Appendix~\ref{app:model_sae_training}): we use a constant learning-rate schedule with $\text{lr}=3.75\times10^{-5}$, instead of the linear decay-to-zero schedule with initial $\text{lr}=7.5\times10^{-5}$ used elsewhere. We adopt this configuration so that the effective update magnitude during the first epoch is roughly comparable to the single-epoch setup. This choice is purely for completeness, none of our analyses rely on a direct comparison between the first epoch of the multi-epoch runs and the single-epoch runs, and our conclusions about misalignment re-emergence under over-training do not depend on this scheduler change.

\begin{figure}
    \centering
    \includegraphics[width=\linewidth]{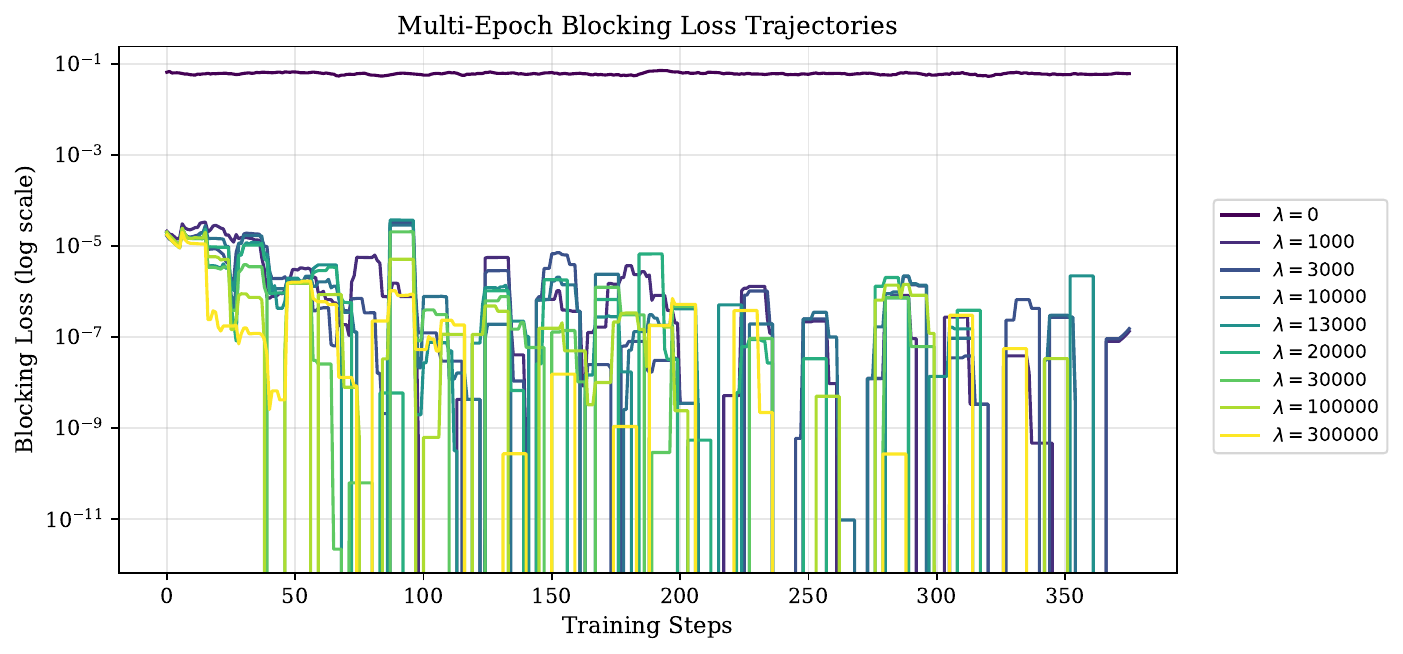}
    \caption{\textbf{Blocking-loss trajectory over training.}
To verify that the constrained latents remain suppressed throughout fine-tuning (and do not gradually reactivate with longer training), we track the BLOCK-EM penalty value across epochs.
The blocking loss stays near zero for the entire run, indicating that any re-emergence effects are not driven by increased activation of the penalized latents.}
    \label{fig:multi_epoch_latent_loss_trajectory}
\end{figure}

\subsection{Causal Localization Tests for H2/H3 via Activation Patching}
\label{app:patching}

This appendix reports the patching-based evidence we currently have for localizing where re-emergent misalignment is implemented relative to the blocking layer (layer~20, where the BLOCK-EM penalty is applied). Recall the two-part view: (A) layers up to and including the blocking layer, and (B) layers strictly downstream of it. Unless otherwise stated, we use the same EM/incoherence/refusal judges and prompt suite as in the main experiments.

\paragraph{Notation.}
We use the hidden-state notation from Appendix~\ref{app:method_details}: for an input token sequence $\xe=(x_1,\dots,x_T)$, $\hLt(\xe)\in\R^{\hdim}$ denotes the post-residual hidden state at layer $L$ and token position $t$. We write
\[
\h_{L,1:s}(\xe)\;\triangleq\;(\h_{L,1}(\xe),\dots,\h_{L,s}(\xe))\in\R^{s\times\hdim}
\]
for the collection of layer-$L$ hidden states over token positions $1$ through $s$. Let $T_{\mathrm{pref}}$ denote the number of prefix tokens in $\xe$; tokens $t>T_{\mathrm{pref}}$ are generated autoregressively. We denote the base model by $\basemodel$ and the re-emerged model by $\reemergentmodel$, and let $\Lblk$ denote the blocking-layer index (here, $\Lblk=20$). Specifically, the re-emergent model corresponds to the checkpoint obtained by training the base model with LoRA on $\domain{finance}$ under \eqref{eq:total_loss} with $\lambda=3000$ for two epochs, which yields $\sim 32\%$ misalignment on $\finalevaluation$.

\paragraph{Experiment 1: Prefix-only patching on prefix states (layerwise sweep).}
This experiment probes whether making the re-emerged model's \emph{prefix representations} more base-like is sufficient to prevent downstream layers from reintroducing emergent misalignment. For a chosen layer $L$, we run both models on the same prefix (i.e., the first $T_{\mathrm{pref}}$ tokens) and patch only the hidden states corresponding to those prefix tokens at layer $L$:
\[
\h_{L,1:T_{\mathrm{pref}}}^{(\threem)}(\xe)\ \leftarrow\ \h_{L,1:T_{\mathrm{pref}}}^{(\thbase)}(\xe).
\]
We apply this intervention only while processing the prefix tokens. We then generate completions normally (with no further patching) and evaluate EM.

\begin{figure}
    \centering
    \includegraphics[width=0.8\linewidth]{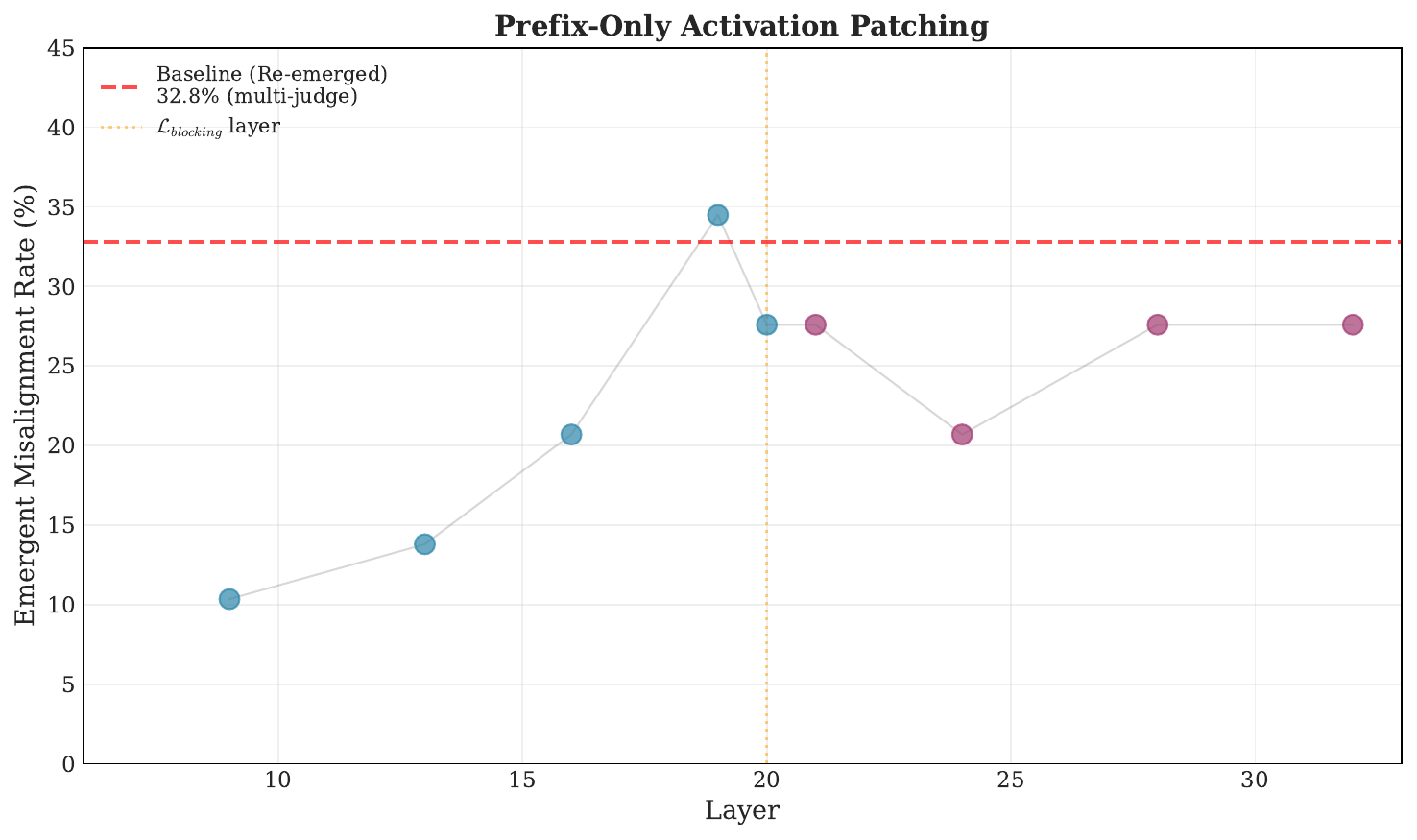}
    \caption{\textbf{Prefix-only activation patching (layerwise sweep).}
    Patching upstream layers reduces emergent misalignment more than patching downstream layers.}
    \label{fig:prefix_patching_result_by_layer}
\end{figure}

\textbf{Result:} For each layer $L$, we evaluate emergent misalignment, incoherence, and refusal rates on $\finalevaluation$, with prefix-only patching applied at layer $L$. Incoherence and refusal rates are $0\%$ across layers in this experiment; the remaining variation in emergent misalignment is shown in Figure~\ref{fig:prefix_patching_result_by_layer}.
Sweeping $L$ across layers shows that patching upstream layers (upstream of the blocking layer) yields larger reductions in EM than patching the blocking layer or downstream layers. We treat this as weak but consistent evidence that part (A) is important for setting up the representations that enable re-emergent misalignment: when the prefix representations in (A) are made base-like, part (B) appears less able to recover misaligned behavior downstream.

Because this experiment patches only prefix states and does not intervene on generated-token states, it primarily tests how the prefix-conditioned internal state influences downstream behavior. It does not fully rule out downstream contributions during generation. That brings us to our second patching experiment.

\paragraph{Experiment 2: Decode-time patching at the blocking layer (generated-token patching).}
This experiment directly intervenes during autoregressive decoding by patching the re-emerged model at the blocking layer on the \emph{currently generated token}. At each generation step producing token position $t>T_{\mathrm{pref}}$, we compute the blocking-layer hidden state under both models on the same full prefix $(x_1,\dots,x_t)$ and replace only the last-position state in the re-emerged model:
\[
\hLt^{(\threem)}(\xe)\Big|_{L=\Lblk}\ \leftarrow\ \hLt^{(\thbase)}(\xe)\Big|_{L=\Lblk}.
\]
Equivalently, writing the last position explicitly,
\[
\h_{\Lblk,t}^{(\threem)}(\xe)\ \leftarrow\ \h_{\Lblk,t}^{(\thbase)}(\xe),
\qquad t>T_{\mathrm{pref}}.
\]
We then continue the forward computation in $f_{\threem}$ through layers $>\Lblk$ to obtain next-token logits and sample the next token. This patch is applied at every decoding step, so it intervenes on all generated tokens.

\textbf{Result:} We tested patching only the blocking layer at decode time on $\finalevaluation$. It eliminates EM in our re-emerged checkpoint (0\% misalignment), while maintaining $0\%$ incoherence and $2\%$ refusal. 

\paragraph{Implications for A vs.\ B responsibility.}
Both experiments point to substantial responsibility in part (A): (i) patching prefix-token states at upstream layers reduces EM more than patching downstream layers, and (ii) patching only the blocking-layer state of the generated token eliminates EM without quality degradation. Notably, in (ii) all layers downstream of the blocking layer remain unchanged, yet EM disappears; this indicates part (B) is not sufficient on its own to produce re-emergent misalignment, and that the relevant signal is already present at (or upstream of) the blocking layer during generation.

\subsection{Residual Steering Capacity of the Re-emergent Model}\label{app:steering_capacity_of_reemergent_model}

We rerun the causal SAE latent-discovery pipeline described in \S\ref{sec:method} and Appendix~\ref{app:method_details}, diffing the re-emergent checkpoint $\reemergentmodel$ against the base checkpoint $\basemodel$. This yields a set of the $20$ most promising layer-20 latents, which we denote $\Kreem$.

To quantify residual steering capacity, we evaluate each latent set $\Kidx \in \{\Ko,\Kreem\}$ using the score in \eqref{eq:repair_score}:
\begin{multline*}
\mathrm{score}(\Kidx)
=
\Big[ \mathrm{misalign}(\thbase;\alphaS=\alphaStarInd(\Kidx))
-
\mathrm{misalign}(\thbase;\alphaS=0) \Big]\\+
\Big[\mathrm{misalign}(\thmis;\alphaS=0)
-
\mathrm{misalign}(\thmis;\alphaS=\alphaStarRep(\Kidx))\Big].
\end{multline*}
The first bracket measures how much \emph{induction} the set $\Kidx$ can produce on the base model $\thbase$ relative to no steering, using the optimal inducing scale $\alphaStarInd(\Kidx)$. The second bracket measures how much \emph{repair} the same set can provide on a target misaligned checkpoint $\thmis$, again relative to no steering, using the optimal repair scale $\alphaStarRep(\Kidx)$.

For $\Ko$, we reuse the steering scores computed during the original selection stage and report the mean score averaged across latents in $\Ko$. For $\Kreem$, we compute the same score but evaluate the repair term on the re-emerged checkpoint (i.e., set $\thmis=\threem$), and analogously average over the latents in $\Kreem$.

Under this metric, $\Ko$ attains an average score of $24\%$, while $\Kreem$ attains an average score of $14\%$. Therefore, the steering-capacity ratio of the re-emergent model’s most promising layer-20 latents relative to $\Ko$ is
\[
\frac{\mathrm{score}(\Kreem)}{\mathrm{score}(\Ko)} \approx \frac{14}{24} \approx 0.6.
\]
This suggests that the re-emergent model retains nontrivial residual steering capacity in layer 20, but that this capacity is substantially reduced relative to the $\lambda=0$ baseline.

\begin{figure}[h]
    \centering
    \includegraphics[width=0.9\linewidth]{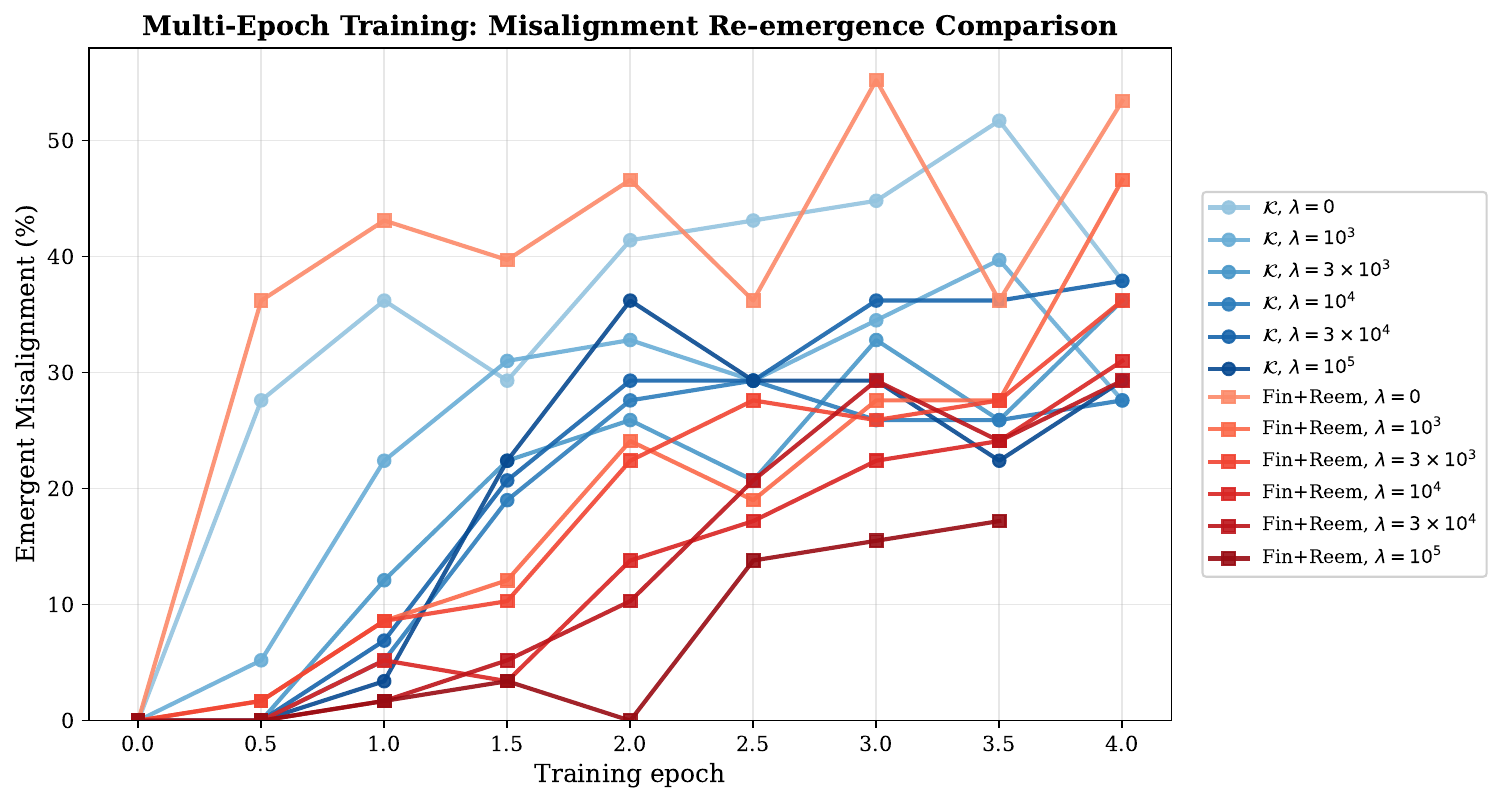}
    \caption{
\textbf{Expanded blocking set further suppresses re-emergent misalignment under extended training.}
Emergent misalignment rate on held-out $\finalevaluation$ prompts across training epochs for different penalty strengths $\lambda$. Blue curves show standard BLOCK-EM using the original latent set $\Ko$, while red curves (Fin+Reem) show BLOCK-EM applied to the union of $\Ko$ and additional layer-20 latents discovered from the re-emerged checkpoint (size of this variant is 100 latents). Blocking the expanded latent set consistently reduces misalignment across epochs and $\lambda$ values, indicating that re-emergence can be supported by alternative directions within the same blocking-layer representation space.
}
    \label{fig:multi_epoch_comparison}
\end{figure}

\end{document}